\documentclass{article}
\usepackage{amsmath,amsthm,amssymb,amsfonts,mathtools,bm,bbm,array,float,mathdots,dsfont,mathrsfs}
\usepackage{caption,subcaption,multirow,longtable,lscape,wrapfig,tikz,comment}
\usepackage{hyperref}
\hypersetup{allcolors=black}
\usepackage{graphicx,tikz,tikz-cd}
\usetikzlibrary{positioning}
\tikzset{
  node/.style={circle, draw=blue!70!black, fill=blue!20, thick, minimum size=4.5mm, inner sep=0pt},
  edge/.style={->, line width=0.9pt, >=Latex, shorten >=1pt, shorten <=1pt},
  every label/.style={font=\small}
}
\usepackage[top=1in, bottom=1in, left=1in, right=1in]{geometry}
\usepackage[absolute]{textpos}
\usepackage[toc,page]{appendix}
\usepackage[utf8]{inputenc}
\usepackage[autostyle=true]{csquotes}
\allowdisplaybreaks

\def\bea{\begin{eqnarray}}
\def\eea{\end{eqnarray}}
\def\cA{ { \mathcal{A}}}
\def\cB{ \mathcal{B}}
\def\cS{ \mathcal{S}}
\def\Dim{ {\rm{Dim}}}

\begin{document}
\begin{textblock}{5}(12,1)
\noindent QMUL-PH-25-28
\end{textblock}
\title{\bf Approximate Gaussianity \\ Beyond Initialisation in Neural Networks}
\author{\bf Edward Hirst\textsuperscript{1}\footnote{e.hirst@qmul.ac.uk}, Sanjaye Ramgoolam\textsuperscript{1}\footnote{s.ramgoolam@qmul.ac.uk}}
\date{\small\today \\ \textsuperscript{1} \textit{\small Centre for Theoretical Physics, \\Queen Mary University of London,\\ Mile End Road, London E1 4NS, UK}\\
}

\maketitle

\begin{abstract}
Ensembles of neural network weight matrices are studied through the training process for the MNIST classification problem, testing the efficacy of matrix models for representing their distributions, under assumptions of Gaussianity and permutation-symmetry. The general 13-parameter permutation invariant Gaussian matrix models are found to be effective models for the correlated Gaussianity in the weight matrices, beyond the range of applicability of the simple Gaussian with independent identically distributed matrix variables, and notably well beyond the initialisation step. The  representation theoretic model parameters, and the graph-theoretic characterisation of the permutation invariant matrix observables give an interpretable framework for the best-fit model and for small departures from Gaussianity. Additionally, the Wasserstein distance is calculated for this class of models and used to quantify the movement of the distributions over training. Throughout the work, the effects of varied initialisation regimes, regularisation, layer depth, and layer width are tested for this formalism, identifying limits where particular departures from Gaussianity are enhanced and how more general, yet still highly-interpretable, models can be developed.
\end{abstract}

\thispagestyle{empty}
\clearpage
\setcounter{page}{1}
\numberwithin{equation}{section}
\numberwithin{figure}{section}
\numberwithin{table}{section}
\tableofcontents
\clearpage 

%%%%%%%%%%%%%%%%%%%%%%%%%%%%%%%%%%%%%%%%%%%%%%%%%
\section{Introduction}
As universal function approximators \cite{Hornik1989MultilayerFN, Hanin_2019}, neural networks (NNs), have seen a recent explosive growth in applications  across scientific research \cite{anderson1995introduction, Cartwright2009, He:2023csq}. 
Their successes may largely be attributed to their ubiquitous efficacy, structural scalability, and ease of implementation.
However, despite these widespread practical applications, they are  often vastly over-parameterised so that interpretation and distillation of mathematical insights into the problems they solve are difficult.

Furthermore, as each instantiation of a NN comes from random samples of the initialisation distributions, and due to the highly non-convex nature of the loss function for many problems, the final parameter configuration of trained NN models often varies wildly.
Therefore an understanding of what features of the parameter set lead to a good model is highly desired, and has inspired a whole field of research into interpretability measures.
Examples of productive directions of research in this area include deduction of redundant parameters through information-theoretic techniques \cite{Berman:2023rqb, Berman:2024pax, Manning-Coe:2025uhs}, and gradient saliency approaches to uncover mathematical correlations between inputs \cite{Davies2021, Berglund:2021ztg, Cheung:2022itk, Costantino:2024joa}, yet direct analysis of trained parameter distributions has identified broad behaviours \cite{ctcs, bnnps} with limited production of explicit compressed models.

This work focuses on symmetry-based techniques for understanding the NN functions' parameter space, and creating lower-dimensional representations.
Symmetry-based  methods have already inspired design of invariant NNs \cite{pmlr-v48-cohenc16, invariant_nns} which implicitly respect desired symmetry, and have uncovered insight into the impact of loss \cite{sym_learningconstraints} and data \cite{sym_mfv} symmetries onto the parameter space.
Continuous symmetries of NN parameters have been studied in \cite{Hashimoto:2024rms}, using gauge symmetry ideas for geometric interpretation as spacetime diffeomorphisms, 
aligning with a program of work connecting NNs to quantum field theories \cite{Halverson:2020trp, Maiti:2021fpy, Erbin:2021kqf}.
Here, we focus on the complement of continuous symmetries, those which are discrete, and use another set of techniques known as \textit{matrix models} complemented with ideas from quantum field theory. 

Probability distributions over matrices have long been studied in mathematical physics with the motivation of modelling real world matrix data arising in a variety of contexts. 
Works of Wigner and Dyson introduced and developed  the idea of modelling the statistical properties of highly excited discrete energy levels of complex nuclei, eigenvalues of complex Hamiltonians, using eigenvalue distributions derived from Gaussian matrix measures \cite{Wigner1951, Dyson1962, Wigner1967}. 
The matrix distributions in these traditional applications have continuous symmetries, such as $SO(N), U(N), Sp(N)$ for matrices of size $N$.
Robust statistical evidence for the agreement between nuclear spectra and random matrix theory (RMT) predictions, based on an ensemble of nuclei, was provided in \cite{Haq1982}. 
RMT was also found to provide an effective description of eigenvalue distributions in chaotic quantum dynamical systems \cite{Bohigas1984}, and in Rydberg atom energy levels in strong electromagnetic fields \cite{Delande1986}. 
In recent years, evidence has been built that eigenvalue distributions from RMT (specifically the Marchenko–Pastur distribution) are applicable to stock price correlations \cite{Laloux1999, Plerou1999}. 
Reviews of these applications in physics as well as other disciplines are given in \cite{Guhr1998, Akemann2011}.

Recent developments have also highlighted the utility of traditional RMT methods in the study of neural networks, where eigenvalue spectra of weight matrices reveal universal statistical properties akin to those observed in physical systems. 
Analyses of trained deep networks demonstrate that their weight matrices often exhibit spectral signatures predicted by RMT, such as heavy-tailed distributions or deviations from the Marchenko–Pastur law, which serve as indicators of network capacity, generalisation, and implicit regularisation \cite{JMLR:v22:20-410, Martin2021}.
These findings extend the scope of RMT from physical and financial systems to machine learning, reinforcing its role as a unifying framework for understanding the emergent statistical behaviour of high-dimensional matrix ensembles.

Gaussian matrix distributions with $U(N)$ symmetry, and the expectation values of $U(N)$ invariant polynomial functions of matrix variables, have theoretical applications in the context of the AdS/CFT correspondence \cite{Maldacena1999, gkp,witten} which relates the physics of four-dimensional conformal quantum field theory (CFT) to ten-dimensional string theory. 
Highly symmetric quantum states in the CFT can be classified and their interactions computed using representation theoretic structures, notably Young diagrams, which organise the map to gravitons and branes in the string theory \cite{CJR}. 
Representation theory techniques can also be used to organise the combinatorics of invariant polynomial matrix functions when the symmetry group is switched from continuous groups such as $U(N)$ to finite groups such as $S_N$. The general 13-parameter permutation invariant Gaussian matrix models (PIGMMs) were introduced in \cite{Ramgoolam:2018xty}. 
A combination of quantum field theory techniques, notably Wick contraction combinatorics, combined with the representation theory of symmetric groups, was used to compute expectation values of permutation invariant matrix polynomials (also referred to as permutation invariant observables or simply the observables of the model). 
This was motivated by the analysis of matrix data from computational linguistics \cite{Kartsaklis:2017lfq} and extended to data science applications in natural language processing and statistical finance \cite{Ramgoolam:2019ldg, Huber:2022ohf, Barnes:2023kqc}. 
The matrix data analysis based on PIGMM 
shares the philosophy of traditional RMT of characterising universal features across different types of real world matrix ensembles, while using the understanding of the characteristics of this background noise to identify physical system-specific features. 
The technical focus is  on the computation of expectation values of low degree invariant polynomials in matrix variables using techniques from the combinatorics of quantum fields, and the comparison of these to the experimental data of averages of these same invariant polynomials extracted from appropriately defined real world matrix ensembles.  
In this paper we apply the framework of permutation invariant matrix models to ensembles of weight matrices for neural networks. 

This paper starts in §\ref{sec:bkg} with a discussion of the relevant background for Gaussian matrix models, and neural network weight matrices.
Then §\ref{sec:data} introduces the MNIST classification problem trained on, and the weight matrix ensembles generated and studied with the PIGMM formalism.

The main results of the investigations in this work are presented in §\ref{sec:results}, notably including computation of invariants of the permutation symmetry assumed by the models up to order 4.
With these invariants, permutation-invariant Gaussian matrix models (PIGMMs) are fitted to best represent each weight matrix ensemble, and used to predict theoretical values of the higher-order invariants.
Deviation measures are defined which first demonstrate the appropriateness of a simple Gaussian model for the initialised distributions, which are then showed to fit poorly post-training. 
With this motivation for a more general framework for modelling weight distributions throughout training, the PIGMMs are fitted to the lower-order invariants, and deviation measures from their predicted values quantify the appropriateness of these models, establishing them as good fits for the general behaviour across training, initialisation regimes, and architecture layers.
The Wasserstein distance is defined for these PIGMMs and used to measure how the distributions change over training, whilst again also allowing comparison between initialisations and layers at different depths.

Then in §\ref{sec:results2} final investigations text the robustness of the formalism where regularisation is introduced to the training regime, and where the width of the layers grows.
The former is well described by the PIGMM formalism, yet the latter identifies specific deviations from the Gaussianity assumption and suggests how these highly-efficiently parameterised models can be developed to understand the next order of behaviour in the weight distributions.
The final section §\ref{sec:conc} summarises key findings, and outlines future developments; whilst supporting computations are presented in the appendices.

The repository of \texttt{python} code written for this work is available at the respective \href{https://github.com/edhirst/PIGWMM}{\texttt{GitHub}}\footnote{\href{https://github.com/edhirst/PIGWMM}{\texttt{https://github.com/edhirst/PIGWMM}}} repository. It builds on previous work for computing representation theory variables and PIGMM invariants in \cite{Ramgoolam:2019ldg, Huber:2022ohf, PIG2MM}.

%%%%%%%%%%%%%%%%%%%%%%%%%%%%%%%%%%%%%%%%%%%%%%%%%
\section{Background}\label{sec:bkg}
In this section the relevant mathematics behind Gaussian matrix models is introduced, followed by the equivalently relevant essential components of NNs.
The section finishes with a description of permutation symmetry in NN architectures, and how under the Gaussianity assumption these matrix models can be applied to provide efficient and informative representations of trained NN models.

%%%%%%%%%%%%%%%%%%%%%%%%%%%%%%%%%%%%%%%%%%%%%%%%%
\subsection{Permutation Invariant Gaussian Matrix Models}\label{sec:gmms}
The eigenvalue distributions of traditional random matrix theory are derived from Gaussian distributions over matrix variables, which are are invariant under continuous symmetries (see the original work  \cite{Wigner1951, Dyson1962} or the textbook treatment \cite{Mehta2004}). 
For example in the Gaussian orthogonal ensemble (GOE), the matrix variables $\{ M_{ ij}: 1 \le  i , j \le  N\}  $ are real, the matrix is symmetric (equal to its transpose, $M = M^T$).  
The measure is 
\begin{align}\label{eq:GOEMeas}
e^{ - \frac { 1 }{ 2 \sigma^2 }  {\rm tr } M^2  } \prod_{ i \le j } d M_{ ij } 
= e^{ - \frac{ 1 }{2 \sigma^2 }  \sum_{ i =1 } M_{ ii}^2 - \frac{ 1 }{\sigma^2 }  \sum_{ i < j } M_{ ij}^2 } \prod_{ i \le j } d M_{ ij } \;,
\end{align}
where $\prod_{ i\le j } M_{ij}$ is the standard Euclidean  measure on $\frac{N(N-1)}{2}$ variables.
The measure in \eqref{eq:GOEMeas} is invariant under transformations  $ M \rightarrow AM A^T$, where the matrix $A$ is orthogonal, i.e. the transpose is equal to the inverse $A^T = A^{-1}$. 

In the 13-parameter Gaussian matrix model \cite{Ramgoolam:2018xty}, there are $N^2$  real variables $W_{ij}$ and the measure is 
\bea 
e^{ - S ( W ; f_1 , \cdots , f_{ 13} ) } \prod_{ i \le j } d W_{ ij } \;,
\eea
where $S$ is a function of $13$ parameters which are coefficients of linear and quadratic functions of the matrix $W$. W
Note we switch notation from $M$ to $W$ for the purposes of this paper, where the application is the weight matrices. 
Permutations $ \sigma $ are bijective  maps from the set  $ \{ 1, 2, \cdots , N \} $ to itself, equivalently $ \sigma : i \rightarrow \sigma (i) $ define the $N!$ rearrangements of the ordered list of numbers. The permutation matrix $A_{ \sigma } $ acts on a set of  basis vectors $ \{ e_1 , \cdots , e_N \}  $ as 
\bea 
A_{ \sigma }  e_{ i } = e_{ \sigma (i) }\;,
\eea
while $W_{ij} $ are the coefficients of a linear operator 
\bea 
W e_i = W_{ ji} e_j \;.
\eea
With these definitions the transformation 
\bea 
 A_{ \sigma } : W \rightarrow A_{ \sigma } W A^{ -1}_{ \sigma } \;,
\eea 
acts on the matrix elements as 
\bea 
 W_{ij}  \rightarrow W_{ \sigma (i) \sigma(j ) }\;.
\eea 
Every directed graph with $k$ edges defines a permutation invariant polynomial function of degree $k$, by associating indices to the vertices and $W_{ij}$ to edges directed from the vertex labelled $i$ to the vertex labelled $j$ \cite{Kartsaklis:2017lfq, Ramgoolam:2018xty, Barnes:2021ehp, Padellaro:2023blb}. For $N \ge 2k$, these functions are linearly independent. In the data science applications of permutation invariant matrix models, this condition is generally satisfied since the matrices are of size around at least $19$ \cite{Barnes:2023kqc} and often into hundreds \cite{Kartsaklis:2017lfq}, while the degrees of interest have  ranged from $k=1$ to $ k=4$. For $ N < 2k$, the number of linearly independent invariant polynomials  is the number of directed graphs with $k$ edges and $N$ nodes (this is proved in \cite{Barnes:2021ehp} and generalised to 2-colour graphs, while an accessible exposition is in  the proof of proposition 28 of \cite{Padellaro:2023blb}). 

The general permutation invariant Gaussian action for matrices  of size $N$  is constructed by defining an orthogonal transformation, of the form 
\bea 
S^{ \mathcal{A} } = \sum_{ i, j =1 }^D  C^{ \cA}_{ ij } W_{ ij } \;,
\eea
with the orthogonality relation 
\bea\label{orthog} 
\sum_{ i , j } C_{ i  j }^{ \mathcal{A}} C_{ ij}^{ \cB } = \delta^{ \mathcal{A}\cB  } \;.
\eea
The index $ \mathcal{A}$ is an abbreviation for a list of labels $ ( V_{ A } , m_A , \alpha_A  ) $  where $V_A$ is an  irreducible representation of $S_N$, one of  $ \{ V_0 , V_H , V_2 , V_3    \} $, $m_A$ runs over a set of orthonormal basis vectors for $V_A$ while  $ \alpha_A $ is a multiplicity label for $V_A$. Further description of the coefficients 
$C_{ i  j }^{ \mathcal{A}}$ is given in \cite{Ramgoolam:2018xty}. The vector spaces
$\{ V_0 , V_H , V_2 , V_3    \}$ have dimensions
\bea 
 \{ \Dim V_0  , \Dim V_H  , \Dim V_2  , \Dim V_3  \} = \{ 1 , ( D-1) , { D  ( D-3) / 2 } , { ( D-1) ( D-2) /2 } \} \;,
 \eea 
 while the respective multiplicities are $ \{ 2, 3, 1, 1 \} $. 
The probability density function in terms of the representation theoretic variables is given by $ e^{ - \cS } $ with
\begin{equation}\label{actionSvar}
\begin{split}
    \cS = & - \mu_1 S^{  V_0 ;  1 } - \mu_2 S^{ V_0 ; 2} + \frac{1}{2} \sum_{ a, b =1}^2 (\Lambda_{ V_0} )_{ab}S^{ V_0 ; a}S^{ V_0 ; b} +  \frac{1}{2} \sum_{ a, b =1}^{ 3 }  ( \Lambda_{V_H} )_{ ab} \sum_{ m =1}^{ {\rm {Dim }} V_H } S^{V_H ; a }S^{V_H ; a } \\
    & + \frac{1}{2} (\Lambda_{ V_2} ) \sum_{ m=1}^{ {\rm {Dim }} V_2 } S_m^{V_2} S_m^{ V_2} +  \frac{1}{2} \Lambda_{ V_3 } \sum_{ m=1}^{ {\rm {Dim }} V_3  } S_m^{ V_3 } S_m^{ V_3 } \;.
\end{split}
\end{equation}
The matrices $ \Lambda_{ V_0} , \Lambda_{ V_H} $ are symmetric $ 2 \times 2 $ and $ 3 \times  3 $ matrices respectively. 
The orthogonality of the transformation \eqref{orthog} means that the  real Euclidean measure for the $N^2 $ variables $W_{ij}$ is equal to the Euclidean measure in the $S^{ \cA}$ variables. 
This allows  the computation of the expectation values 
of polynomial functions of the matrix variables using Wick's theorem. 
The expectation values 
of linear and quadratic permutation invariant polynomials corresponding to the graphs in Figure \ref{fig:LQ_invariant_graphs}, along with a selection of cubic and quartic invariants were computed in \cite{Ramgoolam:2018xty}. 
The equations for the linear and quadratic invariants were inverted numerically in matrix data analysis problems in the context of computational linguistics in \cite{Ramgoolam:2019ldg, Huber:2022ohf}. 
An analytic inversion computed in this work is given in §\ref{app:param_analytics}. 

Matrix data analysis based on the PIGMM used ensemble averages of the linear and quadratic invariants calculated from matrix datasets to determine the 13 model parameters, by fitting to these averages. 
Analysis has tested permutation invariant  Gaussianity of the ensembles by comparing the theoretical expectation values of cubic and quartic invariants with the ensemble averages of these invariants calculated from the data. 
Strong evidence for approximate permutation invariant Gaussianity in the computational linguistics context was found \cite{Ramgoolam:2019ldg, Huber:2022ohf}. 
The small deviations from Gaussianity were used successfully in an algorithm to distinguish synonyms and antonyms in \cite{Huber:2022ohf}. 
The same matrix data analysis program was applied in the context of financial correlation matrices, by extending the construction of the $13$-parameter models for generic real matrices to $4$-parameter models for symmetric real matrices with vanishing diagonal entries \cite{Barnes:2023kqc}.  
Here we build on these results and initiate the investigation of permutation invariant matrix Gaussianity to weight matrices in neural networks. 

In this paper, our notation for the PIGMM model parameters is 
\begin{equation}
\begin{split}
        (f_1, f_2, ..., f_{13}) = \ & (\tilde{\mu}_1, \tilde{\mu}_2, (\Lambda^{-1}_{V_0})_{11}, (\Lambda^{-1}_{V_0})_{12}, (\Lambda^{-1}_{V_0})_{22}, \\
        & (\Lambda^{-1}_{V_H})_{11}, (\Lambda^{-1}_{V_H})_{12}, (\Lambda^{-1}_{V_H})_{13}, (\Lambda^{-1}_{V_H})_{22}, (\Lambda^{-1}_{V_H})_{23}, (\Lambda^{-1}_{V_H})_{33}, \Lambda^{-1}_{V_2}, \Lambda^{-1}_{V_3}) \;.
\end{split}
\end{equation}

The simplest Gaussian probability density function $ e^{ - \frac{1}{2 \sigma^2} \sum_{ i, j } M_{ ij }^2 }  $ corresponding to drawing the $W_{ij } $ independently from the same Gaussian univariate distribution with standard deviation $\sigma$ is equivalent to the probability density function  specified by \eqref{actionSvar} with the parameters 
\begin{equation}\label{eq:special-param}
\begin{split}
    & f_1 = f_2 = f_4 = f_7  = f_8 = f_{10} = 0 \;, \\
    & f_3  =  f_5 = f_6  =  f_9  = f_{11}  = f_{12}  = f_{13}  = \sigma^{ 2} \;.
\end{split}
\end{equation}
This simple Gaussian is used as one of the standard initialisations in neural networks, and is studied in this work. 
In the case of a Gaussian model matching the variance of an ensemble with $W_{ij}$ independently drawn from a uniform distribution of width $2\sigma$, the equivalent version of \eqref{actionSvar} has parameters
\begin{equation}\label{eq:special-param-uniform}
\begin{split}
    & f_1 = f_2 = f_4 = f_7  = f_8 = f_{10} = 0 \;,\\
    & f_3  =  f_5 = f_6  =  f_9  = f_{11}  = f_{12}  = f_{13}  = \frac{\sigma^2}{3} \;.
\end{split}
\end{equation}
For the uniform initialisation, this will be shown to give a good approximation to the expectation values of permutation invariant linear and quadratic functions of the weight matrices in §\ref{sec:lq} at initialisation. 
This can be understood as due to the central limit theorem, which is at play since the permutation invariants sum over many distinct neurons.  
The departures, both for the simple Gaussian and the uniform initialisation, will be tracked through training, as a way to motivate the more general 13-parameter space Gaussians.

%%%%%%%%%%%%%%%%%%%%%%%%%%%%%%%%%%%%%%%%%%%%%%%%%
\subsection{Neural Networks}
The machine learning subfield of supervised learning encompasses an array of techniques for non-linear function fitting. 
The subfield in general looks to fit maps between input and output data, where data is represented as tensors of varying shape and dimension, and entries are floating point numbers (i.e. decimals up to a set degree of precision).
The architecture of the supervised learning technique specifies the function form as well as the method of fitting it to the data.

Once data has been identified, which for supervised learning will consist of $N$ (input, output) pairs, the standard procedure is to partition the dataset into train:validation:test independent subsets.
The train data is used to perform the function fitting, whilst validation data monitors performance independently throughout the training process; finally the test data is used after training to evaluate the performance of the architecture.
To provide confidence in the test performance measures the same architecture may be reinitialised, trained, and tested on different random partitions of the dataset.
The final performance measures can then be averaged and standard errors calculated, this process is known as cross-validation.

Further subsplitting the supervised learning field, where the output tensors can take value in a continuous range the problem is known as a regression problem. 
Whilst if the outputs take value in some finite set the problem is a classification problem.
Although many classification problems with a large finite set of outputs can be well modelled as regression problems, in the cases where the output set is small the architecture hyperparameters can be tailored to match this problem style.
As generally simpler problems, classification is the focus in this work, where in spirit the architecture is learning to sort the inpust into distinct categories or classes.

Whilst there is an immense selection of techniques one may apply from supervised learning, the perhaps most popular sub-selection are \textit{neural networks}.
Neural networks are functions inspired by the synapse structures in the brain, they are built from constituent neurons, which may be concatenated in a variety of ways in parallel and sequence to build the full architecture.
Each neuron\footnote{Neuron actions in parallel form a layer, and this allows the action of the whole layer to be considered as the activation action on a matrix operation, $\text{act}(W \cdot \underline{x} + \underline{b})$, where $W$ is the stacking of each neuron's weight vector into a matrix, and $\underline{b}$ the stacking of each neuron's bias number into a vector.} takes as input a vector ($\underline{x}$) and outputs a number, the vector is first acted on linearly through means of a weight vector ($\underline{w}$) and bias number ($b$), to produce a number, then non-linearly by an activation function (act$(\cdot)$).
The full neuron functional form is hence: $\underline{x} \longmapsto \text{act}(\underline{w} \cdot \underline{x} + b)$. 
The choice of activation function amounts to the architecture desired, however the most commonly selected due to its simplicity is: $\text{ReLU}(x) \vcentcolon = \max (0,x)$.

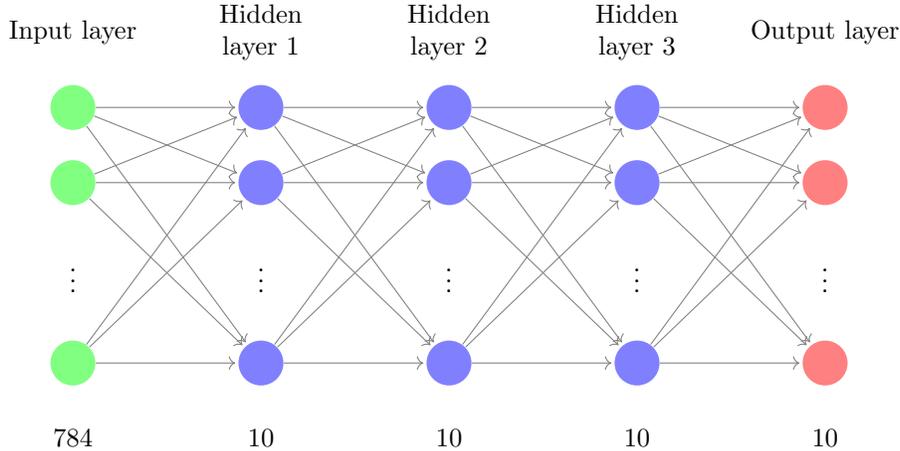
\begin{figure}[!t]
\centering
\begin{comment}
\begin{tikzpicture}[shorten >=1pt,->,draw=black!50, node distance=\layersep]
    \tikzstyle{every pin edge}=[<-,shorten <=1pt]
    \tikzstyle{neuron}=[circle,fill=black!25,minimum size=17pt,inner sep=0pt]
    \tikzstyle{input neuron}=[neuron, fill=green!50];
    \tikzstyle{output neuron}=[neuron, fill=red!50];
    \tikzstyle{hidden neuron}=[neuron, fill=blue!50];
    \tikzstyle{annot} = [text width=4em, text centered]
    \def\layersep{2.5cm}

    % Draw the input layer nodes
    \foreach \name / \y in {1,...,3}
        \node[input neuron, pin=left:Input \#\y] (I-\y) at (0,-\y) {};

    % Draw the hidden layer nodes
    \foreach \name / \y in {1,...,4}
        \path[yshift=0.5cm]
            node[hidden neuron] (H-\name) at (\layersep,-\y cm) {};

    % Draw the output layer nodes
    \foreach \name / \y in {1,...,2}
        \path[yshift=-0.5cm]
            node[output neuron, pin={[pin edge={->}]right:Output \#\y}] (O-\name) at (2*\layersep,-\y cm) {};

    % Connect the input layer nodes with the hidden layer nodes
    \foreach \source in {1,...,3}
        \foreach \dest in {1,...,4}
            \path (I-\source) edge (H-\dest);

    % Connect the hidden layer nodes with the output layer nodes
    \foreach \source in {1,...,4}
        \foreach \dest in {1,...,2}
            \path (H-\source) edge (O-\dest);

    % Annotate the layers
    \node[annot,above of=H-1, node distance=1cm] (hl) {Hidden layer};
    \node[annot,left of=hl] {Input layer};
    \node[annot,right of=hl] {Output layer};
\end{tikzpicture}
\end{comment}
\begin{tikzpicture}[shorten >=1pt,->,draw=black!50, node distance=\layersep]
    \tikzstyle{every pin edge}=[<-,shorten <=1pt]
    \tikzstyle{neuron}=[circle,fill=black!25,minimum size=17pt,inner sep=0pt]
    \tikzstyle{input neuron}=[neuron, fill=green!50];
    \tikzstyle{output neuron}=[neuron, fill=red!50];
    \tikzstyle{hidden neuron}=[neuron, fill=blue!50];
    \tikzstyle{annot} = [text width=6em, text centered]
    \def\layersep{2.5cm}

    % Draw the input layer nodes (3 neurons with ellipses)
    \foreach \y in {1,...,2}
        \node[input neuron] (I-\y) at (0,-\y) {};
    \node at (0,-3.2) {$\vdots$};
    \node[input neuron] (I-3) at (0,-4.4) {};

    % Hidden layers: 3 of them
    \foreach \layer in {1,...,3} {
        \foreach \y in {1,...,2}
            \node[hidden neuron] (H\layer-\y) at (\layer*\layersep,-\y) {};
        \node at (\layer*\layersep,-3.2) {$\vdots$};
        \node[hidden neuron] (H\layer-3) at (\layer*\layersep,-4.4) {};
    }

    % Output layer nodes (3 with ellipses)
    \foreach \y in {1,...,2}
        \node[output neuron] (O-\y) at (4*\layersep,-\y) {};
    \node at (4*\layersep,-3.2) {$\vdots$};
    \node[output neuron] (O-3) at (4*\layersep,-4.4) {};

    % Connect input to first hidden
    \foreach \source in {1,...,3}
        \foreach \dest in {1,...,3}
            \path (I-\source) edge (H1-\dest);

    % Connect hidden layers
    \foreach \layer in {1,...,2}
        \foreach \source in {1,...,3}
            \foreach \dest in {1,...,3}
                \path (H\layer-\source) edge (H\the\numexpr\layer+1\relax-\dest);

    % Connect last hidden to output
    \foreach \source in {1,...,3}
        \foreach \dest in {1,...,3}
            \path (H3-\source) edge (O-\dest);

    % Annotations above layers
    \node[annot,above of=I-1, node distance=1cm] {Input layer};
    \node[annot,above of=H1-1, node distance=1cm] {Hidden layer 1};
    \node[annot,above of=H2-1, node distance=1cm] {Hidden layer 2};
    \node[annot,above of=H3-1, node distance=1cm] {Hidden layer 3};
    \node[annot,above of=O-1, node distance=1cm] {Output layer};

    % Numbers under layers
    \node[below of=I-3, node distance=1cm] {784};
    \node[below of=H1-3, node distance=1cm] {10};
    \node[below of=H2-3, node distance=1cm] {10};
    \node[below of=H3-3, node distance=1cm] {10};
    \node[below of=O-3, node distance=1cm] {10};
\end{tikzpicture}
\caption{A diagrammatic representation of a general neural network with three hidden layers; this matches the main architecture used in this work, where the numbers below indicate the number of neurons in each layer in the used architecture. Each neuron represents the action of a linear and then a non-linear function on its input vector, the linear action in this work is multiplication by a weight matrix (no biases) represented by the arrows in the diagram \cite{Armstrong-Williams:2024nzy}. The graphical nature between layers is that of a complete bipartite graph.}
\label{fig:NN_diagram}
\end{figure}

The popularity of neural networks may be attributed to their vast adaptability, ease of implementation, and surprisingly strong performances in a broad range of problems and fields.
Although the design freedom for arranging neurons allows for more complicated architectures, including convolutional neural networks (famous for many image data applications) \cite{oshea2015introcnns} and recurrent neural networks \cite{schmidt_2019} (the predecessor to the attention mechanism \cite{NIPS2017_3f5ee243} in transformers and large language models), the most commonly used are the prototypical dense feed-forward neural networks \cite{anderson1995introduction}.
An abstract representation of the standard dense neural network design is shown in Figure \ref{fig:NN_diagram}, in these architectures neurons are organised into layers, where each neuron in a layer receives the same vector input, and each output number of that layer is concatenated into a vector to become the next layer's input.
The number of neurons may vary between layers, where the first layer receives the input vector and the last layer has as many neurons as the output vector; where there is more than one layer the network is described as deep.

A general dense neural network hence has many parameters (all the weights and biases for each neuron), which makes the parameter space very high-dimensional.
Further to a choice of neural network hyperparameters, an architecture requires a loss function to compare predicted outputs to true outputs.
For the classification tasks considered in this work the standard choice for this is categorical cross-entropy; this first acts on the outputs with a softmax layer which normalises the output into a discrete probability distribution over the classes, then computing the cross-entropy between the output probability distribution and the true class distribution (a one-hot encoded class label, i.e. vector of all zeros except a single one at the entry corresponding to the true class).
For network output $\underline{y}$, and true data output $\underline{\hat{y}}$ this looks like:
\begin{equation}
    \text{CEL} \vcentcolon = \frac{1}{\tilde{N}} \sum_n^{\tilde{N}} \sum_c^C \hat{y}_{n,c} \text{log} \bigg( \frac{\text{exp}(y_{n,c})}{\sum_{\tilde{c}}^C \text{exp}(y_{n,\tilde{c}})} \bigg)\;,
\end{equation}
where $c$ and $\tilde{c}$ run over the $C$ classes (for the problem considered here $C=10$), and $n$ runs over the outputs in the batch of size $\tilde{N}$.

Throughout the training process, a selected optimiser (most often Adam \cite{kingma2017adam}), will update the network parameters (weights and biases) to minimise the loss over some batch of data (a subset of the training data) in some gradient descent inspired manner, with step-size controlled by the learning rate parameter.
The updates are repeated for all batches across the training data, and this single run through of the full training set is called an epoch.
At this point the validation data may be used to assess intermediate performance.
The training process is then repeated for many epochs, where the data in each batch is shuffled as each new epoch starts.

Throughout training additional metrics to the loss function may be evaluated and tracked, as well as further used for the testing procedure.
For the classification problems of focus in this work, general performance metrics are usually functions of the confusion matrix $(CM)_{ij}$ which counts the number of test data inputs in class $i$ that the trained network classifies into class $j$.
This matrix can then be normalised, and the most typical metric used is then the accuracy, defined:
\begin{equation}
    \text{Accuracy} \vcentcolon = \quad \sum_i (CM)_{ii} \quad \in [0,1]\;,
\end{equation}
which may be interpreted as the proportion of correctly classified data inputs, such that 1 indicates perfect learning.

%%%%%%%%%%%%%%%%%%%%%%%%%%%%%%%%%%%%%%%%%%%%%%%%%
\subsubsection{Datasets}\label{sec:mnist}
In examining the Gaussianity of neural network parameters the prototypical MNIST classification task is selected for focus.
The database is made directly available in the two standard \texttt{python} machine learning libraries: \texttt{tensorflow} \cite{tensorflow2015-whitepaper} and \texttt{pytorch} \cite{NEURIPS2019_9015}; of which the latter is used in this work.

\begin{figure}[tb]
    \centering
    \includegraphics[width=0.8\textwidth]{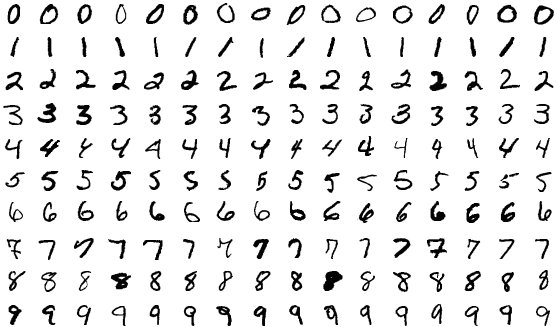}
    \caption{Sample images from the MNIST database of handwritten digits. Each row shows examples from the database of the digits $0 - 9$.}
    \label{fig:dataset_sampleimages}
\end{figure}

The MNIST \cite{LeCun2005TheMD,deng2012mnist} database consists of images of handwritten digits, it amounts to 70000 images of the digits $0-9$ (i.e. 7000 of each), each paired with the respective digit as the classification label, and the full dataset is split such that 60000 images are for training and 10000 for testing.
The images have $28 \times 28$ pixels, and are greyscaled.
A sample of these images is shown in Figure \ref{fig:dataset_sampleimages}.
This database is probably the most standard database used for introducing and bench-marking machine learning processes, as demonstrated in the works \cite{mnist_comparisons,mnist_paperswithcode}, where performances are shown to reach as high as 0.9987 accuracy.

%%%%%%%%%%%%%%%%%%%%%%%%%%%%%%%%%%%%%%%%%%%%%%%%%
\subsection{Symmetry in Neural Networks and PIGMMs}\label{sec:perm_intro}
Permutation symmetry of a system amounts to
an invariance of the appropriately defined observables of the system under a permutation of the variables defining the system. The collection of all $n!$ permutations of $n$ variables 
$\{a_1, a_2, ..., a_n\}$ forms the symmetric group $S_n$. Each permutation $ \pi \in S_n $ is a bijective map from  the set $\{ 1 , 2 , \cdots , n \} $ to  the same set, which takes $ i \rightarrow \pi (i) $. The variables $ \{ a_1 , \cdots , a_n \} $ are mapped to
$ \{ a_{ \pi (1)} , \cdots , a_{ \pi (n) } \} $. 
The symmetric groups  are a fundamental  family of finite symmetry groups and every finite group is a  subgroup of a symmetric group by Cayley's theorem.  In this paper, we investigate the applicability of permutation invariant matrix distributions 
to the statistics of weight matrices of neural networks throughout the training process, from initialisation to the learned matrices.  Learned matrices arising at the end of training belong to ensembles whose statistical characteristics are brought into focus, much as the foundational works of random matrix theory sought the universal patterns of simplicity described by traditional matrix models in the  complex energy spectra of highly excited nuclei \cite{Wigner1951, Dyson1962}. The convergence of known permutation symmetries of neural network architectures and recent developments in permutation invariant matrix models make this a particularly timely project, and we will outline further avenues of investigation in the conclusions. 

When discussing symmetries of dense feed-forward neural networks, it is useful to distinguish different levels  of structure. 
At the level of connectivity, each dense layer corresponds to a complete bipartite graph, whose adjacency matrix is the all-ones matrix. 
This structure carries a large continuous symmetry group: if we regard the weights of a layer as a vector in 
$\mathbb{R}^{n_1 n_2}$, then the standard Gaussian initialisation corresponds to the quadratic form 
$\exp(-\tfrac{1}{2\sigma^2}\sum_{A=1}^{n_1 n_2} w_A^2)$, which is invariant under the full orthogonal group $O(n_1 n_2)$. 
Taking into account the bipartite matrix structure of the weights breaks this to $O(n_2)\times O(n_1)$ (as nodes are not moved between layers), acting by $W \mapsto U W V^\top$ 
with $U\in O(n_2)$ and $V\in O(n_1)$. 
Finally, once non-linear activations are included, the continuous orthogonal invariance is reduced further, 
to discrete permutation symmetries of the hidden units. 
This can be imagined as activation on a set of outputs is preserved as activation is applied element-wise (so the permutation reordering does not change the set of outputs), but most linear combinations of outputs will change the set and what is outputted, for example allowing a linear combination which changes an output from positive to negative and changing dramatically how a ReLU activation acts on it. 
In particular, it has been shown in \cite{navon2023equiv} that the parameter space of a dense feed-forward neural network is invariant under 
the product of symmetric groups
\bea\label{prodAI} 
S_{d_1}\times\cdots\times S_{d_{k-1}} \, , 
\eea 
where $d_i$ is the number of nodes in layer $i$ for the $k$ layers of the NN.
This inter-layer symmetry corresponds to permutations of hidden units in each layer, acting between consecutive layers and is natural to any dense NN function.
For layer $i$ with weight matrix of shape $(d_\alpha, d_\beta)$ and for layer $(i+1)$ with shape $(d_\gamma, d_\alpha)$, then the action of $S_{d_\alpha}$ simultaneously on the weight matrices either side (just the rows on the $i$ layer and just the columns on the $(i+1)$ layer) has no effect on the output and is a symmetry of the function.

In convolutional neural networks (CNNs), the imposition of locality together with weight sharing breaks the permutation symmetries present in dense feed-forward networks, and there're similar restrictions with recursive neural networks (RNNs) and other architectures, so the resulting parameter redundancies take a different form. 
Hence the dense feed forward neural networks provide the natural setting to investigate the applicability of permutation invariant matrix statistics. 

Following the idea of exploring statistical regularities in the weight matrices arising at the end of and through training, in light of the symmetries discussed above, an additional ingredient in our approach is the central limit theorem. The Central Limit Theorem (CLT) asserts that when many independent random contributions are aggregated, the total tends toward a Gaussian law.
In its multivariate form, the result extends to random vectors: sums of many independent vector-valued contributions converge in distribution to a multivariate Gaussian, with a covariance matrix determined by the variances and correlations of the underlying variables \cite{billingsley1995probability,anderson2003multivariate}.
Whereas analyses based on the Neural Network Gaussian Process (NNGP) framework \cite{neal1996bayesian,matthews2018gaussian,lee2018deep} and the Neural Tangent Kernel (NTK) \cite{jacot2018neural,lee2019wide} focus on Gaussianity in the distribution of network outputs and training dynamics in function space, here we turn to the empirical investigation of the approximate Gaussianity that emerges in the weight matrices themselves during training, providing a parameter-space analogue. 
Alongside the motivation of Gaussianity outlined, will use the discussion of symmetry above as an avenue to choose a convenient set of existing matrix models  as a tool for data-reduction. 

The standard Gaussian initialisation draws the $d^2$ matrix entries of the weight matrices from a single Gaussian distribution which has a mean and a standard deviation, thus two parameters.  
The general multi-variate Gaussian for $d^2$ variables has $d^2$ means, $\mu_{ij}$, and $ d^2  ( d^2 + 1 ) /2 $ parameters in the quadratic terms, $\mathcal{T}_{ij;kl}$, taking the form: 
\bea 
 e^{ + \sum_{ i,j} \mu_{ ij} W_{ ij}  - \sum_{ i,j , k , l } \mathcal { T}_{ ij ; kl } W_{ ij } W_{ kl } } \prod_{ ij } dW_{ ij} \;.
\eea
This will generally provide a vast set of parameters, not conducive to an interpretable framework. 
Simplifying to an architecture with fixed layer size ($d_{i-1}=d_i$), and constraining the linear and quadratic terms to be permutation invariant, in accordance with an assumption of a diagonal subsymmetry $\text{diag}(S_{d_{i}} \times S_{d_i})$, leads to the 13-parameter models of \cite{Ramgoolam:2018xty}.
These are amenable to computations of expectation values of invariant functions using representation theory of $S_d$ and Wick's theorem, described at the start of this section.

%%%%%%%%%%%%%%%%%%%%%%%%%%%%%%%%%%%%%%%%%%%%%%%%%
\section{Data Generation}\label{sec:data}
The matrices to be modelled in this work are neural network weight matrices, which were collected across the network architecture layers throughout the training process on the MNIST classification problem, as described in §\ref{sec:mnist}.

The prototypical dense feed-forward neural network architecture was used, with layer sizes chosen to produce square weight matrices of size 10 to reflect the 10 output classes in MNIST problem considered.
The networks had 3 hidden layers of 10 neurons each, producing weight matrix data\footnote{The first weight matrix $W^0_{d,10}$ was omitted in analysis due to its non-square nature, where square is desired for this first analysis assuming the diagonal subgroup permutation symmetry.}: $\{W^0_{d,10},W^1_{10,10},W^2_{10,10},W^3_{10,10}\}$; where each weight matrix is the concatenation of all neuron weight vectors ($\underline{w}$) in the layer, for all layers between the input layer of size $d$ (784 for MNIST) and output layer of size $10$.
No biases were used in the networks to maintain focus on the weight matrix parameters.
All the layers (except the output layer) used the standard ReLU activation, as motivated by \cite{pedamonti2018comparison}. 

For consistency in comparison, the same remaining architecture hyperparameters were used throughout, including a cross-entropy loss, Adam optimiser (with learning rate 0.01), training for 50 epochs using batches of size 100.
The same architecture was reinitialised and retrained 1000 times to produce the weight databases used for statistical analysis and modelling.

At initialisation the parameters of the network are drawn from a given distribution.
Across neural network literature there are two standard choices for this distribution, a Gaussian distribution and a uniform distribution.
Since the aim of this work is to examine the Gaussianity of the weights throughout the training process, this choice of scheme becomes immediately relevant.
Therefore to probe the effect of initialisation scheme on the subsequent matrix distributions the data generation is repeated for both initialisation schemes.
The Gaussian initialisation scheme draws initial weight values from the distribution
\begin{align}\label{eq:gauss_init}
    f(w) = N(\mu, \sigma) & = \frac{1}{\sigma \sqrt{2\pi}}\text{exp}\bigg( -\frac{1}{2}\bigg( \frac{w-\mu}{\sigma} \bigg)^2\bigg)\;,\\ 
    & = \sqrt{\frac{d_{\text{in}}}{2\pi}}\text{exp}\bigg( -\frac{d_{\text{in}}w^2}{2} \bigg) \;,
\end{align}
for mean $\mu=0$, and standard deviation $\sigma = \frac{1}{\sqrt{d_{\text{in}}}}$ where $d_{\text{in}}$ is the dimension of that layer's input ($d_{\text{in}}=10$ for all the layers we study, $W^{i>0}_{10,10}$).
Conversely, the \texttt{pytorch} standard uniform initialisation scheme draws initial weight values from the distribution
\begin{align}\label{eq:uniform_init}
    f(w) = U(a,b) & = \begin{cases} \frac{1}{b-a}\;, & \text{if } a\leq w \leq b\\ 0,              & \text{otherwise} \end{cases}\;,\\ %note linear layers used have no 'gain'
    & = \begin{cases} \frac{\sqrt{d_{\text{in}}}}{2}\;, & \text{if } a\leq w \leq b\\ 0,              & \text{otherwise} \end{cases}\;,
\end{align}
for lower bound $a=-\frac{1}{\sqrt{d_{\text{in}}}}$ and higher bound $b=\frac{1}{\sqrt{d_{\text{in}}}}$, and again layer input dimension $d_{\text{in}}$, as motivated by the method \cite{He_uniforminit}.

The final databases of weight matrices may hence be indexed: $W^{i,l,r}$ for initialisation scheme $i \in$ \{Gaussian, Uniform\}, layer $l \in \{1,2,3\}$, and experiment run $r \in \{1,2,...,1000\}$; where the first 2 indices are selected for an investigation, leaving the $r$ index to denote the database of weight matrices statistically analysed in each case.

The machine learning for weight matrix data generation, and subsequent analysis was coded in \texttt{python}, with use of the \texttt{pytorch} library \cite{NEURIPS2019_9015}.
Scripts and data are made available at this works' respective \href{https://github.com/edhirst/PIGWMM}{\texttt{GitHub}}.

\begin{table}[!t]
\centering
\begin{tabular}{|c|c|c|}
\hline
Problem                & Initialisation & Average Accuracy    \\ \hline
\multirow{2}{*}{MNIST} & Gaussian       & $0.8909 \pm 0.0023$ \\ \cline{2-3} 
                       & Uniform        & $0.9005 \pm 0.0017$ \\ \hline
\end{tabular}
\caption{Final accuracies of the trained NN models after the 50 epochs of training for the MNIST classification investigation, averaged over the 1000 runs, reported with standard error.} 
\label{tab:accuracies}
%MG 0.89089 \pm 0.00229169498406747, MU 0.900499 \pm 0.0017421917228020575
\end{table}

The trained NN models had average final accuracies as displayed in Table \ref{tab:accuracies}. 
The performances are notably similar between initialisations, suggesting that 50 epochs is sufficiently long a training time to move away from the sampled positions in parameter space towards optimum positions.
Since this classification problem has 10 classes, accuracies of 0.1 represent null learning. 
The usefulness of models with accuracies above this 0.1 score is then open to interpretation and is heavily dependent on the problem context and perceived difficulty.
Practically, applications of NNs in industry typically look for accuracy scores exceeding 0.9, a benchmark achieved in this MNIST problem.
However this choice is somewhat arbitrary, and dependent on a problem's difficulty. %An attribute quantifiably measured with information-theoretic techniques in §\ref{sec:entropy}.

%%%%%%%%%%%%%%%%%%%%%%%%%%%%%%%%%%%%%%%%%%%%%%%%%
\subsection{Analytic Considerations}
The PIGMMs use the set of linear and quadratic matrix invariants of the imposed diagonal permutation symmetry to fit their model parameters, as described in §\ref{sec:gmms}.

At initialisation the weights are sampled using one of the quoted initialisation distributions, as described in \eqref{eq:gauss_init} \& \eqref{eq:uniform_init}.
To analyse the PIGMM fitting, it is important to know expected values of these invariants for samples from these distributions, as well as some measure of variation such that deviation from the initialisation distributions can be measured throughout training.

Calculation of these invariant and model parameter expectation values and standard error / deviations requires subtle statistical rigour, and was completed for each of the 13 linear and quadratic invariants and the 13 model parameters.
A thorough introduction to the statistical techniques required as well as the full calculations are available in §\ref{app:inv_analytics} for the invariants and §\ref{app:param_analytics} for the PIGMM parameters respectively.

The outcome of the calculations are expectation values and standard errors for each invariant, calculated generally in the appendix, but shown in Table \ref{tab:inv_analytic} with the specific substitutions to give numerical values for this MNIST classification problem with the implemented architecture.
Equivalently the expectation values and standard deviations\footnote{Whereas invariants are computed for each weight matrix in a run and then averaged such that standard error is the appropriate variation measure of the invariant mean values, the model parameters are computed once from these invariant means such that standard deviation is the appropriate variation measure.} are shown for the model parameters in Table \ref{tab:param_analytic}.
Values are shown for both initialisation distributions of $10 \times 10$ weight matrices, over the 1000 runs generated, noting a match to values quoted in \eqref{eq:special-param} \& \eqref{eq:special-param-uniform}.

The values in both cases have varying scales, with some notably having expectations of zero. More importantly perhaps are the diverse variation measure values, particularly in the model parameter case, which will be essential for assessing the fitting of these PIGMMs in the subsequent sections.

\begin{table}[!t]
\centering
\addtolength{\leftskip}{-4cm}
\addtolength{\rightskip}{-4cm}
\begin{tabular}{|cc|cc|ccccccccccc|}
\hline
\multicolumn{2}{|c|}{\multirow{2}{*}{Invariant}}  & \multicolumn{2}{c|}{Linear}        & \multicolumn{11}{c|}{Quadratic}  \\ \cline{3-15} 
\multicolumn{2}{|c|}{}                                                                                        & \multicolumn{1}{c|}{$I_1$} & $I_2$ & \multicolumn{1}{c|}{$I_3$} & \multicolumn{1}{c|}{$I_4$} & \multicolumn{1}{c|}{$I_5$} & \multicolumn{1}{c|}{$I_6$} & \multicolumn{1}{c|}{$I_7$} & \multicolumn{1}{c|}{$I_8$} & \multicolumn{1}{c|}{$I_9$} & \multicolumn{1}{c|}{$I_{10}$} & \multicolumn{1}{c|}{$I_{11}$} & \multicolumn{1}{c|}{$I_{12}$} & $I_{13}$ \\ \hline
\multicolumn{2}{|c|}{Equation} & \multicolumn{1}{c|}{\tiny{$W_{ii}$}}      &    \tiny{$W_{ij}$}   & \multicolumn{1}{c|}{\tiny{$W_{ij}^2$}}      & \multicolumn{1}{c|}{\tiny{$W_{ij}W_{ji}$}}      & \multicolumn{1}{c|}{\tiny{$W_{ii}W_{ij}$}}      & \multicolumn{1}{c|}{\tiny{$W_{ii}W_{ji}$}}      & \multicolumn{1}{c|}{\tiny{$W_{ij}W_{ik}$}}      & \multicolumn{1}{c|}{\tiny{$W_{ij}W_{kj}$}}      & \multicolumn{1}{c|}{\tiny{$W_{ij}W_{jk}$}}      & \multicolumn{1}{c|}{\tiny{$W_{ij}W_{kl}$}}       & \multicolumn{1}{c|}{\tiny{$W_{ii}^2$}}       & \multicolumn{1}{c|}{\tiny{$W_{ii}W_{jj}$}}       &    \tiny{$W_{ii}W_{jk}$}    \\ \hline
\multicolumn{1}{|c|}{\multirow{2}{*}{Gaussian}} & $\langle \cdot \rangle$ & \multicolumn{1}{c|}{\tiny{00[0]}}      &    \tiny{00[0]}   & \multicolumn{1}{c|}{\tiny{10[0]}}      & \multicolumn{1}{c|}{\tiny{10[-1]}}      & \multicolumn{1}{c|}{\tiny{10[-1]}}      & \multicolumn{1}{c|}{\tiny{10[-1]}}      & \multicolumn{1}{c|}{\tiny{10[0]}}      & \multicolumn{1}{c|}{\tiny{10[0]}}      & \multicolumn{1}{c|}{\tiny{10[-1]}}      & \multicolumn{1}{c|}{\tiny{10[0]}}       & \multicolumn{1}{c|}{\tiny{10[-1]}}       & \multicolumn{1}{c|}{\tiny{10[-1]}}       &    \tiny{10[-1]}     \\ \cline{2-15}   
\multicolumn{1}{|c|}{}                                                                             &  $\hat{\sigma}_{SE}$ & \multicolumn{1}{c|}{\tiny{32[-3]}}      &    \tiny{10[-2]}   & \multicolumn{1}{c|}{\tiny{45[-3]}}      & \multicolumn{1}{c|}{\tiny{33[-3]}}      & \multicolumn{1}{c|}{\tiny{33[-3]}}      & \multicolumn{1}{c|}{\tiny{33[-3]}}      & \multicolumn{1}{c|}{\tiny{10[-2]}}      & \multicolumn{1}{c|}{\tiny{10[-2]}}      & \multicolumn{1}{c|}{\tiny{10[-2]}}      & \multicolumn{1}{c|}{\tiny{32[-2]}}       & \multicolumn{1}{c|}{\tiny{14[-3]}}       & \multicolumn{1}{c|}{\tiny{33[-3]}}       &    \tiny{10[-2]}   \\ \hline
\multicolumn{1}{|c|}{\multirow{2}{*}{Uniform}} & $\langle \cdot \rangle$  & \multicolumn{1}{c|}{\tiny{00[0]}}      &    \tiny{00[0]}   & \multicolumn{1}{c|}{\tiny{33[-1]}}      & \multicolumn{1}{c|}{\tiny{33[-2]}}      & \multicolumn{1}{c|}{\tiny{33[-2]}}      & \multicolumn{1}{c|}{\tiny{33[-2]}}      & \multicolumn{1}{c|}{\tiny{33[-1]}}      & \multicolumn{1}{c|}{\tiny{33[-1]}}      & \multicolumn{1}{c|}{\tiny{33[-2]}}      & \multicolumn{1}{c|}{\tiny{33[-1]}}       & \multicolumn{1}{c|}{\tiny{33[-2]}}       & \multicolumn{1}{c|}{\tiny{33[-2]}}       &    \tiny{33[-2]}     \\ \cline{2-15} 
\multicolumn{1}{|c|}{}    & $\hat{\sigma}_{SE}$  & \multicolumn{1}{c|}{\tiny{18[-3]}}      &    \tiny{58[-3]}   & \multicolumn{1}{c|}{\tiny{94[-4]}}      & \multicolumn{1}{c|}{\tiny{10[-3]}}      & \multicolumn{1}{c|}{\tiny{10[-3]}}      & \multicolumn{1}{c|}{\tiny{10[-3]}}      & \multicolumn{1}{c|}{\tiny{33[-3]}}      & \multicolumn{1}{c|}{\tiny{33[-3]}}      & \multicolumn{1}{c|}{\tiny{33[-3]}}      & \multicolumn{1}{c|}{\tiny{11[-2]}}       & \multicolumn{1}{c|}{\tiny{30[-4]}}       & \multicolumn{1}{c|}{\tiny{10[-3]}}       &    \tiny{33[-3]}    \\ \hline
\end{tabular}
\caption{Numerical expected values ($\langle \cdot \rangle$) and standard errors ($\hat{\sigma}_{SE}$) of the linear and quadratic invariants ($I_i$) at initialisation in both the Gaussian and Uniform schemes. The table shows the invariant equations with respect to the weight matrix entries \tiny $W_{ij}$ \footnotesize, where summing over the indices (\tiny $i,j,k,l$ \footnotesize) is implicit. Values are shown to two significant figures with notation $a[b] = a \times 10^b$.}
\label{tab:inv_analytic}
\end{table}

\begin{table}[!t]
\centering
\addtolength{\leftskip}{-4cm}
\addtolength{\rightskip}{-4cm}
\begin{tabular}{|cc|cc|ccccccccccc|}
\hline
\multicolumn{2}{|c|}{\multirow{2}{*}{Model Parameter}}  & \multicolumn{2}{c|}{Linear}        & \multicolumn{11}{c|}{Quadratic}  \\ \cline{3-15} 
\multicolumn{2}{|c|}{}  & \multicolumn{1}{c|}{$f_1$} & $f_2$ & \multicolumn{1}{c|}{$f_3$} & \multicolumn{1}{c|}{$f_4$} & \multicolumn{1}{c|}{$f_5$} & \multicolumn{1}{c|}{$f_6$} & \multicolumn{1}{c|}{$f_7$} & \multicolumn{1}{c|}{$f_8$} & \multicolumn{1}{c|}{$f_9$} & \multicolumn{1}{c|}{$f_{10}$} & \multicolumn{1}{c|}{$f_{11}$} & \multicolumn{1}{c|}{$f_{12}$} & $f_{13}$ \\ \hline
\multicolumn{1}{|c|}{\multirow{2}{*}{Gaussian}} & $\langle \cdot \rangle$ & \multicolumn{1}{c|}{\tiny{00[0]}}      &    \tiny{00[0]}   & \multicolumn{1}{c|}{\tiny{10[-2]}}      & \multicolumn{1}{c|}{\tiny{00[0]}}      & \multicolumn{1}{c|}{\tiny{10[-2]}}      & \multicolumn{1}{c|}{\tiny{10[-2]}}      & \multicolumn{1}{c|}{\tiny{00[0]}}      & \multicolumn{1}{c|}{\tiny{00[0]}}      & \multicolumn{1}{c|}{\tiny{10[-2]}}      & \multicolumn{1}{c|}{\tiny{00[0]}}       & \multicolumn{1}{c|}{\tiny{10[-2]}}       & \multicolumn{1}{c|}{\tiny{10[-2]}}       &    \tiny{10[-2]}     \\ \cline{2-15}   
\multicolumn{1}{|c|}{} &  $\hat{\sigma}$ & \multicolumn{1}{c|}{\tiny{10[-3]}}      &    \tiny{11[-3]}   & \multicolumn{1}{c|}{\tiny{32[-4]}}      & \multicolumn{1}{c|}{\tiny{35[-4]}}      & \multicolumn{1}{c|}{\tiny{46[-4]}}      & \multicolumn{1}{c|}{\tiny{12[-4]}}      & \multicolumn{1}{c|}{\tiny{12[-4]}}      & \multicolumn{1}{c|}{\tiny{15[-4]}}      & \multicolumn{1}{c|}{\tiny{12[-4]}}      & \multicolumn{1}{c|}{\tiny{15[-4]}}       & \multicolumn{1}{c|}{\tiny{25[-4]}}       & \multicolumn{1}{c|}{\tiny{11[-4]}}       &    \tiny{81[-5]}   \\ \hline
\multicolumn{1}{|c|}{\multirow{2}{*}{Uniform}} & $\langle \cdot \rangle$  & \multicolumn{1}{c|}{\tiny{00[0]}}      &    \tiny{00[0]}   & \multicolumn{1}{c|}{\tiny{33[-3]}}      & \multicolumn{1}{c|}{\tiny{00[0]}}      & \multicolumn{1}{c|}{\tiny{33[-3]}}      & \multicolumn{1}{c|}{\tiny{33[-3]}}      & \multicolumn{1}{c|}{\tiny{00[0]}}      & \multicolumn{1}{c|}{\tiny{00[0]}}      & \multicolumn{1}{c|}{\tiny{33[-3]}}      & \multicolumn{1}{c|}{\tiny{00[0]}}       & \multicolumn{1}{c|}{\tiny{33[-3]}}       & \multicolumn{1}{c|}{\tiny{33[-3]}}       &    \tiny{33[-3]}     \\ \cline{2-15} 
\multicolumn{1}{|c|}{}    & $\hat{\sigma}$  & \multicolumn{1}{c|}{\tiny{58[-4]}}      &    \tiny{64[-4]}   & \multicolumn{1}{c|}{\tiny{11[-4]}}      & \multicolumn{1}{c|}{\tiny{12[-4]}}      & \multicolumn{1}{c|}{\tiny{15[-4]}}      & \multicolumn{1}{c|}{\tiny{38[-5]}}      & \multicolumn{1}{c|}{\tiny{39[-5]}}      & \multicolumn{1}{c|}{\tiny{48[-5]}}      & \multicolumn{1}{c|}{\tiny{38[-5]}}      & \multicolumn{1}{c|}{\tiny{48[-5]}}       & \multicolumn{1}{c|}{\tiny{64[-5]}}       & \multicolumn{1}{c|}{\tiny{29[-5]}}       &    \tiny{21[-5]}    \\ \hline
\end{tabular}
\caption{Numerical expected values ($\langle \cdot \rangle$) and standard deviations ($\hat{\sigma}$) of the linear and quadratic model parameters ($f_i$) at initialisation in both the Gaussian and Uniform schemes. Values are shown to two significant figures with notation $a[b] = a \times 10^b$.}
\label{tab:param_analytic}
\end{table}

%%%%%%%%%%%%%%%%%%%%%%%%%%%%%%%%%%%%%%%%%%%%%%%%%
\section{Approximate Gaussianity in Weight Matrices}\label{sec:results}
With 306 ensembles of 1000 matrices, from the 2 initialisations, 3 layers, and 51 epochs (including an `epoch 0' for before training), over the 1000 runs, there is now plenty of data to apply the PIGMM formalism to.
Having ensembles for multiple initialisation distributions and layers allows further probing of the underlying weight matrix data symmetries and correlations via the testing of these PIGMM models which rely on the diagonal permutation symmetry and the Gaussianity assumptions.

In this section, results from the computation of the relevant invariants, and fitting of the equivalent PIGMM models are reported.
Results are organised first into 2 sections based on the order of the invariants considered: either §\ref{sec:lq} for linear \& quadratic, or §\ref{sec:cq} for cubic \& quartic; then there is a section on formal model parameter space distances in §\ref{sec:wasserstein}.

The first section's results in §\ref{sec:lq} include the computation of the 13 linear and quadratic invariants for each ensemble and the equivalent best fit 13 PIGMM model parameters, a deviation measure is defined and used to quantify the change of the more general PIGMM from the initialisation distributions, with subsections dedicated to before, after, and throughout training.
The second section's results in §\ref{sec:cq} equivalently compute the 39 cubic and quartic invariants from the ensembles, and independently predict them from the fitted PIGMMs, another deviation measure is defined which is used to quantify the extent to which the PIGMM generalised and is a good fit to the weight matrix data - providing insight into the appropriateness of the Gaussianity and diagonal permutation symmetry assumptions, again separated according to after, or throughout training.

The third section's results in §\ref{sec:wasserstein} define the Wasserstein distance for PIGMMs (a novel result in itself), and apply it as another measure for quantifying the change in the best fitted PIGMM over the training as it changes from the initialisation distribution.

%%%%%%%%%%%%%%%%%%%%%%%%%%%%%%%%%%%%%%%%%%%%%%%%%
\subsection{Linear \& Quadratic Invariants}\label{sec:lq}
In this section the 13 linear and quadratic invariants of the diagonal permutation symmetry are computed for the 306 matrix ensembles. 
Which are then used to compute the equivalent 13 model parameters in fitting the PIGMMs to each ensemble of weights. 

To quantify the departure of the fitted PIGMM from the simple distribution at initialisation (simple Gaussian or Gaussian approximation to the uniform distribution) we define a \textit{deviation} measure for the ensemble expectation values of  linear and quadratic permutation  invariants at any training epoch from the theoretical expectation values arising from the initialisation distributions, 
\begin{equation}\label{eq:LQinv_deviation} \text{Deviation}_{LQ}(I_i) : = \frac{|\hat{I}_i - \langle I_i \rangle|}{\hat{\sigma}_{SE}(I_i)}\;,
\end{equation}
where $\hat I_i$ is the ensemble average at a specified training epoch, whilst $\langle I_i \rangle$ and $\hat{\sigma}_{SE}(I_i)$ are the analytically computed expectation values and standard errors for these initialisation distributions by application of Wick's theorem in §\ref{app:inv_analytics}, as displayed in Table \ref{tab:inv_analytic}. 
This measure dictates how many standard errors the mean invariant value is away from its expectation value\footnote{We emphasise also that in computing the deviations the true values of each of the inputs were used, not the versions rounded to 2 significant figures in the tables.}.

Then propagating this comparison to the level of the PIGMM parameters, another deviation measure for the parameter values is then defined
\begin{equation}\label{eq:param_deviation}
    \text{Deviation}_{LQ}(f_i)  := \frac{|f_i - \langle f_i \rangle|}{\hat{\sigma}(f_i)}\;,
\end{equation}
where $f_i$ are the observed model parameters computed using the NN weight data, as shown in Table \ref{tab:pre-modelparams}, whilst $\langle f_i \rangle$ and $\hat{\sigma}(f_i)$ are the analytically computed expectation values and standard deviations for these model parameters when fitted to these initialisation distributions, as computed in §\ref{app:param_analytics} and displayed in Table \ref{tab:param_analytic}, using the functional forms of the model parameters in terms of the expectation values of linear/quadratic invariants. 
%Equivalently, this measure dictates how many standard deviations the observed model parameter value is away from its expectation value. 

These analytic values used in the deviation measures represent the `simple Gaussian' as a special case of the PIGMMs and the deviations measure how far the observed distribution for each ensemble and best fitted general PIGMM model is from this simple Gaussian initial assumption associated to the initialisation distributions.
These deviation measures are now used to test this assumption and the PIGMMs in this prototypical learning scenario at each stage of training. 
The observations on the extent of departures from the simple Gaussians based on the deviation measures \eqref{eq:LQinv_deviation} \& \eqref{eq:param_deviation} will be later corroborated in §\ref{sec:wasserstein} using the Wasserstein distance specialised to these PIGMMs. 

%%%%%%%%%%%%%%%%%%%%%%%%%%%%%%%%%%%%%%%%%%%%%%%%%
\subsubsection{Before Training}\label{sec:lq_before}
At initialisation (epoch 0) there're 6 weight matrix ensembles to consider for the 2 initialisation distributions and the 3 layers. 

%%%%%%%%%%%%%%%%%%%%%%%%%%%%%%%%%%%%%%%%%%%%%%%%%
\paragraph{Invariants}
\begin{table}[!t]
\centering
\addtolength{\leftskip}{-3cm}
\addtolength{\rightskip}{-3cm}
\begin{tabular}{|c|c|ccccccccccccc|}
\hline
\multirow{3}{*}{Initialisation} & \multirow{3}{*}{Layer} & \multicolumn{13}{c|}{Experimental Invariants} \\ \cline{3-15} 
& & \multicolumn{2}{c|}{Linear} & \multicolumn{11}{c|}{Quadratic} \\ \cline{3-15} 
& & \multicolumn{1}{c|}{$I_1$} & \multicolumn{1}{c|}{$I_2$} & \multicolumn{1}{c|}{$I_3$} & \multicolumn{1}{c|}{$I_4$} & \multicolumn{1}{c|}{$I_5$} & \multicolumn{1}{c|}{$I_6$} & \multicolumn{1}{c|}{$I_7$} & \multicolumn{1}{c|}{$I_8$} & \multicolumn{1}{c|}{$I_9$} & \multicolumn{1}{c|}{$I_{10}$} & \multicolumn{1}{c|}{$I_{11}$} & \multicolumn{1}{c|}{$I_{12}$} & $I_{13}$                 \\ \hline
\multirow{3}{*}{Gaussian}       & 1                      & \multicolumn{1}{c|}{\tiny{-13[-3]}} & \multicolumn{1}{c|}{\tiny{63[-3]}}  & \multicolumn{1}{c|}{\tiny{10[0]}}  & \multicolumn{1}{c|}{\tiny{92[-2]}} & \multicolumn{1}{c|}{\tiny{96[-2]}} & \multicolumn{1}{c|}{\tiny{95[-2]}} & \multicolumn{1}{c|}{\tiny{99[-1]}} & \multicolumn{1}{c|}{\tiny{98[-1]}} & \multicolumn{1}{c|}{\tiny{69[-2]}} & \multicolumn{1}{c|}{\tiny{95[-1]}} & \multicolumn{1}{c|}{\tiny{10[-1]}} & \multicolumn{1}{c|}{\tiny{98[-2]}} & \tiny{97[-2]} \\ \cline{2-15} 
& 2                      & \multicolumn{1}{c|}{\tiny{-34[-3]}} & \multicolumn{1}{c|}{\tiny{14[-3]}}  & \multicolumn{1}{c|}{\tiny{10[0]}}  & \multicolumn{1}{c|}{\tiny{97[-2]}} & \multicolumn{1}{c|}{\tiny{10[-1]}} & \multicolumn{1}{c|}{\tiny{99[-2]}} & \multicolumn{1}{c|}{\tiny{10[0]}}  & \multicolumn{1}{c|}{\tiny{10[0]}}  & \multicolumn{1}{c|}{\tiny{87[-2]}} & \multicolumn{1}{c|}{\tiny{99[-1]}} & \multicolumn{1}{c|}{\tiny{99[-2]}} & \multicolumn{1}{c|}{\tiny{10[-1]}} & \tiny{11[-1]} \\ \cline{2-15} 
& 3                      & \multicolumn{1}{c|}{\tiny{22[-3]}}  & \multicolumn{1}{c|}{\tiny{-22[-2]}} & \multicolumn{1}{c|}{\tiny{10[0]}}  & \multicolumn{1}{c|}{\tiny{11[-1]}} & \multicolumn{1}{c|}{\tiny{99[-2]}} & \multicolumn{1}{c|}{\tiny{10[-1]}} & \multicolumn{1}{c|}{\tiny{98[-1]}} & \multicolumn{1}{c|}{\tiny{99[-1]}} & \multicolumn{1}{c|}{\tiny{11[-1]}} & \multicolumn{1}{c|}{\tiny{97[-1]}} & \multicolumn{1}{c|}{\tiny{10[-1]}} & \multicolumn{1}{c|}{\tiny{11[-1]}} & \tiny{11[-1]} \\ \hline    
\multirow{3}{*}{Uniform}        & 1                      & \multicolumn{1}{c|}{\tiny{-33[-3]}} & \multicolumn{1}{c|}{\tiny{-63[-3]}} & \multicolumn{1}{c|}{\tiny{33[-1]}} & \multicolumn{1}{c|}{\tiny{33[-2]}} & \multicolumn{1}{c|}{\tiny{34[-2]}} & \multicolumn{1}{c|}{\tiny{32[-2]}} & \multicolumn{1}{c|}{\tiny{34[-1]}} & \multicolumn{1}{c|}{\tiny{34[-1]}} & \multicolumn{1}{c|}{\tiny{34[-2]}} & \multicolumn{1}{c|}{\tiny{34[-1]}} & \multicolumn{1}{c|}{\tiny{33[-2]}} & \multicolumn{1}{c|}{\tiny{32[-2]}} & \tiny{29[-2]} \\ \cline{2-15} 
& 2                      & \multicolumn{1}{c|}{\tiny{37[-3]}}  & \multicolumn{1}{c|}{\tiny{80[-3]}}  & \multicolumn{1}{c|}{\tiny{33[-1]}} & \multicolumn{1}{c|}{\tiny{33[-2]}} & \multicolumn{1}{c|}{\tiny{33[-2]}} & \multicolumn{1}{c|}{\tiny{33[-2]}} & \multicolumn{1}{c|}{\tiny{33[-1]}} & \multicolumn{1}{c|}{\tiny{33[-1]}} & \multicolumn{1}{c|}{\tiny{33[-2]}} & \multicolumn{1}{c|}{\tiny{33[-1]}} & \multicolumn{1}{c|}{\tiny{33[-2]}} & \multicolumn{1}{c|}{\tiny{34[-2]}} & \tiny{35[-2]} \\ \cline{2-15} 
& 3                      & \multicolumn{1}{c|}{\tiny{-99[-4]}} & \multicolumn{1}{c|}{\tiny{38[-3]}}  & \multicolumn{1}{c|}{\tiny{33[-1]}} & \multicolumn{1}{c|}{\tiny{35[-2]}} & \multicolumn{1}{c|}{\tiny{34[-2]}} & \multicolumn{1}{c|}{\tiny{33[-2]}} & \multicolumn{1}{c|}{\tiny{33[-1]}} & \multicolumn{1}{c|}{\tiny{33[-1]}} & \multicolumn{1}{c|}{\tiny{40[-2]}} & \multicolumn{1}{c|}{\tiny{34[-1]}} & \multicolumn{1}{c|}{\tiny{34[-2]}} & \multicolumn{1}{c|}{\tiny{32[-2]}} & \tiny{33[-2]} \\ \hline
\end{tabular}
\caption{Observed average values of the linear and quadratic Gaussian matrix model invariants $\hat{I}_i$ for the MNIST classification problem, for each weight initialisation scheme and layer, averaged over the 1000 runs, at initialisation \textit{prior} to training. Values are shown to two significant figures with notation $a[b] = a \times 10^b$. The invariants $I_i$ are indexed as: $\{ \sum_i W_{ii}, \sum_{i,j} W_{ij}, \sum_{i,j} W_{ij}^2, \sum_{i,j} W_{ij}W_{ji}, \sum_{i,j} W_{ii}W_{ij}, \sum_{i,j} W_{ii}W_{ji}, \sum_{i,j,k} W_{ij}W_{ik},\\ \sum_{i,j,k} W_{ij}W_{kj}, \sum_{i,j,k} W_{ij}W_{jk}, \sum_{i,j,k,l} W_{ij}W_{kl}, \sum_{i} W_{ii}^2, \sum_{i,j} W_{ii}W_{jj}, \sum_{i,j,k} W_{ii}W_{jk} \}$.}
\label{tab:pre-invariants}
\end{table}

The deviations, as defined in \eqref{eq:LQinv_deviation}, of the linear \& quadratic invariants computed on the NN weight data at initialisation from the expected values are reported in Table \ref{tab:pre-invariants}.
First we consider the Gaussian initialisation case; for all invariants across all layers the mean deviation was $1.06$, expectedly close to 1 to verify the initialisation.  %1.0555710887747602, across the layers (1.22941561, 0.53290788, 1.40438978)
The (min, max) deviations were $(0.054, 3.75)$ respectively coming from the layer 2 $I_7$ and layer 3 $I_{12}$ invariants. %(0.054322851812488165, 3.7542067159477694)
Ordering the invariant deviations within each layer according to size showed no significant correlation between the layers' orderings, indicating the invariants are equitably represented in this symmetric decomposition. %(i.e. ranking the 13 invariant deviations for the first layer, then repeating for the second layer, then the third, then comparing these rankings); could do spearmans rank here but probably overkill
Equivalently, ordering the invariant deviations across the layers according to size showed no significant correlation between the invariants, indicating the layers are also equally well represented by the same decomposition. %(i.e. ranking the 3 layers according to the first invariant deviation, then repeating for the second, then the third, ..., up to the 13th, then comparing these rankings)
Since the invariants are all statistically acceptable, this provides reassurance in the applicability of this permutation-symmetry decomposition of the sampled weight matrices.
%...discuss proportions of invariants within [1,2,3,4] standard errors of the expectaition value, within each layer: [[0.46153846153846156, 0.7692307692307693, 0.9230769230769231, 1.0], [0.8461538461538461, 1.0, 1.0, 1.0], [0.38461538461538464, 0.6923076923076923, 0.9230769230769231, 1.0]]?

Secondly, for the Uniform initialisation case, for all invariants across all layers the mean deviation was $0.88$, again expectedly close to 1 to verify the initialisation. %0.878708469837843
Surprisingly the uniform initialisation appears to fit the matrix model even better, although up to the statistical uncertainty they are both satisfactory.
The (min, max) deviations were $(0.013, 2.07)$ respectively coming from the layer 3 $I_{13}$ and layer 2 $I_{11}$ invariants. %(0.01342797532264499, 2.070380846150333) 
Again, both orderings of the deviations, within the layers or across the layers, showed no significant correlation, implying again this symmetric decomposition is suitably representative for this initialisation scheme.

%%%%%%%%%%%%%%%%%%%%%%%%%%%%%%%%%%%%%%%%%%%%%%%%%
\paragraph{Model Parameters}
\begin{table}[!t]
\centering
\addtolength{\leftskip}{-4cm}
\addtolength{\rightskip}{-4cm}
\begin{tabular}{|c|c|ccccccccccccc|}
\hline
\multirow{2}{*}{Initialisation} & \multirow{2}{*}{Layer} & \multicolumn{13}{c|}{Model Parameters} \\ \cline{3-15} 
& & \multicolumn{1}{c|}{$f_1$} & \multicolumn{1}{c|}{$f_2$} & \multicolumn{1}{c|}{$f_3$} & \multicolumn{1}{c|}{$f_4$} & \multicolumn{1}{c|}{$f_5$} & \multicolumn{1}{c|}{$f_6$} & \multicolumn{1}{c|}{$f_7$} & \multicolumn{1}{c|}{$f_8$} & \multicolumn{1}{c|}{$f_9$} & \multicolumn{1}{c|}{$f_{10}$} & \multicolumn{1}{c|}{$f_{11}$} & \multicolumn{1}{c|}{$f_{12}$} & $f_{13}$ \\ \hline
\multirow{3}{*}{Gaussian}       & 1                      & \multicolumn{1}{c|}{\tiny{63[-4]}} & \multicolumn{1}{c|}{\tiny{-65[-4]}} & \multicolumn{1}{c|}{\tiny{95[-3]}} & \multicolumn{1}{c|}{\tiny{75[-5]}} & \multicolumn{1}{c|}{\tiny{98[-3]}} & \multicolumn{1}{c|}{\tiny{98[-3]}} & \multicolumn{1}{c|}{\tiny{-29[-4]}} & \multicolumn{1}{c|}{\tiny{-22[-5]}} & \multicolumn{1}{c|}{\tiny{100[-3]}} & \multicolumn{1}{c|}{\tiny{-14[-5]}} & \multicolumn{1}{c|}{\tiny{10[-2]}} & \multicolumn{1}{c|}{\tiny{10[-2]}} & \multicolumn{1}{c|}{\tiny{10[-2]}} \\ \cline{2-15} 
& 2 & \multicolumn{1}{c|}{\tiny{14[-4]}} & \multicolumn{1}{c|}{\tiny{-12[-3]}} & \multicolumn{1}{c|}{\tiny{99[-3]}} & \multicolumn{1}{c|}{\tiny{33[-4]}} & \multicolumn{1}{c|}{\tiny{98[-3]}} & \multicolumn{1}{c|}{\tiny{10[-2]}} & \multicolumn{1}{c|}{\tiny{-14[-4]}} & \multicolumn{1}{c|}{\tiny{-46[-5]}} & \multicolumn{1}{c|}{\tiny{10[-2]}} & \multicolumn{1}{c|}{\tiny{59[-5]}} & \multicolumn{1}{c|}{\tiny{99[-3]}} & \multicolumn{1}{c|}{\tiny{10[-2]}} & \multicolumn{1}{c|}{\tiny{10[-2]}} \\ \cline{2-15} 
& 3 & \multicolumn{1}{c|}{\tiny{-22[-3]}} & \multicolumn{1}{c|}{\tiny{15[-3]}} & \multicolumn{1}{c|}{\tiny{97[-3]}} & \multicolumn{1}{c|}{\tiny{54[-4]}} & \multicolumn{1}{c|}{\tiny{11[-2]}} & \multicolumn{1}{c|}{\tiny{99[-3]}} & \multicolumn{1}{c|}{\tiny{16[-4]}} & \multicolumn{1}{c|}{\tiny{47[-5]}} & \multicolumn{1}{c|}{\tiny{98[-3]}} & \multicolumn{1}{c|}{\tiny{-63[-5]}} & \multicolumn{1}{c|}{\tiny{10[-2]}} & \multicolumn{1}{c|}{\tiny{10[-2]}} & \multicolumn{1}{c|}{\tiny{10[-2]}}  \\ \cline{1-15} 
\multirow{3}{*}{Uniform}        & 1                      & \multicolumn{1}{c|}{\tiny{-63[-4]}} & \multicolumn{1}{c|}{\tiny{-90[-4]}} & \multicolumn{1}{c|}{\tiny{34[-3]}} & \multicolumn{1}{c|}{\tiny{-16[-4]}} & \multicolumn{1}{c|}{\tiny{33[-3]}} & \multicolumn{1}{c|}{\tiny{34[-3]}} & \multicolumn{1}{c|}{\tiny{52[-6]}} & \multicolumn{1}{c|}{\tiny{-62[-5]}} & \multicolumn{1}{c|}{\tiny{34[-3]}} & \multicolumn{1}{c|}{\tiny{30[-5]}} & \multicolumn{1}{c|}{\tiny{34[-3]}} & \multicolumn{1}{c|}{\tiny{33[-3]}} & \multicolumn{1}{c|}{\tiny{33[-3]}}  \\ \cline{2-15} 
& 2                      & \multicolumn{1}{c|}{\tiny{80[-4]}} & \multicolumn{1}{c|}{\tiny{98[-4]}} & \multicolumn{1}{c|}{\tiny{33[-3]}} & \multicolumn{1}{c|}{\tiny{61[-5]}} & \multicolumn{1}{c|}{\tiny{33[-3]}} & \multicolumn{1}{c|}{\tiny{33[-3]}} & \multicolumn{1}{c|}{\tiny{-43[-6]}} & \multicolumn{1}{c|}{\tiny{-15[-5]}} & \multicolumn{1}{c|}{\tiny{33[-3]}} & \multicolumn{1}{c|}{\tiny{-28[-5]}} & \multicolumn{1}{c|}{\tiny{33[-3]}} & \multicolumn{1}{c|}{\tiny{33[-3]}} & \multicolumn{1}{c|}{\tiny{33[-3]}} \\ \cline{2-15} 
& 3                      & \multicolumn{1}{c|}{\tiny{38[-4]}} & \multicolumn{1}{c|}{\tiny{-46[-4]}} & \multicolumn{1}{c|}{\tiny{34[-3]}} & \multicolumn{1}{c|}{\tiny{-67[-6]}} & \multicolumn{1}{c|}{\tiny{32[-3]}} & \multicolumn{1}{c|}{\tiny{33[-3]}} & \multicolumn{1}{c|}{\tiny{70[-5]}} & \multicolumn{1}{c|}{\tiny{-19[-5]}} & \multicolumn{1}{c|}{\tiny{33[-3]}} & \multicolumn{1}{c|}{\tiny{38[-6]}} & \multicolumn{1}{c|}{\tiny{34[-3]}} & \multicolumn{1}{c|}{\tiny{33[-3]}} & \multicolumn{1}{c|}{\tiny{33[-3]}}   \\ \hline
\end{tabular}
\caption{Observed values of the Gaussian matrix model parameters $f_i$ (at linear and quadratic orders) for the MNIST classification problem, for each weight initialisation scheme and layer, averaged over the 1000 runs, at initialisation \textit{prior} to training. Values are shown to two significant figures with notation $a[b] = a \times 10^b$.}
\label{tab:pre-modelparams}
\end{table}

The deviations, as defined in \eqref{eq:param_deviation}, of the PIGMM parameters, when fitted to the NN weight data at initialisation, from the expected values are reported in Table \ref{tab:pre-modelparams}.
Starting again with the Gaussian initialisation scheme, for all model parameters across all layers the mean deviation was $0.81$, expectedly close to 1 to verify the initialisation.  % 0.8070323197668782
The (min, max) deviations were $(0.033, 2.44)$ respectively coming from the layer 2 $f_9$ and layer 1 $f_7$ model parameters. %(0.032939708581734534, 2.438605885175737)
Both orderings of the model parameter deviations, within the layers and across the layers, showed no significant correlation, hence indicating that parts of the model corresponding to each term in the action are equally representative for the data.
Since the model parameter deviations are also all statistically acceptable, this provides expected reassurance in the applicability of this Gaussian matrix model for the sampled weight matrices at initialisation.

Following with the consideration of the Uniform initialisation case, for all model parameters across all layers the mean deviation was $0.70$, again expectedly close to 1 to verify the initialisation. %0.6974219939907206
The (min, max) deviations were $(0.0024, 1.81)$ respectively coming from the layer 2 $f_3$ and layer 3 $f_7$ model parameters. %(0.0023907699569807764, 1.8121908390559784)
Again, both orderings of the deviations, within the layers or across the layers, showed no significant correlation, implying again the Gaussian matrix model is suitably representative for this initialisation scheme.

As a final comment, we emphasise the surprising applicability of this permutation-invariant Gaussian matrix model to the NN weight matrix data under the Uniform initialisation scheme.
The general permutation-invariant Gaussian matrix model is designed on sampling matrices from a Gaussian distribution, not a Uniform one.
However, since we use the mean invariant values, and a large number of samples for that mean (1000), the distribution of the mean will be well approximated by a Gaussian distribution in accordance with the Central Limit Theorem.
This is likely why the considered PIGMM is an equally good model for the weight matrices, or more precisely the average weight matrices across the runs.

%%%%%%%%%%%%%%%%%%%%%%%%%%%%%%%%%%%%%%%%%%%%%%%%%
\subsubsection{After Training}\label{sec:lq_after}
During the training process the weights are updated such that the architecture approximates a function which performs well on the task.
As explained in §\ref{sec:perm_intro}, since the training optimiser has no clear motivation to respect the simple Gaussian assumption of the initialisation as the weights are updated, it is expected that the strong fitting at initialisation of the \textit{simple} Gaussian PIGMM model (from the initialisation distribution) is broken down as a more general PIGMM is required.

In this subsection we consider the parameters of the final NN functions after training (epoch 50), and the status of this \textit{simple} Gaussianity property in these final trained models across the relevant 6 weight matrix ensembles (2 initialisations, 3 layers).

%%%%%%%%%%%%%%%%%%%%%%%%%%%%%%%%%%%%%%%%%%%%%%%%%
\paragraph{Invariants}
\begin{table}[!t]
\centering
\addtolength{\leftskip}{-3cm}
\addtolength{\rightskip}{-3cm}
\begin{tabular}{|c|c|ccccccccccccc|}
\hline
\multirow{3}{*}{Initialisation} & \multirow{3}{*}{Layer} & \multicolumn{13}{c|}{Experimental Invariants} \\ \cline{3-15} 
& & \multicolumn{2}{c|}{Linear} & \multicolumn{11}{c|}{Quadratic} \\ \cline{3-15} 
 &                        & \multicolumn{1}{c|}{$I_1$} & \multicolumn{1}{c|}{$I_2$} & \multicolumn{1}{c|}{$I_3$} & \multicolumn{1}{c|}{$I_4$} & \multicolumn{1}{c|}{$I_5$} & \multicolumn{1}{c|}{$I_6$} & \multicolumn{1}{c|}{$I_7$} & \multicolumn{1}{c|}{$I_8$} & \multicolumn{1}{c|}{$I_9$} & \multicolumn{1}{c|}{$I_{10}$} & \multicolumn{1}{c|}{$I_{11}$} & \multicolumn{1}{c|}{$I_{12}$} & $I_{13}$ \\ \hline
\multirow{3}{*}{Gaussian}       & 1                      & \multicolumn{1}{c|}{\tiny{37[-2]}} & \multicolumn{1}{c|}{\tiny{38[-1]}} & \multicolumn{1}{c|}{\tiny{14[0]}} & \multicolumn{1}{c|}{\tiny{14[-1]}} & \multicolumn{1}{c|}{\tiny{13[-1]}} & \multicolumn{1}{c|}{\tiny{13[-1]}} & \multicolumn{1}{c|}{\tiny{13[0]}} & \multicolumn{1}{c|}{\tiny{13[0]}} & \multicolumn{1}{c|}{\tiny{21[-1]}} & \multicolumn{1}{c|}{\tiny{23[0]}} & \multicolumn{1}{c|}{\tiny{14[-1]}} & \multicolumn{1}{c|}{\tiny{15[-1]}} & \multicolumn{1}{c|}{\tiny{22[-1]}}  \\ \cline{2-15}
& 2                      & \multicolumn{1}{c|}{\tiny{47[-2]}} & \multicolumn{1}{c|}{\tiny{51[-1]}} & \multicolumn{1}{c|}{\tiny{19[0]}} & \multicolumn{1}{c|}{\tiny{21[-1]}} & \multicolumn{1}{c|}{\tiny{17[-1]}} & \multicolumn{1}{c|}{\tiny{17[-1]}} & \multicolumn{1}{c|}{\tiny{18[0]}} & \multicolumn{1}{c|}{\tiny{18[0]}} & \multicolumn{1}{c|}{\tiny{38[-1]}} & \multicolumn{1}{c|}{\tiny{39[0]}} & \multicolumn{1}{c|}{\tiny{19[-1]}} & \multicolumn{1}{c|}{\tiny{21[-1]}} & \multicolumn{1}{c|}{\tiny{36[-1]}} \\ \cline{2-15} 
& 3                      & \multicolumn{1}{c|}{\tiny{-61[-2]}} & \multicolumn{1}{c|}{\tiny{-63[-1]}} & \multicolumn{1}{c|}{\tiny{27[0]}} & \multicolumn{1}{c|}{\tiny{33[-1]}} & \multicolumn{1}{c|}{\tiny{18[-1]}} & \multicolumn{1}{c|}{\tiny{34[-1]}} & \multicolumn{1}{c|}{\tiny{20[0]}} & \multicolumn{1}{c|}{\tiny{35[0]}} & \multicolumn{1}{c|}{\tiny{69[-1]}} & \multicolumn{1}{c|}{\tiny{71[0]}} & \multicolumn{1}{c|}{\tiny{26[-1]}} & \multicolumn{1}{c|}{\tiny{32[-1]}} & \multicolumn{1}{c|}{\tiny{68[-1]}}  \\ \cline{1-15} 
\multirow{3}{*}{Uniform}        & 1                      & \multicolumn{1}{c|}{\tiny{38[-2]}} & \multicolumn{1}{c|}{\tiny{40[-1]}} & \multicolumn{1}{c|}{\tiny{89[-1]}} & \multicolumn{1}{c|}{\tiny{10[-1]}} & \multicolumn{1}{c|}{\tiny{74[-2]}} & \multicolumn{1}{c|}{\tiny{88[-2]}} & \multicolumn{1}{c|}{\tiny{77[-1]}} & \multicolumn{1}{c|}{\tiny{90[-1]}} & \multicolumn{1}{c|}{\tiny{21[-1]}} & \multicolumn{1}{c|}{\tiny{20[0]}} & \multicolumn{1}{c|}{\tiny{89[-2]}} & \multicolumn{1}{c|}{\tiny{10[-1]}} & \multicolumn{1}{c|}{\tiny{19[-1]}}   \\ \cline{2-15} 
& 2                      & \multicolumn{1}{c|}{\tiny{50[-2]}} & \multicolumn{1}{c|}{\tiny{46[-1]}} & \multicolumn{1}{c|}{\tiny{14[0]}} & \multicolumn{1}{c|}{\tiny{16[-1]}} & \multicolumn{1}{c|}{\tiny{13[-1]}} & \multicolumn{1}{c|}{\tiny{12[-1]}} & \multicolumn{1}{c|}{\tiny{12[0]}} & \multicolumn{1}{c|}{\tiny{12[0]}} & \multicolumn{1}{c|}{\tiny{28[-1]}} & \multicolumn{1}{c|}{\tiny{29[0]}} & \multicolumn{1}{c|}{\tiny{14[-1]}} & \multicolumn{1}{c|}{\tiny{16[-1]}} & \multicolumn{1}{c|}{\tiny{31[-1]}} \\ \cline{2-15} 
& 3                      & \multicolumn{1}{c|}{\tiny{-64[-2]}} & \multicolumn{1}{c|}{\tiny{-63[-1]}} & \multicolumn{1}{c|}{\tiny{19[0]}} & \multicolumn{1}{c|}{\tiny{22[-1]}} & \multicolumn{1}{c|}{\tiny{13[-1]}} & \multicolumn{1}{c|}{\tiny{20[-1]}} & \multicolumn{1}{c|}{\tiny{14[0]}} & \multicolumn{1}{c|}{\tiny{19[0]}} & \multicolumn{1}{c|}{\tiny{54[-1]}} & \multicolumn{1}{c|}{\tiny{55[0]}} & \multicolumn{1}{c|}{\tiny{18[-1]}} & \multicolumn{1}{c|}{\tiny{23[-1]}} & \multicolumn{1}{c|}{\tiny{54[-1]}}  \\ \hline
\end{tabular}
\caption{Observed average values of the linear and quadratic Gaussian matrix model invariants $\hat{I}_i$ for the MNIST classification problem, for each weight initialisation scheme and layer, averaged over the 1000 runs, \textit{after} 50 epochs of training. Values are shown to two significant figures with notation $a[b] = a \times 10^b$ The invariants $I_i$ are indexed as: $\{ \sum_i W_{ii}, \sum_{i,j} W_{ij}, \sum_{i,j} W_{ij}^2, \sum_{i,j} W_{ij}W_{ji}, \sum_{i,j} W_{ii}W_{ij}, \sum_{i,j} W_{ii}W_{ji}, \sum_{i,j,k} W_{ij}W_{ik},\\ \sum_{i,j,k} W_{ij}W_{kj}, \sum_{i,j,k} W_{ij}W_{jk}, \sum_{i,j,k,l} W_{ij}W_{kl}, \sum_{i} W_{ii}^2, \sum_{i,j} W_{ii}W_{jj}, \sum_{i,j,k} W_{ii}W_{jk} \}$.}
\label{tab:post-invariants}
\end{table}

The deviations, as defined in \eqref{eq:LQinv_deviation}, of the linear \& quadratic invariants computed on the NN weight data after training from the expected values of the initialisation distributions are reported in Table \ref{tab:post-invariants}.
First considering the Gaussian initialisation scheme the mean deviation of the invariants across all layers was $64$, with (min, max) values of $(8,372)$ respectively coming from layer 1 $I_5$ and layer 3 $I_3$. %(min,mean,max): (20.656067220698667, 234.32247374812354, 1650.7690597480946)  
Equivalently, for the Uniform initialisation scheme the mean deviation of the invariants across all layers was $234$, with (min, max) values of $(21,1651)$ respectively coming from layer 1 $I_0$ and layer 3 $I_3$. %(min,mean,max): (20.656067220698667, 234.32247374812354, 1650.7690597480946)  
Both schemes show that after training the invariants have significantly deviated from the initialisation distributions and the respective simple Gaussian assumption.

%%%%%%%%%%%%%%%%%%%%%%%%%%%%%%%%%%%%%%%%%%%%%%%%%
\paragraph{Model Parameters}
\begin{table}[!t]
\centering
\addtolength{\leftskip}{-4cm}
\addtolength{\rightskip}{-4cm}
\begin{tabular}{|c|c|ccccccccccccc|}
\hline
\multirow{2}{*}{Initialisation} & \multirow{2}{*}{Layer} & \multicolumn{13}{c|}{Model Parameters} \\ \cline{3-15} 
& & \multicolumn{1}{c|}{$f_1$} & \multicolumn{1}{c|}{$f_2$} & \multicolumn{1}{c|}{$f_3$} & \multicolumn{1}{c|}{$f_4$} & \multicolumn{1}{c|}{$f_5$} & \multicolumn{1}{c|}{$f_6$} & \multicolumn{1}{c|}{$f_7$} & \multicolumn{1}{c|}{$f_8$} & \multicolumn{1}{c|}{$f_9$} & \multicolumn{1}{c|}{$f_{10}$} & \multicolumn{1}{c|}{$f_{11}$} & \multicolumn{1}{c|}{$f_{12}$} & $f_{13}$ \\ \hline
\multirow{3}{*}{Gaussian}       & 1                      & \multicolumn{1}{c|}{\tiny{38[-2]}} & \multicolumn{1}{c|}{\tiny{-40[-4]}} & \multicolumn{1}{c|}{\tiny{82[-3]}} & \multicolumn{1}{c|}{\tiny{62[-5]}} & \multicolumn{1}{c|}{\tiny{14[-2]}} & \multicolumn{1}{c|}{\tiny{12[-2]}} & \multicolumn{1}{c|}{\tiny{-18[-4]}} & \multicolumn{1}{c|}{\tiny{26[-4]}} & \multicolumn{1}{c|}{\tiny{12[-2]}} & \multicolumn{1}{c|}{\tiny{-29[-5]}} & \multicolumn{1}{c|}{\tiny{14[-2]}} & \multicolumn{1}{c|}{\tiny{14[-2]}} & \multicolumn{1}{c|}{\tiny{14[-2]}}  \\ \cline{2-15} 
& 2 & \multicolumn{1}{c|}{\tiny{51[-2]}} & \multicolumn{1}{c|}{\tiny{-14[-3]}} & \multicolumn{1}{c|}{\tiny{13[-2]}} & \multicolumn{1}{c|}{\tiny{-28[-4]}} & \multicolumn{1}{c|}{\tiny{20[-2]}} & \multicolumn{1}{c|}{\tiny{15[-2]}} & \multicolumn{1}{c|}{\tiny{-12[-4]}} & \multicolumn{1}{c|}{\tiny{-44[-5]}} & \multicolumn{1}{c|}{\tiny{15[-2]}} & \multicolumn{1}{c|}{\tiny{-10[-4]}} & \multicolumn{1}{c|}{\tiny{19[-2]}} & \multicolumn{1}{c|}{\tiny{20[-2]}} & \multicolumn{1}{c|}{\tiny{20[-2]}} \\ \cline{2-15} 
& 3 & \multicolumn{1}{c|}{\tiny{-63[-2]}} & \multicolumn{1}{c|}{\tiny{69[-4]}} & \multicolumn{1}{c|}{\tiny{31[-2]}} & \multicolumn{1}{c|}{\tiny{-53[-4]}} & \multicolumn{1}{c|}{\tiny{28[-2]}} & \multicolumn{1}{c|}{\tiny{31[-2]}} & \multicolumn{1}{c|}{\tiny{-19[-4]}} & \multicolumn{1}{c|}{\tiny{-17[-4]}} & \multicolumn{1}{c|}{\tiny{14[-2]}} & \multicolumn{1}{c|}{\tiny{-56[-4]}} & \multicolumn{1}{c|}{\tiny{26[-2]}} & \multicolumn{1}{c|}{\tiny{28[-2]}} & \multicolumn{1}{c|}{\tiny{27[-2]}} \\ \cline{1-15} 
\multirow{3}{*}{Uniform}        & 1                      & \multicolumn{1}{c|}{\tiny{40[-2]}} & \multicolumn{1}{c|}{\tiny{-85[-4]}} & \multicolumn{1}{c|}{\tiny{43[-3]}} & \multicolumn{1}{c|}{\tiny{-19[-4]}} & \multicolumn{1}{c|}{\tiny{93[-3]}} & \multicolumn{1}{c|}{\tiny{77[-3]}} & \multicolumn{1}{c|}{\tiny{11[-5]}} & \multicolumn{1}{c|}{\tiny{-11[-5]}} & \multicolumn{1}{c|}{\tiny{63[-3]}} & \multicolumn{1}{c|}{\tiny{-32[-5]}} & \multicolumn{1}{c|}{\tiny{92[-3]}} & \multicolumn{1}{c|}{\tiny{91[-3]}} & \multicolumn{1}{c|}{\tiny{91[-3]}}  \\ \cline{2-15} 
& 2                      & \multicolumn{1}{c|}{\tiny{46[-2]}} & \multicolumn{1}{c|}{\tiny{13[-3]}} & \multicolumn{1}{c|}{\tiny{75[-3]}} & \multicolumn{1}{c|}{\tiny{35[-4]}} & \multicolumn{1}{c|}{\tiny{14[-2]}} & \multicolumn{1}{c|}{\tiny{97[-3]}} & \multicolumn{1}{c|}{\tiny{-26[-5]}} & \multicolumn{1}{c|}{\tiny{75[-5]}} & \multicolumn{1}{c|}{\tiny{10[-2]}} & \multicolumn{1}{c|}{\tiny{13[-4]}} & \multicolumn{1}{c|}{\tiny{15[-2]}} & \multicolumn{1}{c|}{\tiny{15[-2]}} & \multicolumn{1}{c|}{\tiny{15[-2]}}  \\ \cline{2-15} 
& 3                      & \multicolumn{1}{c|}{\tiny{-63[-2]}} & \multicolumn{1}{c|}{\tiny{-27[-4]}} & \multicolumn{1}{c|}{\tiny{16[-2]}} & \multicolumn{1}{c|}{\tiny{-42[-4]}} & \multicolumn{1}{c|}{\tiny{19[-2]}} & \multicolumn{1}{c|}{\tiny{15[-2]}} & \multicolumn{1}{c|}{\tiny{-88[-5]}} & \multicolumn{1}{c|}{\tiny{11[-4]}} & \multicolumn{1}{c|}{\tiny{89[-3]}} & \multicolumn{1}{c|}{\tiny{-30[-4]}} & \multicolumn{1}{c|}{\tiny{20[-2]}} & \multicolumn{1}{c|}{\tiny{20[-2]}} & \multicolumn{1}{c|}{\tiny{20[-2]}}  \\ \hline
\end{tabular}
\caption{Observed values of the Gaussian matrix model parameters $f_i$ (at linear and quadratic orders) for the MNIST classification problem, for each weight initialisation scheme and layer, averaged over the 1000 runs, \textit{after} 50 epochs of training. Values are shown to two significant figures with notation $a[b] = a \times 10^b$.}
\label{tab:post-modelparams}
\end{table}

The deviations, as defined in \eqref{eq:param_deviation}, of the PIGMM parameters, when fitted to the NN weight data after training, from the expected values at initialisation are reported in Table \ref{tab:pre-modelparams}.
For the Gaussian initialisation scheme the mean deviation of the model parameters across all layers was $37$, with (min, max) values of $(0.18,207)$ respectively coming from layer 1 $I_4$ and layer 3 $I_{13}$. %(min,mean,max): (0.17598082331456943, 36.57920500591556, 207.13292054236723)
Equivalently, for the Uniform initialisation scheme the mean deviation of the model parameters across all layers was $128$, with (min, max) values of $(0.24,795)$ respectively coming from layer 1 $I_8$ and layer 3 $I_{13}$. %(min,mean,max): (0.23564350498368017, 127.87286382112553, 795.1221870476978) 
Both schemes show that after training the model parameters have significantly deviated from the initialisation distributions and the parameters expected for a PIGMM fitted to the simple Gaussian, therefore motivating the need for a more general model for the trained weight distributions.
More general PIGMMs are posited as candidates for this in this work, and their effectiveness is tested in §\ref{sec:cq}.

%%%%%%%%%%%%%%%%%%%%%%%%%%%%%%%%%%%%%%%%%%%%%%%%%
\paragraph{Between Layers}
To make comparisons between layers we start by considering the average invariant values per layer, which for the invariant deviations are:
\begin{align}
\text{Gaussian} \quad (L1, L2, L3) & = (25, 57, 110)\;,\\ %(25.42273257,  56.60502316, 109.92771605)
\text{Uniform} \quad (L1, L2, L3) & = (127, 229, 347)\;, %(127.29988626, 228.68583029, 346.98170469)
\end{align}
and for the model parameter deviations:
\begin{align}
\text{Gaussian} \quad (L1, L2, L3) & = (15, 32, 62)\;,\\ %(14.96918484, 32.36821901, 62.40021117)
\text{Uniform} \quad (L1, L2, L3) & = (68, 130, 186)\;. %( 67.57844775, 129.85019972, 186.18994399)
\end{align}
Before training these averages were approximately equal across the layers in each case, all $\sim 1$ with the simple Gaussianity assumption not violated; whereas after, in each case, the later layers have larger values.
For the Gaussian initialisation, the invariants and model parameters approximately double with each consecutive layer, and for the Uniform initialisation the average deviations approximately scale linear with the layer number; these relations would be interesting to probe further with much deeper networks in future work.
Overall, these increasing deviations for later layers demonstrate a clear difference in how the training influences the breaking of the simple Gaussianity assumption, it is broken more substantially for later layers.

\begin{table}[!t]
\centering
\begin{tabular}{|ccc|ccc|}
\hline
\multicolumn{3}{|c|}{Measure} & \multicolumn{3}{c|}{PMCC} \\ \hline
\multicolumn{3}{|c|}{Layer Pair}  & \multicolumn{1}{c|}{L1-L2}      & \multicolumn{1}{c|}{L1-L3}       & L2-L3       \\ \hline
\multicolumn{1}{|c|}{\multirow{4}{*}{Invariants}} & \multicolumn{1}{c|}{\multirow{2}{*}{Gaussian}} & Before & \multicolumn{1}{c|}{0.40670172} & \multicolumn{1}{c|}{-0.16146451} & -0.26306281 \\ \cline{3-6} 
\multicolumn{1}{|c|}{} & \multicolumn{1}{c|}{}                          & After  & \multicolumn{1}{c|}{0.97322891} & \multicolumn{1}{c|}{0.88109717}  & 0.93603143  \\ \cline{2-6} 
\multicolumn{1}{|c|}{} & \multicolumn{1}{c|}{\multirow{2}{*}{Uniform}}  & Before & \multicolumn{1}{c|}{0.15929264} & \multicolumn{1}{c|}{-0.64791578} & 0.10118799  \\ \cline{3-6} 
\multicolumn{1}{|c|}{} & \multicolumn{1}{c|}{}                          & After  & \multicolumn{1}{c|}{0.99397488} & \multicolumn{1}{c|}{0.99696663}  & 0.98951672  \\ \hline
\multicolumn{1}{|c|}{\multirow{4}{*}{\begin{tabular}[c]{@{}c@{}}Model\\ Parameters\end{tabular}}} & \multicolumn{1}{c|}{\multirow{2}{*}{Gaussian}} & Before & \multicolumn{1}{c|}{0.31098476} & \multicolumn{1}{c|}{-0.02494182} & 0.08118117  \\ \cline{3-6} 
\multicolumn{1}{|c|}{} & \multicolumn{1}{c|}{}                          & After  & \multicolumn{1}{c|}{0.96129657} & \multicolumn{1}{c|}{0.80764503}  & 0.88123747  \\ \cline{2-6} 
\multicolumn{1}{|c|}{} & \multicolumn{1}{c|}{\multirow{2}{*}{Uniform}}  & Before & \multicolumn{1}{c|}{0.48284886} & \multicolumn{1}{c|}{-0.53995845} & -0.13484479 \\ \cline{3-6} 
\multicolumn{1}{|c|}{} & \multicolumn{1}{c|}{}                          & After  & \multicolumn{1}{c|}{0.9881429}  & \multicolumn{1}{c|}{0.98529427}  & 0.9845063   \\ \hline
\end{tabular}
\caption{PMCC scores between NN layer deviations, each vector includes 13 deviations for each invariant or model parameter respectively, for both initialisation distributions, and with before and after training.}
\label{tab:layer_pmccs}
\end{table}

To further probe how this average change in `deviation scale' between layers splits amongst the invariants and model parameters we now consider the product moment correlation coefficient (PMCC) score for the deviations between the layers (i.e. 3 vectors for each layer of 13 deviations in the order of the 13 invariants have their pairwise PMCCs computed, then repeated for the 13 model parameters).
These PMCC scores are shown for before and after training in Table \ref{tab:layer_pmccs}.
As expected before training the scores are relatively insignificant, since the invariant deviations are of a similar order due to the identical initialisation; whereas after training many are exceptionally close to 1. 
This indicates that the extent which each invariant (or parameter) deviates from simple Gaussianity scales in a similar manner between layers, such that the deviation ratios between invariants (or parameters) in the same layer is maintained and the relation is likely linear between layers.

%%%%%%%%%%%%%%%%%%%%%%%%%%%%%%%%%%%%%%%%%%%%%%%%%
\subsubsection{Throughout Training}\label{sec:lq_throughout}
\begin{figure}[!t]
    \centering
    \begin{subfigure}{0.32\textwidth}
        \centering
        \includegraphics[width=0.98\textwidth]{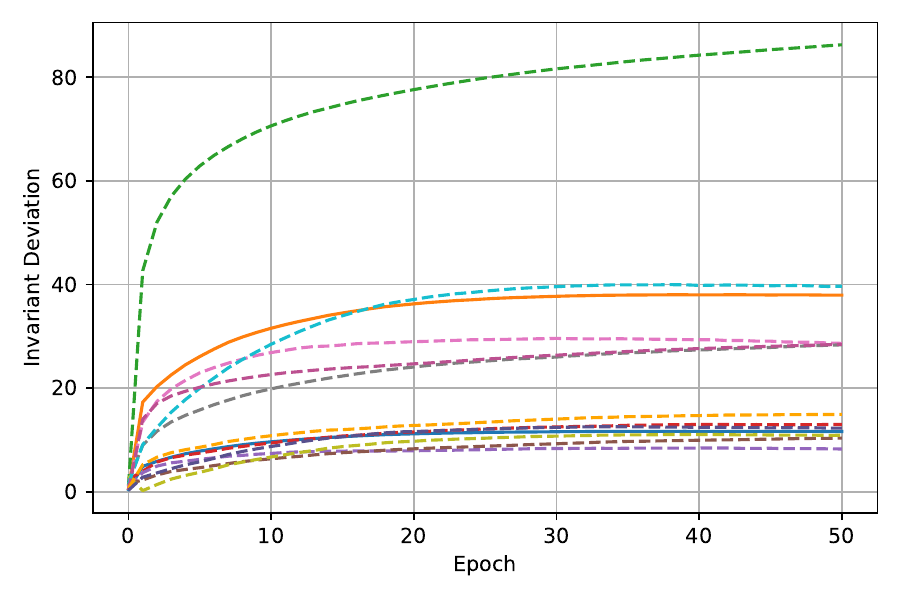}
        \caption{Gaussian L1}
    \end{subfigure} 
    \begin{subfigure}{0.32\textwidth}
        \centering
        \includegraphics[width=0.98\textwidth]{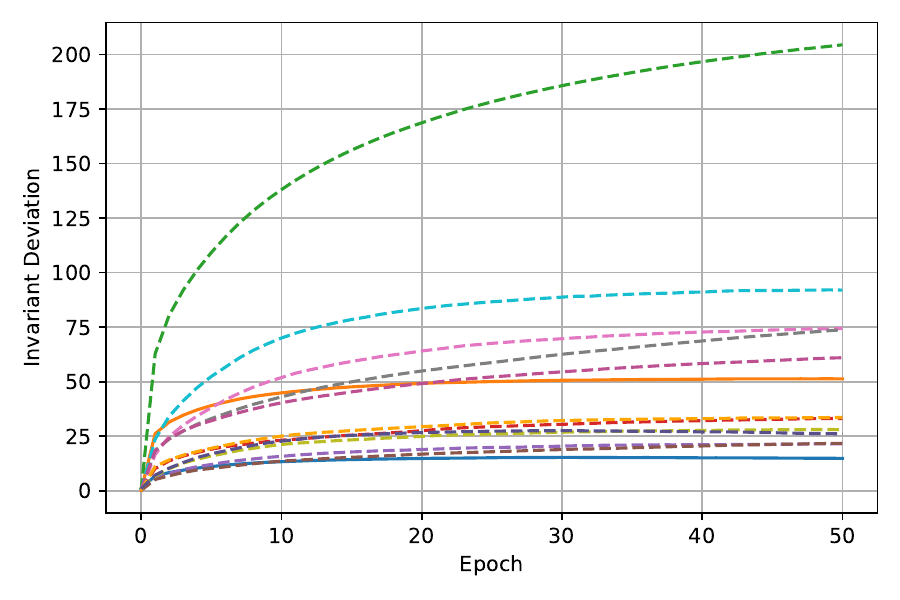}
        \caption{Gaussian L2}
    \end{subfigure}
    \begin{subfigure}{0.32\textwidth}
        \centering
        \includegraphics[width=0.98\textwidth]{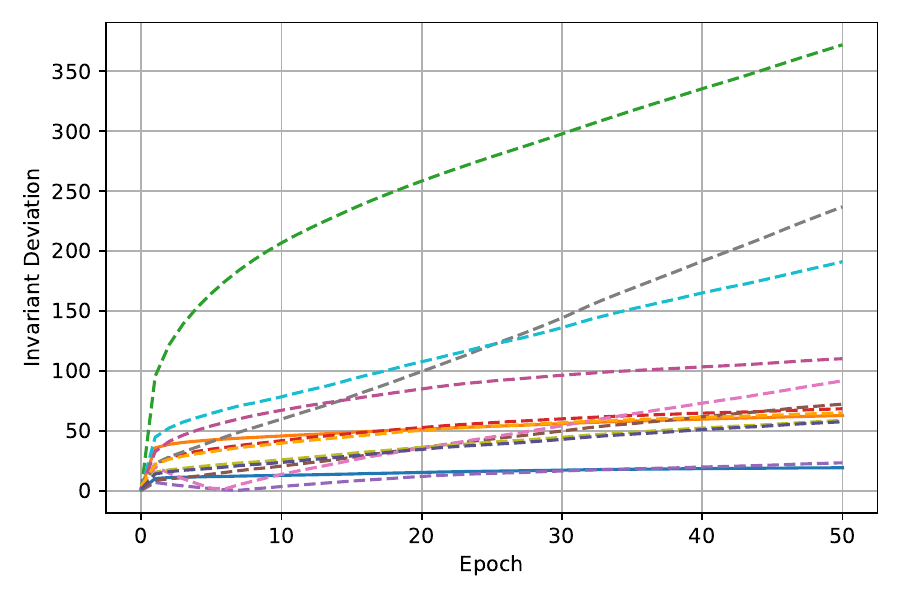}
        \caption{Gaussian L3}
    \end{subfigure}\\ 
    \begin{subfigure}{0.32\textwidth}
        \centering
        \includegraphics[width=0.98\textwidth]{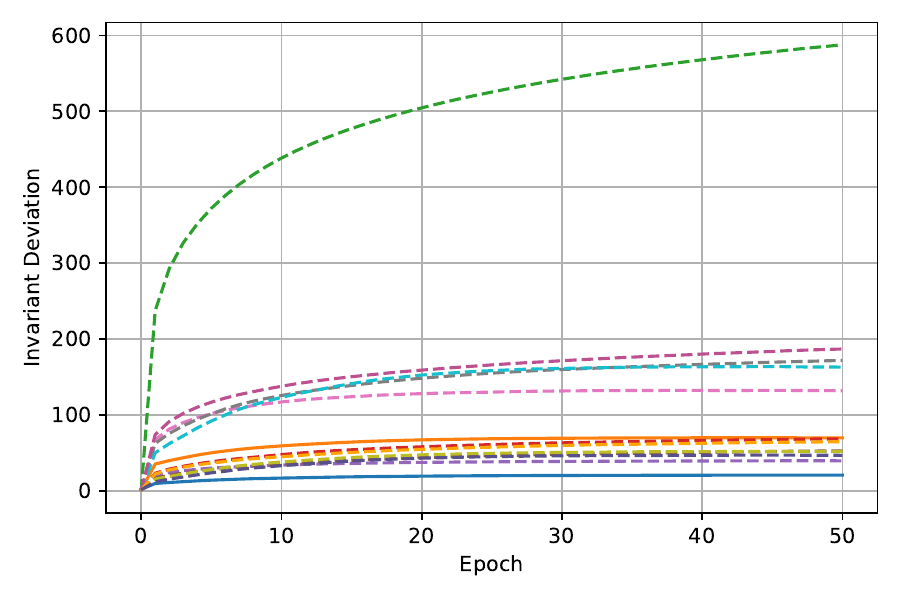}
        \caption{Uniform L1}
    \end{subfigure} 
    \begin{subfigure}{0.32\textwidth}
        \centering
        \includegraphics[width=0.98\textwidth]{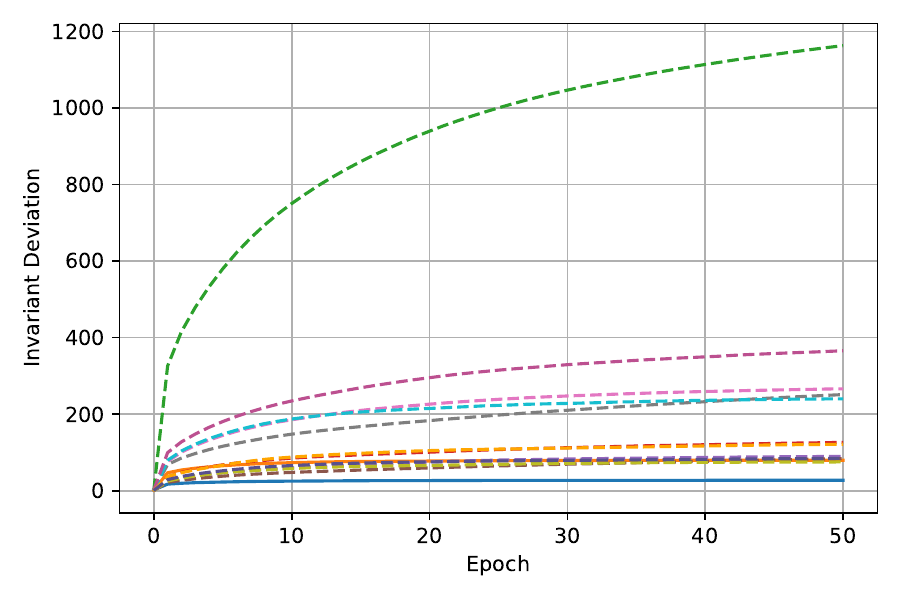}
        \caption{Uniform L2}
    \end{subfigure}
    \begin{subfigure}{0.32\textwidth}
        \centering
        \includegraphics[width=0.98\textwidth]{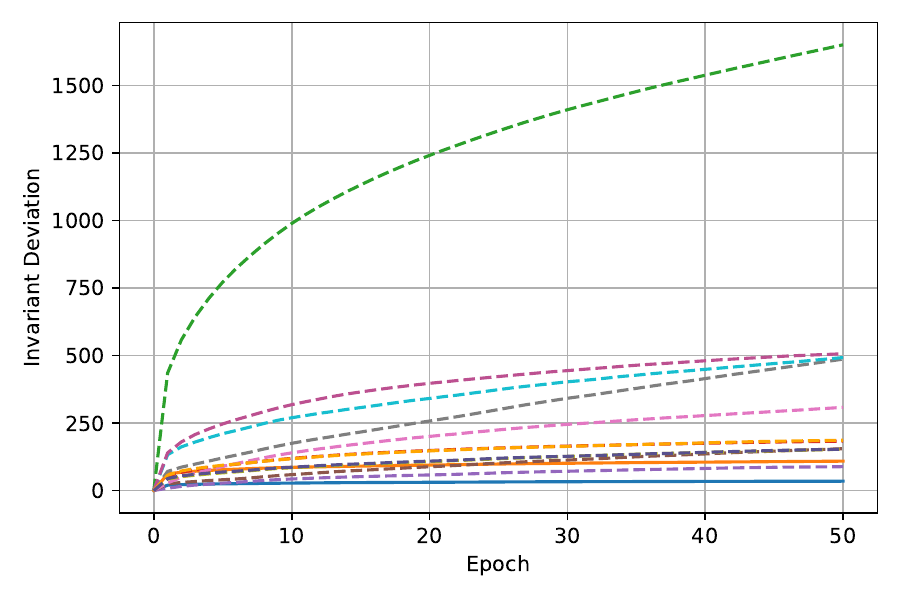}
        \caption{Uniform L3}
    \end{subfigure}\\
    \begin{subfigure}{0.98\textwidth}
        \centering
        \includegraphics[width=0.98\textwidth]{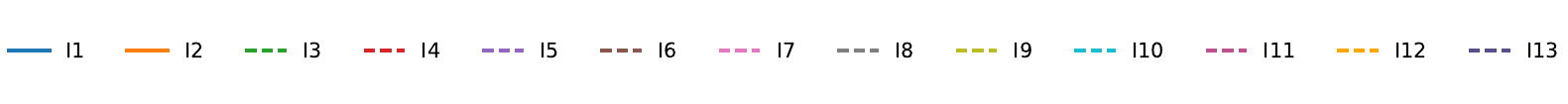}
        \caption{Legend}
    \end{subfigure}
    \caption{Variation of the linear invariant deviations (solid lines) and quadratic invariant deviations (dashed lines), labelled respectively by their invariants $I_1-I_{13}$, across the 50 epochs of training. The legend is the same throughout and collectively shown at the bottom (g) for readability. We emphasise the varying scales in y-axis.}
    \label{fig:LQ_during}
\end{figure}

Using the computed linear and quadratic invariants, and the fitted model parameters, the results in the previous sections show that the simple Gaussian (fitted as this simplest special case of the PIGMM class) is a good fit of the weight matrix ensembles prior to training, but is a bad fit after training.
To examine how smooth this change is, these invariants where also computed after each epoch of training, and the deviation measures (as defined in \eqref{eq:LQinv_deviation}) are shown as plots in Figure \ref{fig:LQ_during}.

In each case the lines are relatively smooth, indicating the deviation from the simplest Gaussian at initialisation is a continuous change.
For both initialisations and for every layer the invariant which deviates the most throughout all the epochs is $I_3 = \sum_{i,j} W_{ij}^2$, since this invariant treats each weight independently and uses all of them perhaps this shows a consistent change of all weight scales away from the original distribution.

The uniform initialisation ensembles express higher deviation values (with larger y-axes in the plots) such that their deviations are usually greater perhaps indicating this initialisation starts the NN function further in model space from the area of good solutions to the problem so the training process causes it to move further.
Comparing between layers, in both initialisations the earlier layers appear to have their weight distributions settle earlier in the training shown by the flatter deviation curves, and additionally the earlier layers have smaller deviations in general (with smaller y-axes scales) such that the change from the simplest Gaussian at initialisation is the smallest.

Therefore, the simplest Gaussian model of initialisation is broken most significantly for deeper layers in a NN and for uniform initialisation.

%%%%%%%%%%%%%%%%%%%%%%%%%%%%%%%%%%%%%%%%%%%%%%%%%
\subsection{Cubic \& Quartic Invariants}\label{sec:cq}
Now that the previous section has established that over training the distribution of weight matrices deviates significantly from the simplest Gaussian assumption affiliated to the initialisation distributions, into model parameter values associated to more general PIGMMs, we now wish to test to what extent these more general PIGMMs are suitable models for the distributions.

PIGMMs rely on the assumptions of general Gaussianity (up to second order terms only in the weights) and permutation-symmetry.
To probe the suitability of the general Gaussianity assumption the PIGMM models are now tested on higher-order invariants, specifically those of cubic and quartic order with the near-minimum and near-maximum number of nodes, as introduced in §\ref{sec:gmms} and §\ref{app:inv_graphs}.
If the Gaussianity assumption is suitable for the weight matrix ensembles then the PIGMM models will well predict the true values of these higher-order invariants for the weight matrices.
Additionally, if the weight matrix distributions significantly break the diagonal permutation symmetry assumption in the PIGMMs then the models will also not be representative, and then badly predict the true values of these symmetric invariants. %is this true? can comments on the perm-sym be made from these results?

To quantify the difference in these 39 higher-order invariant values a new deviation measure is defined, where the predicted theoretical values from the fitted PIGMM model for each of the 306 ensembles are compared to the experimentally observed values computed from the respective matrix data in that ensemble. 
These deviations for the higher-order (Cubic \& Quartic) invariants are defined 
\begin{equation}\label{eq:CQinv_deviation}
    \text{Deviation}_{CQ}(I_i) \vcentcolon = \frac{|I^T_i - \hat{I}^E_i|}{\sigma(I^E_i)}\;,
\end{equation}
where $\hat{I}^E_i$ are the mean experimental values of these higher-order invariants as computed from the data, with standard deviation over the data runs $\sigma(I^E_i)$, and $I^T_i$ the theoretically predicted value from the permutation-invariant Gaussian matrix model fitted to the lower-order invariants and parameters\footnote{We emphasise here the differing nature of this deviation, the variational measure $\sigma$ is experimental and computed across the runs, instead of calculated analytically as before in \eqref{eq:LQinv_deviation} and \eqref{eq:param_deviation}. Also theoretical values are used instead of analytic expectations, and are computed from the model fitted with the lower-order experimental model parameters.}. The deviation defined here for the cubic and quartic invariants has been found to be useful in the application of permutation invariant matrix models in natural language processing \cite{Huber:2022ohf} and finance \cite{Barnes:2023kqc} (§5.3 of the latter gives a discussion of this deviation measure in terms of perturbations of the Gaussian action). 

To further examine these deviations, another measure which we call a `normalised change' is defined in terms of the above deviations (abbreviated Deviation$(I_i) = D_i$), as
\begin{equation}
    \text{Normalised Change}(D_i) \vcentcolon = \frac{D^{final}_i - D^{start}_i}{D^{final}_i + D^{start}_i}\;.
\end{equation}
This measure is bounded in the range $[-1,1]$ due to the normalisation, where a value of 0 indicates the deviation is unchanged after training such that the model is equally appropriate.
The limit of -1 indicates the deviation has decayed towards 0 and the model has become a better fit as the Gaussianity and symmetry are better expressed, whereas the limit 1 indicates the deviation has blown up and the model has become a worse fit as the PIGMM assumptions are broken.

%%%%%%%%%%%%%%%%%%%%%%%%%%%%%%%%%%%%%%%%%%%%%%%%%
\subsubsection{After Training}\label{sec:cq_after}
\begin{figure}[!t]
    \centering
    \begin{subfigure}{0.32\textwidth}
        \centering
        \includegraphics[width=0.98\textwidth]{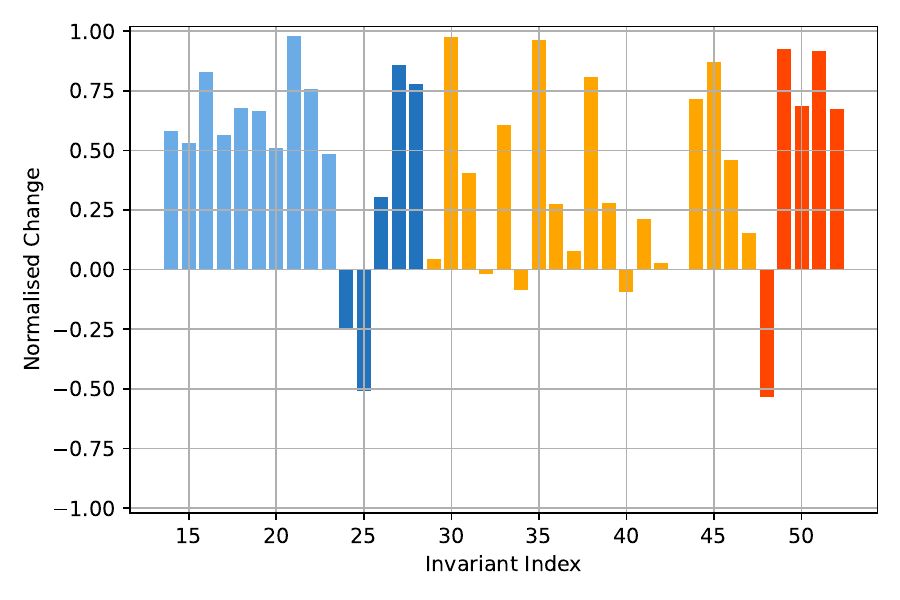}
        \caption{Gaussian L1}
    \end{subfigure} 
    \begin{subfigure}{0.32\textwidth}
        \centering
        \includegraphics[width=0.98\textwidth]{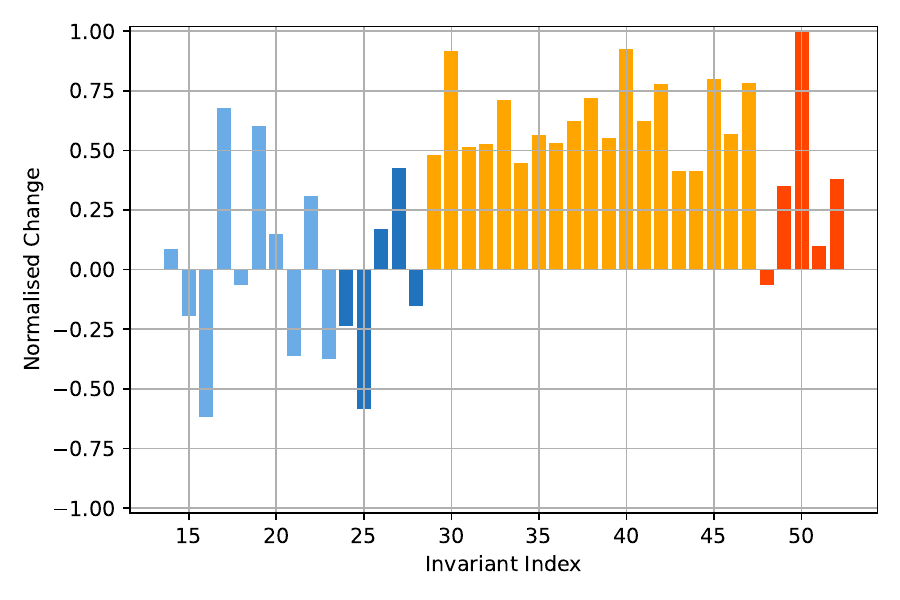}
        \caption{Gaussian L2}
    \end{subfigure}
    \begin{subfigure}{0.32\textwidth}
        \centering
        \includegraphics[width=0.98\textwidth]{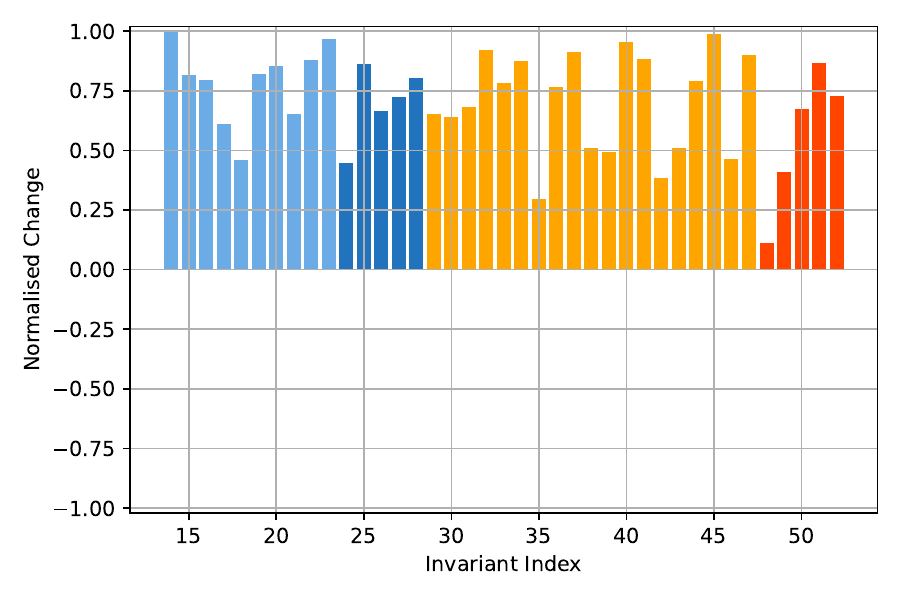}
        \caption{Gaussian L3}
    \end{subfigure}\\ 
    \begin{subfigure}{0.32\textwidth}
        \centering
        \includegraphics[width=0.98\textwidth]{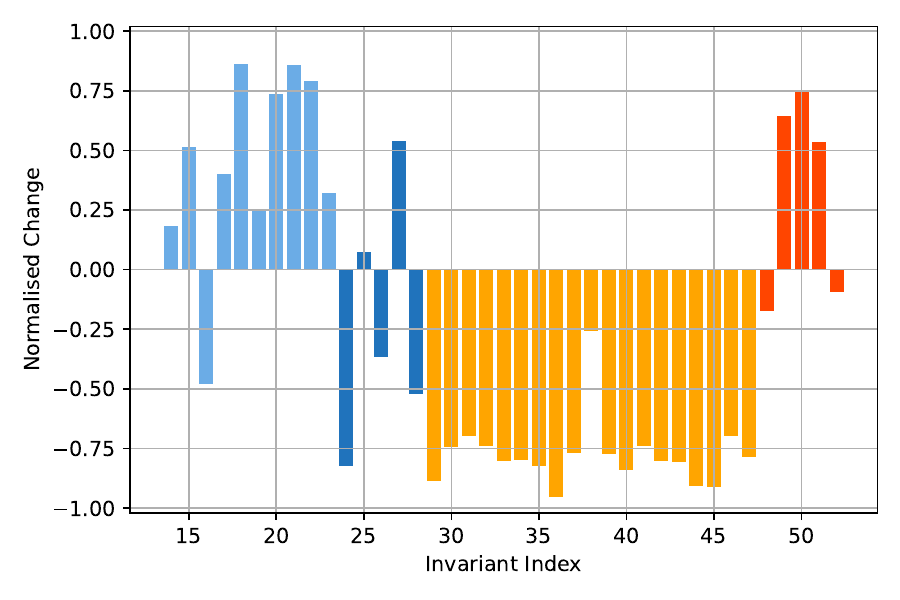}
        \caption{Uniform L1}
    \end{subfigure} 
    \begin{subfigure}{0.32\textwidth}
        \centering
        \includegraphics[width=0.98\textwidth]{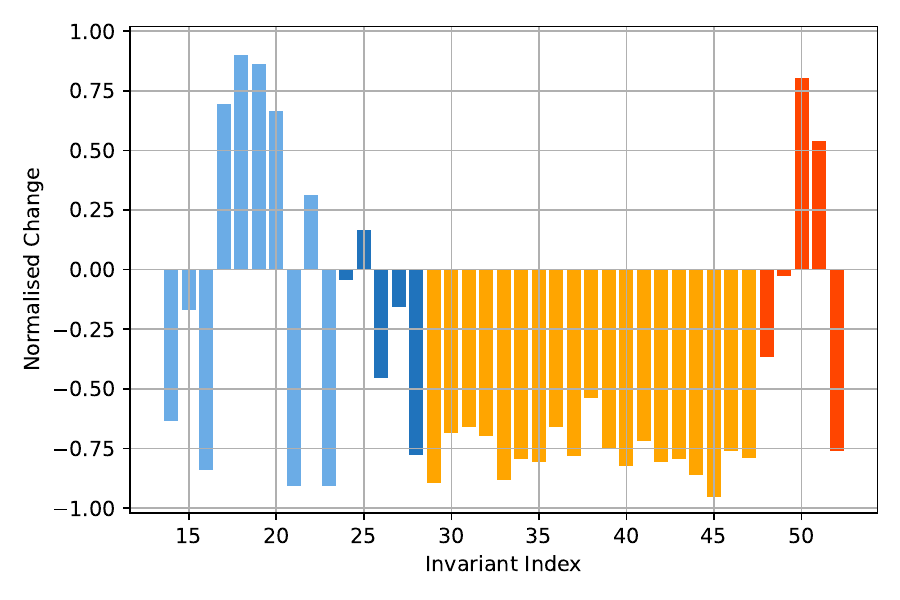}
        \caption{Uniform L2}
    \end{subfigure}
    \begin{subfigure}{0.32\textwidth}
        \centering
        \includegraphics[width=0.98\textwidth]{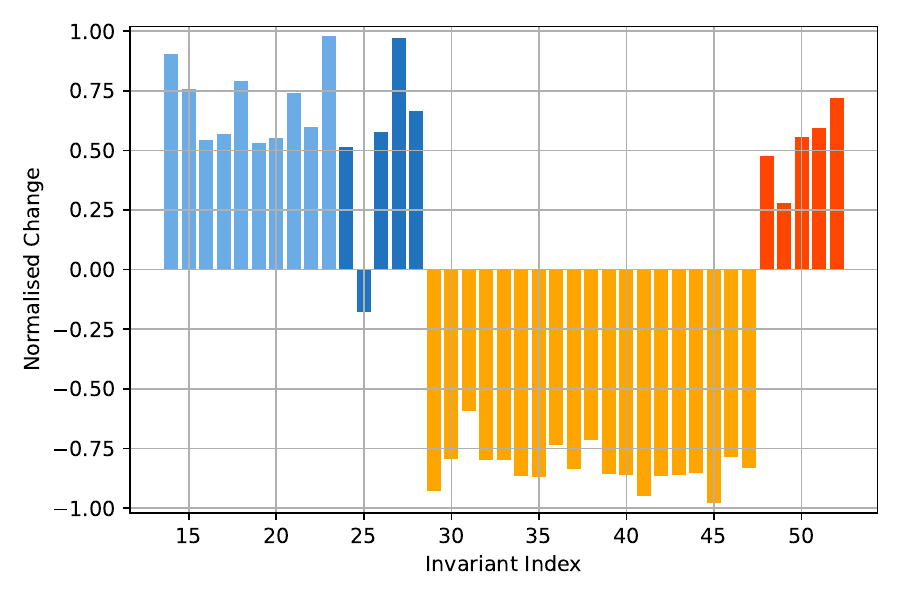}
        \caption{Uniform L3}
    \end{subfigure}
    \caption{Differences in the deviations of the low-node cubic (light blue), high-node cubic (dark blue), low-node quartic (light orange), and high-node quartic (dark orange) invariants $I_{14}-I_{52}$, between the end of training and prior to training, normalised relative to the sum of the values after and prior to training; displayed for all initialisations, and layers (L\#) considered.}
    \label{fig:normdev_startend}
\end{figure}

The plots in Figure \ref{fig:normdev_startend} show these Normalised Changes, already with some very interesting behaviour.
The first observation is that the initialisation schemes behave quite differently, the Gaussian initialisation has Normalised Changes which are predominantly positive (the mean across the invariants is positive for each layer), whereas the Uniform initialisation has predominantly negative (the mean across the invariants is negative for each layer). %(min, mean, max) for each layer [l1, l2, l3] across ivariants gauss: (min,mean,max): min [-0.53350908 -0.61781748  0.11257226], mean [0.43829613 0.34568219 0.70634198], max [0.97911192 0.99879884 0.99873734]; uniform: min [-0.95201816 -0.95234763 -0.98006032], mean [-0.24930568 -0.40398745 -0.09316495], max [0.86122807 0.89981336 0.97978392]
From this one can say that on aggregate the model becomes worse for the Gaussian initialisation over the training as the deviations blow up and the higher-orders are less representative; whereas for the Uniform initialisation the model becomes better over the training.
The quartic low-node invariants are particularity well adjusted to over the training, as the model becomes more representative for their prediction, whereas the cubic low-node are particularly poorly adjusted to.
Across the layers the last layer has the largest mean and the largest absolute mean, supporting the idea that the deviation from the PIGMM best fit of the weight matrix distributions is most changed over the training for deeper layers.

This is perhaps expected since the initialisation distributions are fitted to the simplest Gaussian PIGMM and the results in the previous sections show that fit is good at the start of training.
As the PIGMM model becomes worse after the training this indicates that at least one of the assumptions, either the Gaussianity or the diagonal permutation symmetry is broken by the training process.

%%%%%%%%%%%%%%%%%%%%%%%%%%%%%%%%%%%%%%%%%%%%%%%%%
\subsubsection{Throughout Training}\label{sec:cq_throughout}
The normalised changes in the previous section establish that the PIGMM fits are less representative after training than at the initialisation, yet interestingly their appropriateness for predicting invariants of different orders and node numbers is quite different despite consistency within each of these invariant sets.

\begin{figure}[!t]
    \centering
    \begin{subfigure}{0.32\textwidth}
        \centering
        \includegraphics[width=0.98\textwidth]{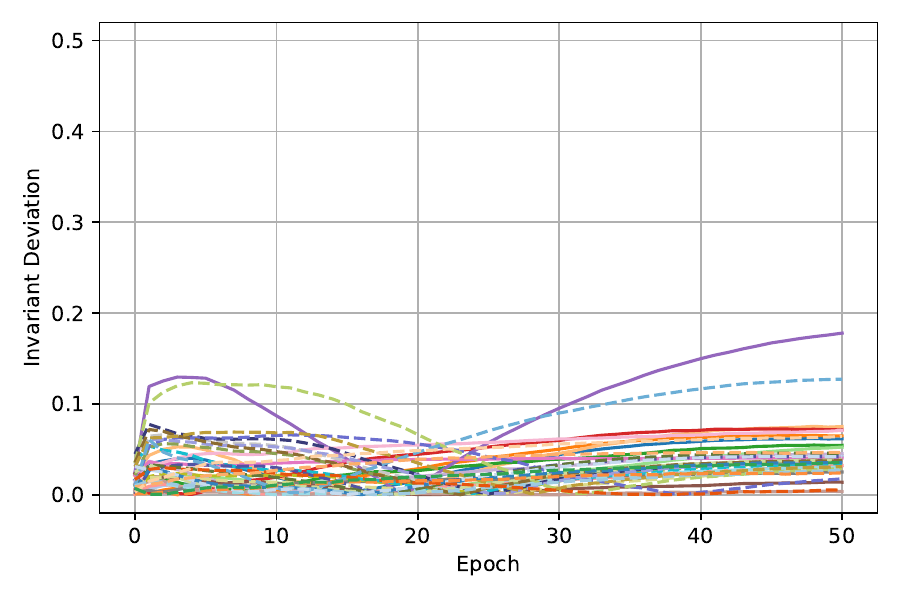}
        \caption{Gaussian L1}
    \end{subfigure} 
    \begin{subfigure}{0.32\textwidth}
        \centering
        \includegraphics[width=0.98\textwidth]{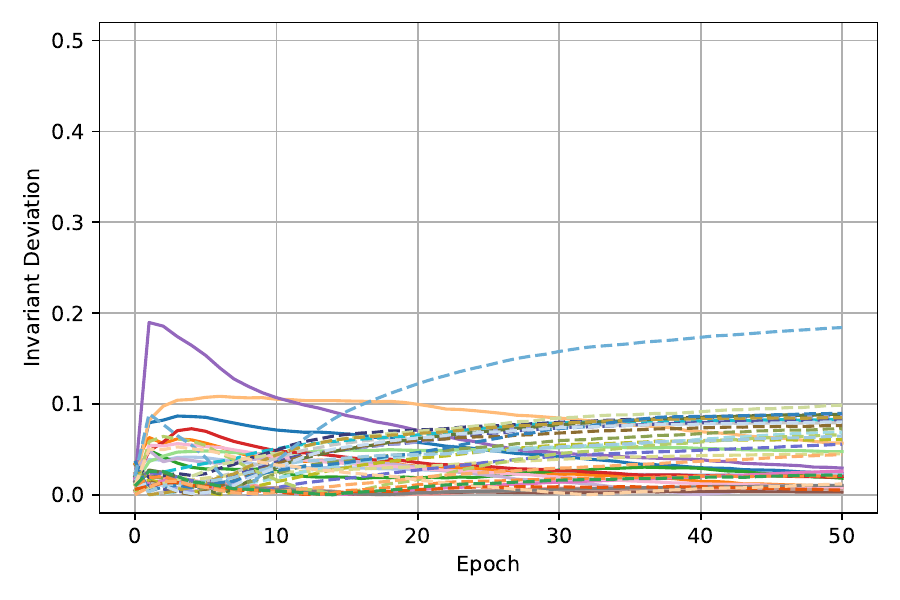}
        \caption{Gaussian L2}
    \end{subfigure}
    \begin{subfigure}{0.32\textwidth}
        \centering
        \includegraphics[width=0.98\textwidth]{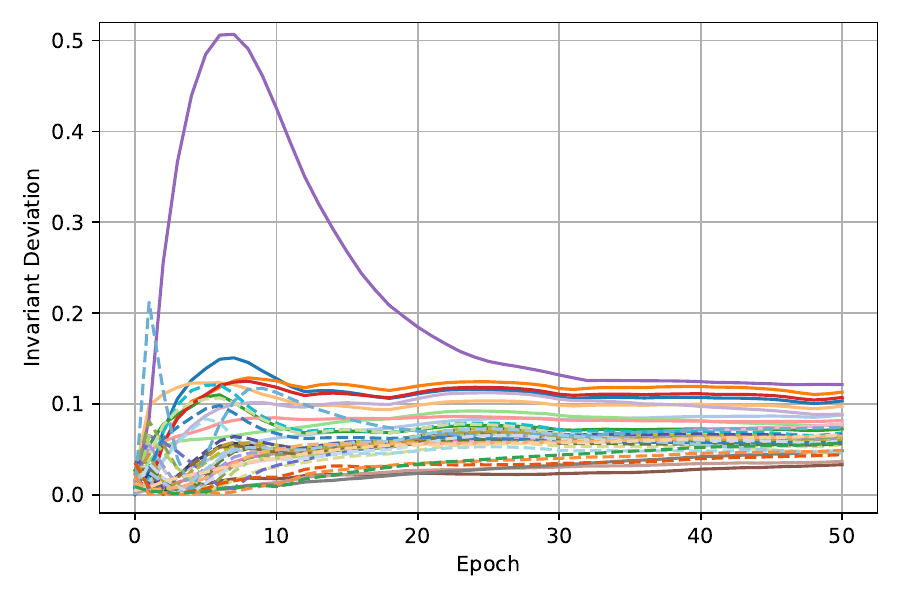}
        \caption{Gaussian L3}
    \end{subfigure}\\ 
    \begin{subfigure}{0.32\textwidth}
        \centering
        \includegraphics[width=0.98\textwidth]{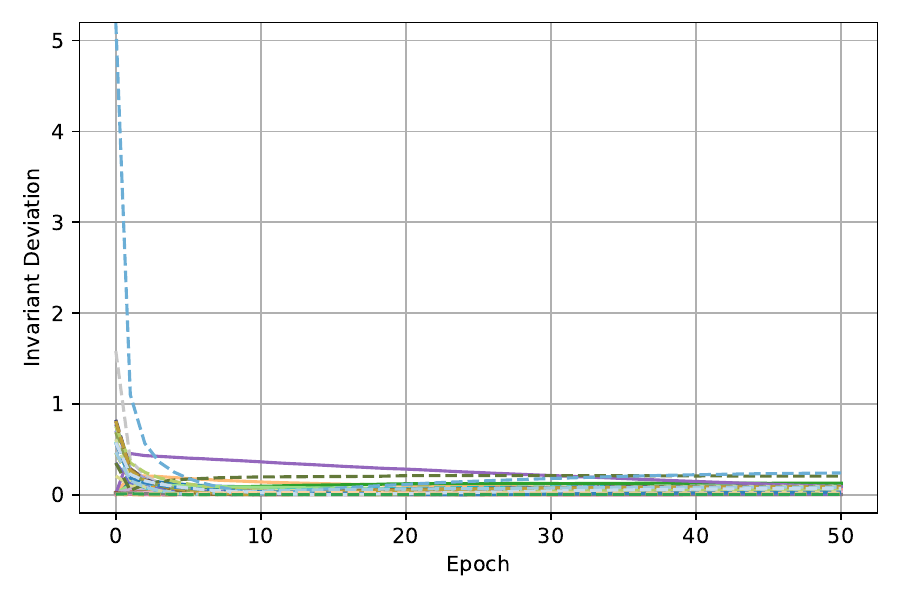}
        \caption{Uniform L1}
    \end{subfigure} 
    \begin{subfigure}{0.32\textwidth}
        \centering
        \includegraphics[width=0.98\textwidth]{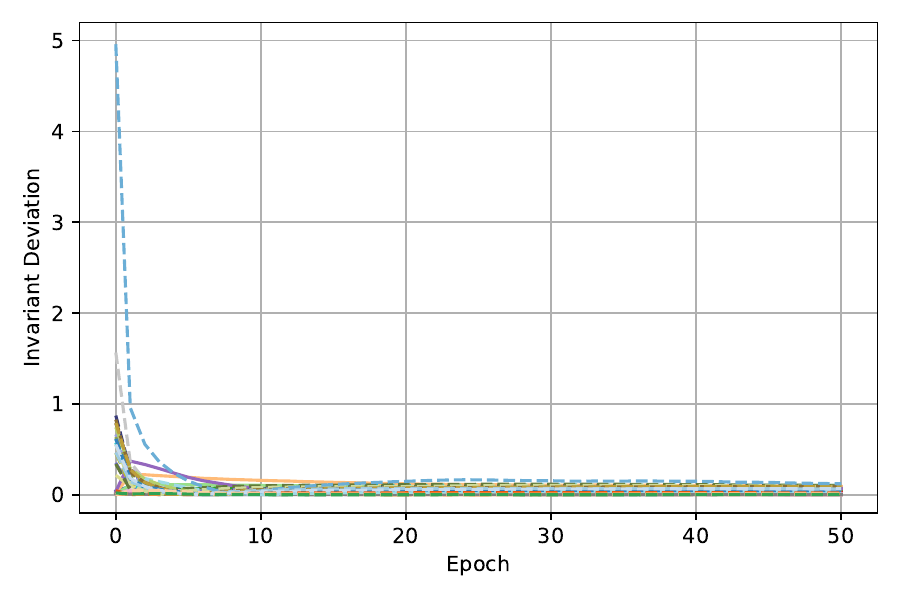}
        \caption{Uniform L2}
    \end{subfigure}
    \begin{subfigure}{0.32\textwidth}
        \centering
        \includegraphics[width=0.98\textwidth]{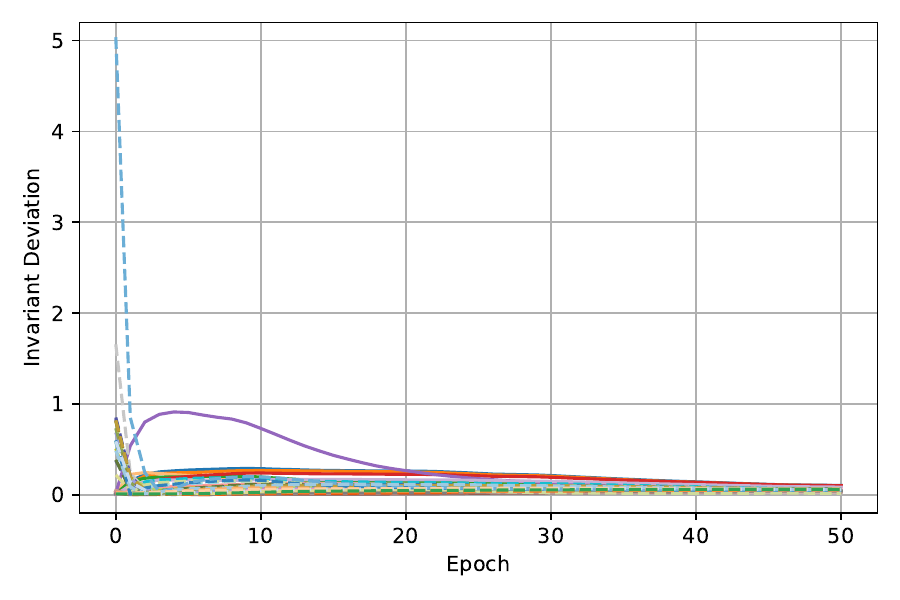}
        \caption{Uniform L3}
    \end{subfigure}\\ 
    \begin{subfigure}{0.98\textwidth}
        \centering
        \includegraphics[width=0.98\textwidth]{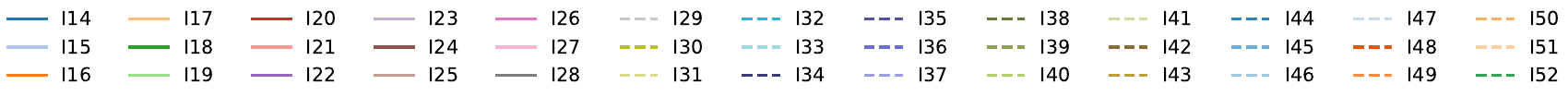}
        \caption{Legend}
    \end{subfigure}
    \caption{Variation of the cubic deviations (solid lines) and quartic deviations (dashed lines), labelled respectively by their invariants $I_{14}-I_{52}$, across the 50 epochs of training. The legend is the same throughout and collectively shown at the bottom (g) for readability. Note the y-axes scales are fixed within initialisations, but differ between them.}
    \label{fig:CQ_during}
\end{figure}

To assess how these deviations of the higher-order invariants, as defined in \eqref{eq:CQinv_deviation}, change throughout the training process, Figure \ref{fig:CQ_during} shows the values of each of these 39 deviations for each training epoch.
Importantly, compared to the normalised changes, these plots show the scale of the deviations (not just relative to the start values), allowing deduction of the extent to which the PIGMM model fits well to the weight matrix ensembles.

The deviations of the higher-order invariants in Figure \ref{fig:CQ_during} show relatively smooth behaviour, and significantly, they show that after training all the deviation values are $<<1$ and therefore that the fitted PIGMM models are accurate predictors of the higher-order invariants, and then finally that the Gaussianity and symmetry assumptions of the PIGMMs are representative of the trained weight matrix data.
The PIGMMs are \textit{good} models for trained weight matrix distributions.

Delving deeper into the behaviour between ensembles, for each layer the uniform initialisation has one invariant which starts with an especially high deviation, this is $I_{45}=\sum_{i,j}W_{ij}^4$.
Respectively for the Gaussian initialisation the invariant $I_{22} = \sum_{i,j}W_{ij}^3$ is the most turbulent (yet still with values well below 1).
Interestingly, both these dominantly deviating invariants match in style the $I_3=\sum_{i,j}W_{ij}^2$ dominant deviations in Figure \ref{fig:LQ_during}. %why? the independent and balanced nature of this invariant style means something?
However as the weights update in training even these invariants quickly becomes well predicted by the PIGMMs for later ensembles.
Otherwise the remaining invariant deviations are a similar scale between the initialisations indicating the PIGMM models are consistently good at modelling weight matrix distributions (especially after training) independently of the chosen initialisation scheme.
Conversely, between layers the deeper layers have larger average deviation values, meaning the PIGMMs are more representative for earlier layers.

\newpage
%%%%%%%%%%%%%%%%%%%%%%%%%%%%%%%%%%%%%%%%%%%%%%%%%
\subsection{Wasserstein Distance}\label{sec:wasserstein}
Using the PIGMM formalism, previous sections have used a change in individual invariants and model parameters as measures of the changing weight distribution through training.
To combine these measures into a single distance metric on the model space of PIGMM parameters, in this section we specialise  the Wasserstein distance on multi-variate Gaussians to PIGMMs and use this  as another measure to quantify these changes.

The Wasserstein distance \cite{wasserstein}, $\mathfrak{d}^2$, is a standard symmetric measure for the distance between two general probability distributions on the same domain, where the form for standard multi-variate Gaussians is known.
Adapting to a pair of  PIGMMs with parameters $\{\tilde{\mu}_1, \tilde{\mu}_2, \Sigma = [\Lambda_{V_0}, \Lambda_{V_H}, \Lambda_{V_2}, \Lambda_{V_3}]\}$ and $\{\tilde{\mu}_1', \tilde{\mu}_2', \Sigma' = [\Lambda_{V_0}', \Lambda_{V_H}', \Lambda_{V_2}', \Lambda_{V_3}']\}$ it is 
\begin{equation}
    \mathfrak{d}^2 = ( \widetilde{\mu}_1 - \widetilde{\mu}_1')^2 + ( \widetilde{\mu}_2 - \widetilde{\mu}_2')^2 
    + \text{Tr} \Sigma + \text{Tr} \Sigma' - 2 \text{Tr}  ( \Sigma^{1/2}  \Sigma' \Sigma^{1/2} )^{1/2} \;,
\end{equation}
where the derivation and further details on the functional form of the trace ($\text{Tr}(\cdot)$) terms are given in §\ref{app:wasserstein_derivation}. 
A \texttt{python} function for the computation of the Wasserstein distance between general PIGMMs is made  available with this paper's \href{https://github.com/edhirst/PIGWMM}{\texttt{GitHub}} repository.

\begin{figure}[!t]
    \centering
    \begin{subfigure}{0.48\textwidth}
        \centering
        \includegraphics[width=0.98\textwidth]{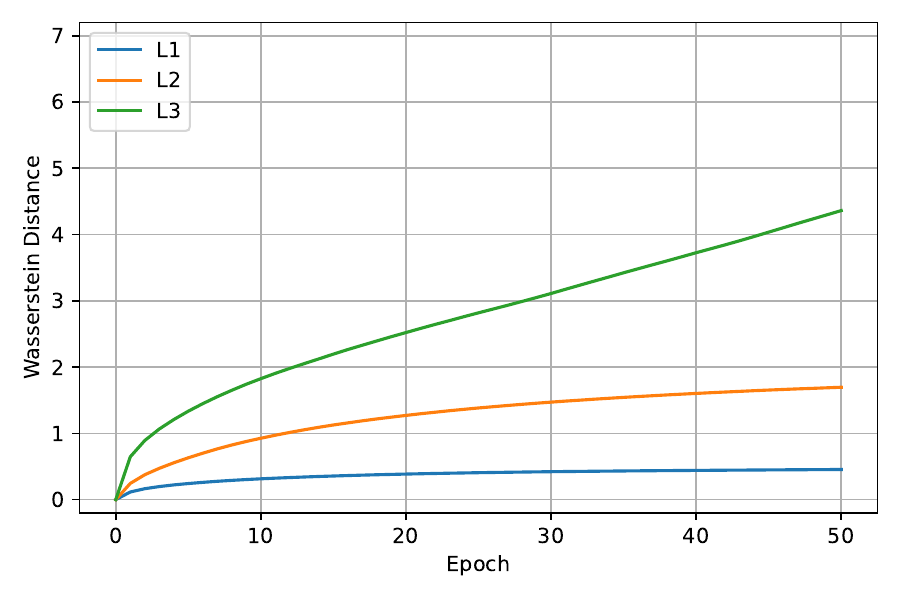}
        \caption{Gaussian}
    \end{subfigure} 
    \begin{subfigure}{0.48\textwidth}
        \centering
        \includegraphics[width=0.98\textwidth]{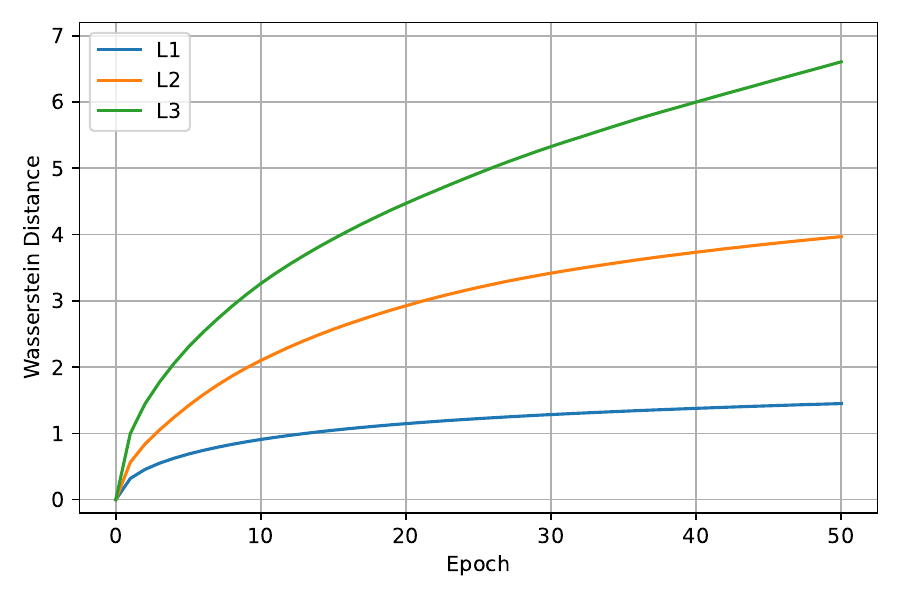}
        \caption{Uniform}
    \end{subfigure}
    \caption{Computations of the Wasserstein distance between the simple Gaussian model for each initialisation and the permutation-invariant Gaussian matrix models at each epoch (based on the observed model parameters), across the 3 layers.}
    \label{fig:wasserstein}
\end{figure}

The PIGMM model parameters $(f_i)$ for both the simple Gaussian model and the fitted PIGMMs for each weight matrix ensemble can be repackaged into the $\Sigma$ format as described in §\ref{sec:gmms}.
The distance in model space of the fitted PIGMMs from the simple Gaussian model\footnote{Note each initialisation distribution has its own (different) simple Gaussian model, here structured as a PIGMM, with parameter values as given in \eqref{eq:special-param} \& \eqref{eq:special-param-uniform}.} is then computed for each of the 306 ensembles across the initialisations, layers, and training epochs.

Plots of these Wasserstein distances are shown in Figure \ref{fig:wasserstein}.
Both plots, show that through training the best fitted PIGMM moves smoothly further away from the simple Gaussian over training, and for each layer the uniform initialisation moves further, expressed by larger Wasserstein distances.
In all cases the rate of movement away from the simple Gaussian slows during the training, indicating the training initially causes large changes to the weight distribution as the best fit PIGMM moves quickly away in the model space, but then settles later in the training.

Between layers, for both initialisations, the earlier layers have lower Wasserstein distances, staying closer to the simple Gaussian, such that the training process changes the weight distributions for later layers more significantly.

%%%%%%%%%%%%%%%%%%%%%%%%%%%%%%%%%%%%%%%%%%%%%%%%%
\section{Impact of Architectural Modifications on Gaussianity}\label{sec:results2}
In the previous section the theoretical and the experimental ensemble averages of permutation invariant polynomials, i.e. the theoretical and experimental observables, gave evidence that the PIGMM provide a parsimonious model for the evolution of multi-variate Gaussianity of the weight matrices through training.  
In this section, two standard variations of the architecture are considered: introducing regularisation; and the limit of increasing layer width. 
We find that the validity of the PIGMM as a description of approximate Gaussianity is robust under these variations and the graph-theoretic observables of the model provide an interpretable framework for features of the two variations. 

The first subsection below,  §\ref{sec:regularisation}, involves the generation of another 153 ensembles for the Gaussian initialisation, but now running the training with L2 regularisation, providing insight into the effects of regularisation on the fitting of the PIGMMs.
In the second subsection, §\ref{sec:asymptotic}, the results again involve the generation of new ensembles, 204 of them, for a modified architecture with focus on a single pair of hidden layers (with one weight matrix between them) that vary in size, choosing 4 sizes to consider increasing from 10 to 640 and giving insight into the effects of the large width limit of NNs on the fitting of these PIGMMs.

%%%%%%%%%%%%%%%%%%%%%%%%%%%%%%%%%%%%%%%%%%%%%%%%%
\subsection{Effects of Regularisation}\label{sec:regularisation}
It is often common practise in supervised learning approaches to problems to introduce a regularisation term into the loss.
This is minimised in the limit of all NN parameters taking the value of zero, and hence encourages the optimiser to find a solution to the problem with parameter values minimised.

These terms often take the form of a $L^n$ norms, particularly where $n \in \{1,2\}$.
When regularisation is applied the full loss landscape changes in accordance with this term, which in turn changes the size and locations of the model space regions which represent good solutions to the problem. 
Therefore, adding regularisation will directly effect the weight distributions of trained solutions, and how best they should be modelled.

To test how introduction of a $L^2 = \sum_{i,j} W_{ij}^2$ regularisation term\footnote{It is interesting to note that this regularisation term takes the same form as the $I_3$ quadratic PIGMM invariant.} affects the weight distribution of the trained solutions, and the suitability of the PIGMM formalism, another 153 ensembles of the weight matrices were generated by training the same NN architecture on the same MNIST problem, with Gaussian initialisation, using the same hyperparameters except with regularisation turned on (here we use a loss weighting factor of 0.01 for the regularisation term).
The average final accuracy performance score across the 1000 runs was $0.9068 \pm 0.0011$, a comparable yet marginally better score than without regularisation as shown in Table \ref{tab:accuracies}. %$0.906836 \pm 0.0011184172316269097$

A priori, any regularisation term is expected to minimise NN parameter values, setting any redundant ones to zero.
This would then naively lead to less weights that need to be modelled, and hence a PIGMM with the same number of degrees of freedom would be expected to perform better at modelling the distributions of trained weights.

\begin{figure}[!t]
    \centering
    \begin{subfigure}{0.32\textwidth}
        \centering
        \includegraphics[width=0.98\textwidth]{Figures/Deviations/MG_Deviations_L1.pdf}
        \caption{Unregularised L1}
    \end{subfigure} 
    \begin{subfigure}{0.32\textwidth}
        \centering
        \includegraphics[width=0.98\textwidth]{Figures/Deviations/MG_Deviations_L2.pdf}
        \caption{Unregularised L2}
    \end{subfigure}
    \begin{subfigure}{0.32\textwidth}
        \centering
        \includegraphics[width=0.98\textwidth]{Figures/Deviations/MG_Deviations_L3.pdf}
        \caption{Unregularised L3}
    \end{subfigure}\\ 
    \begin{subfigure}{0.32\textwidth}
        \centering
        \includegraphics[width=0.98\textwidth]{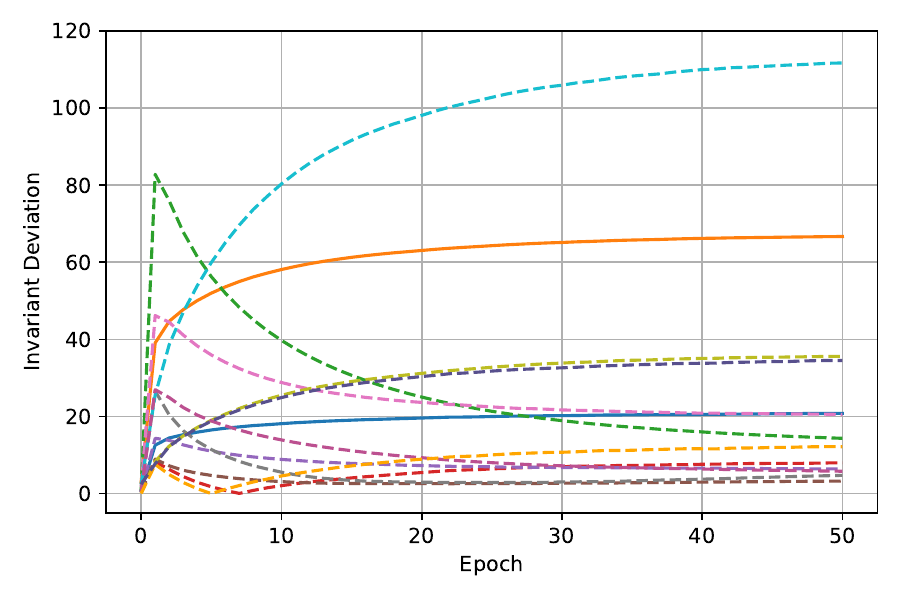}
        \caption{Regularised L1}
    \end{subfigure} 
    \begin{subfigure}{0.32\textwidth}
        \centering
        \includegraphics[width=0.98\textwidth]{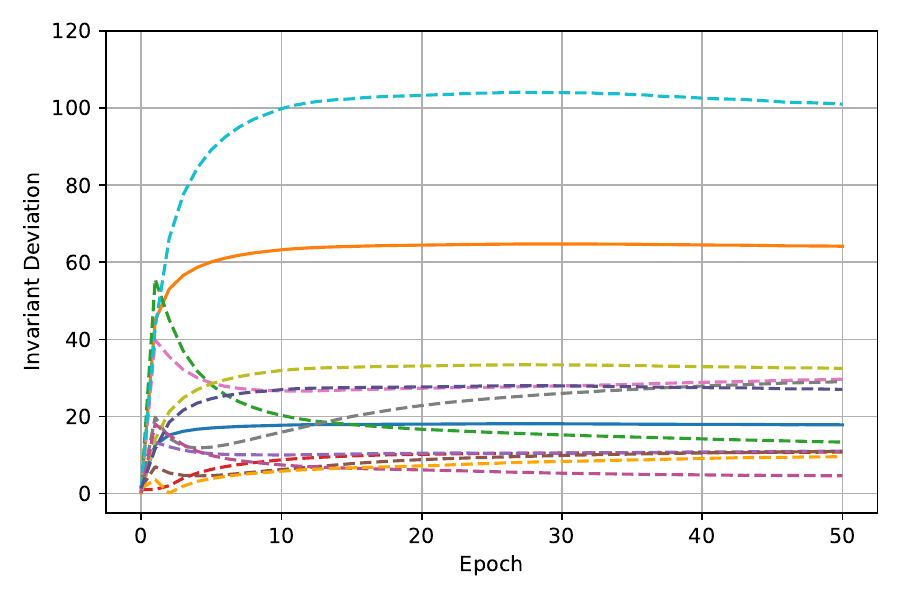}
        \caption{Regularised L2}
    \end{subfigure}
    \begin{subfigure}{0.32\textwidth}
        \centering
        \includegraphics[width=0.98\textwidth]{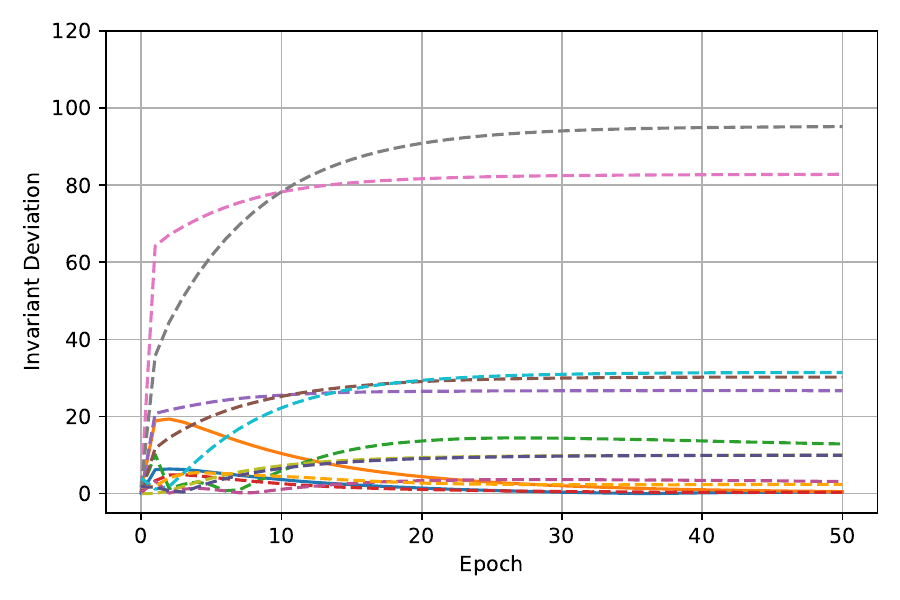}
        \caption{Regularised L3}
    \end{subfigure}\\
    \begin{subfigure}{0.98\textwidth}
        \centering
        \includegraphics[width=0.98\textwidth]{Figures/Deviations/LQ_Legend.pdf}
        \caption{Legend}
    \end{subfigure}
    \caption{Comparison between unregularised (a-c) and regularised (d-f) training schemes for the Gaussian initialised neural networks; showing variation of the linear invariant deviations (solid lines) and quadratic invariant deviations (dashed lines), labelled respectively by their invariants $I_1-I_{13}$, across the 50 epochs of training. The legend is the same throughout and collectively shown at the bottom (g) for readability. We emphasise the varying scales in y-axis.}
    \label{fig:LQr_during}
\end{figure}

Equivalent plots of the linear and quadratic invariant deviations are shown in Figure \ref{fig:LQr_during}, including repetition of the unregularised plots from Figure \ref{fig:LQ_during} for ease of comparison.
Firstly, the introduction of regularisation causes the deviation scales to become consistent across the layers (each layer's plot has the same y-axis scale).
Additionally, the most deviated parameters are now different, instead of $I_3=\sum_{i,j}W_{ij}^2$, it is now $I_{10}=\sum_{i,j,k,l}W_{ij}W_{kl}$ for layers 1 \& 2, and $I_8=\sum_{i,j,k}W_{ij}W_{kj}$ for layer 3.
Importantly, the monotonically increasing deviation observed in the unregularised cases does not happen in the regularised case, many invariants deviate from the simple Gaussian a lot early in training and then approach back towards zero.
A particular example of this is the $I_3$ invariant, whose functional form is the same as the regularisation term, therefore as the optimiser updates weights to minimise this regularisation part of the loss it is also reducing this invariants deviation from the simple Gaussian, which presumably has an especially low value of this invariant for the PIGMM class (note its expectation value for the simple Gaussian is non-zero as computed for Table \ref{tab:inv_analytic}).

\begin{figure}[!t]
    \centering
    \begin{subfigure}{0.32\textwidth}
        \centering
        \includegraphics[width=0.98\textwidth]{Figures/LHDeviations/MG_LHDeviations_L1.pdf}
        \caption{Unregularised L1}
    \end{subfigure} 
    \begin{subfigure}{0.32\textwidth}
        \centering
        \includegraphics[width=0.98\textwidth]{Figures/LHDeviations/MG_LHDeviations_L2.pdf}
        \caption{Unregularised L2}
    \end{subfigure}
    \begin{subfigure}{0.32\textwidth}
        \centering
        \includegraphics[width=0.98\textwidth]{Figures/LHDeviations/MG_LHDeviations_L3.pdf}
        \caption{Unregularised L3}
    \end{subfigure}\\ 
    \begin{subfigure}{0.32\textwidth}
        \centering
        \includegraphics[width=0.98\textwidth]{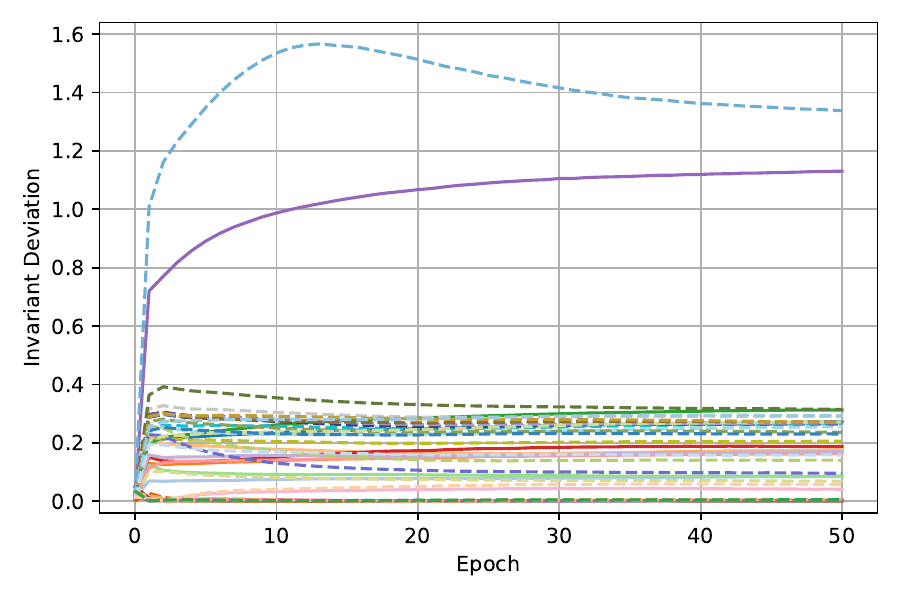}
        \caption{Regularised L1}
    \end{subfigure} 
    \begin{subfigure}{0.32\textwidth}
        \centering
        \includegraphics[width=0.98\textwidth]{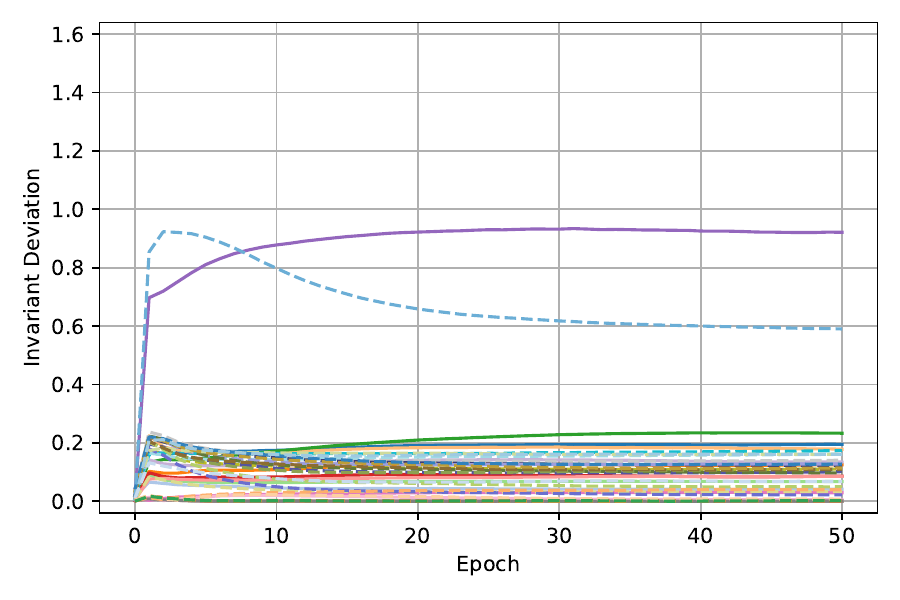}
        \caption{Regularised L2}
    \end{subfigure}
    \begin{subfigure}{0.32\textwidth}
        \centering
        \includegraphics[width=0.98\textwidth]{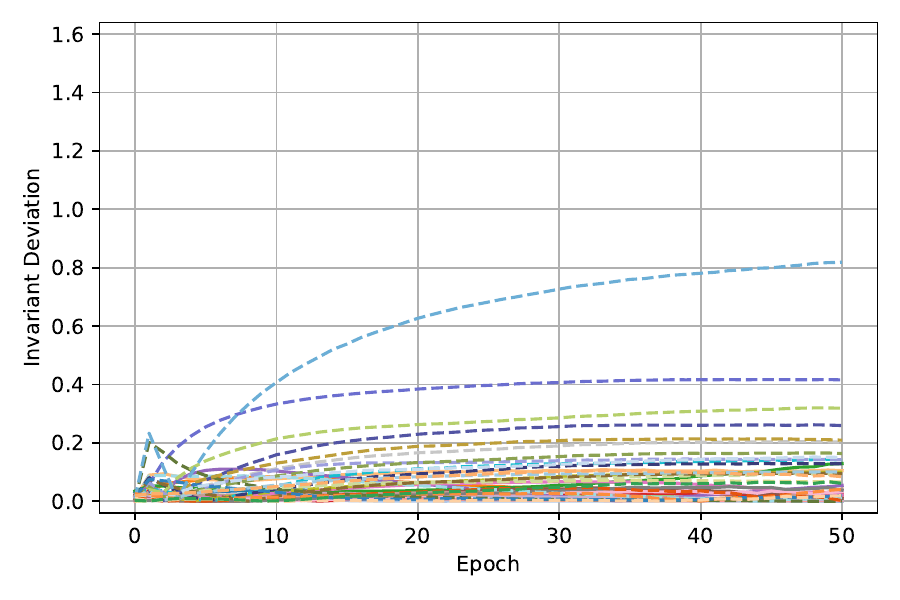}
        \caption{Regularised L3}
    \end{subfigure}\\ 
    \begin{subfigure}{0.98\textwidth}
        \centering
        \includegraphics[width=0.98\textwidth]{Figures/LHDeviations/CQ_Legend.pdf}
        \caption{Legend}
    \end{subfigure}
    \caption{Comparison between unregularised (a-c) and regularised (d-f) training schemes for the Gaussian initialised neural networks; showing variation of the cubic deviations (solid lines) and quartic deviations (dashed lines), labelled respectively by their invariants $I_{14}-I_{52}$, across the 50 epochs of training. The legend is the same throughout and collectively shown at the bottom (g) for readability. Note the y-axes scales are fixed within schemes, but differ between them.}
    \label{fig:CQr_during}
\end{figure}

Likewise, equivalent plots for the cubic and quartic deviations are shown in Figure \ref{fig:CQr_during}, again including repetition of the unregularised plots from Figure \ref{fig:CQ_during} for ease of comparison.
With regularisation, the PIGMMs predict higher-order deviations still predominantly $<<1$, and are therefore still good models for these weight distributions.
Comparatively, with regularisation, the deviations are marginally higher, as the models become marginally less suitable.
Furthermore, the deviation scale difference is more consistent between layers, but interestingly the most deviated become significantly more deviated.
Across the plots these most deviated invariants are the $I_{22} = \sum_{i,j}W_{ij}^3$ \& $I_{45} = \sum_{i,j}W_{ij}^4$ invariants.
Which were also the most deviant without regularisation but now deviate significantly more, with some deviation values exceeding 1.
These observations would indicate that the PIGMMs are still good models for weight matrix ensembles under regularisation, but are slightly worse than without regularisation.
This is likely associated to arguments where regularisation terms break symmetry \cite{Hashimoto:2024rms}, and here these results indicate this idea looks to extend into permutation symmetry causing the symmetric assumptions of the PIGMMs to be less appropriate.

\begin{figure}[!t]
    \centering
    \begin{subfigure}{0.48\textwidth}
        \centering
        \includegraphics[width=0.98\textwidth]{Figures/Wasserstein/WassersteinSimple_MG.pdf}
        \caption{Unregularised}
    \end{subfigure} 
    \begin{subfigure}{0.48\textwidth}
        \centering
        \includegraphics[width=0.98\textwidth]{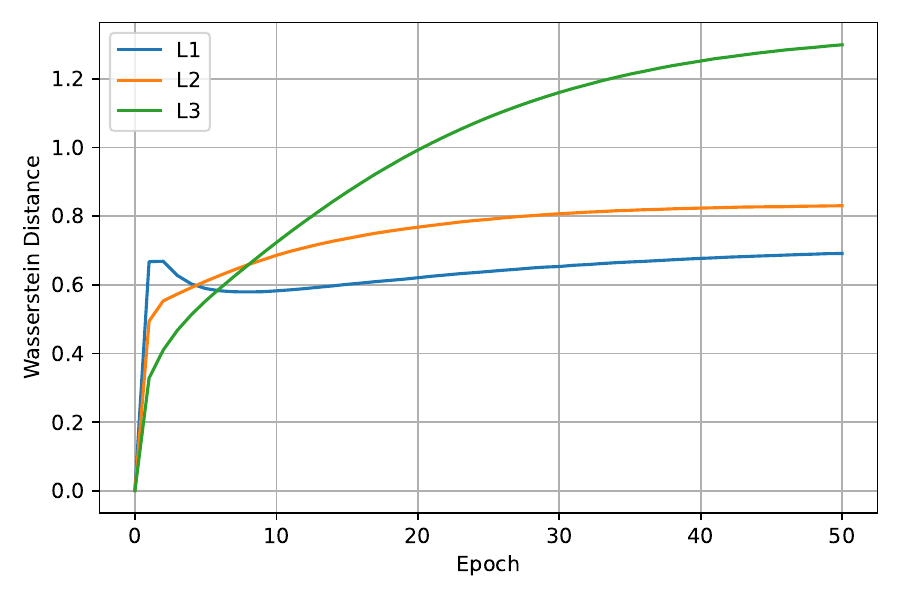}
        \caption{Regularised}
    \end{subfigure}
    \caption{Computations of the Wasserstein distance between the simple Gaussian model for the Gaussian initialisation and the permutation-invariant Gaussian matrix models at each epoch (based on the observed model parameters); shown for the (a) unregularised and (b) regularised training schemes, across the 3 layers.}
    \label{fig:wasserstein_regularised}
\end{figure}

Finally, the same Wasserstein distance measure for PIGMMs (as defined in §\ref{sec:wasserstein}) was applied to the fitted PIGMMs for each weight matrix ensemble throughout training.
A plot of these Wasserstein distances to the simplest Gaussian model are shown in Figure \ref{fig:wasserstein_regularised}, again with the unregularised plot from Figure \ref{fig:wasserstein} repeated for ease of comparison.

This plot shows that with regularisation the PIGMM best fit model for the later layers is closer to the simplest Gaussian model (shown by a smaller y-axis scale).
The behaviour is less consistent between layers, particularly where early in training the earlier layers change the most, which is opposite to the unregularised case, despite later in training the ordering flipping back to deeper layers further away.
Part of this flipping demonstrates that the movement is now non-monotonic for the first layer under regularisation, as at some point in the training the weight distribution shifts back in style towards the simple Gaussian which the best fit PIGMM expresses via a smaller Wasserstein distance. % L1 seems to be set first then L2 then L3, highlighting different learning dynamics 

%%%%%%%%%%%%%%%%%%%%%%%%%%%%%%%%%%%%%%%%%%%%%%%%%
\subsection{Asymptotic Width Limit}\label{sec:asymptotic}
Prior investigations in this work examine a deep architecture with multiple layers of weight matrices, allowing these investigations to make comments on the changing suitability of PIGMMs on weight matrix ensembles at varying depth.
The large depth limit of NNs exhibits unpredictable behaviour with varying degeneracies and vanishing/exploding gradients \cite{largedepthnns}.
However the converse large width limit is much more tractable, where layer pre-activations converge under the Central Limit Theorem to Gaussian distributions and the whole NN towards a Gaussian process \cite{largewidthnns}.

Therefore, as a first step towards probing this alternative limit of large width, in this section, a different yet related architecture was trained on the MNIST problem with the same training hyperparameters.
The architecture here is different in that instead of layer sizes $\{d, 10, 10, 10\}$, they are now $\{d, \alpha, \alpha, 10\}$.
This new architecture then has one central square weight matrix which can be analysed through this PIGMM formalism, of shape $W_{\alpha, \alpha}$.
204 new weight matrix ensembles were then trained, again for 1000 runs, for the Gaussian initialisation scheme but now for 4 different values of $\alpha$.
These values were chosen to follow the function $10 \times 2^{2\alpha}$, using values of $\alpha \in [0,1,2,3]$, producing layer sizes $[10, 40, 160, 640]$.
The $\alpha$ range was selected to start with $10$, such that connection to previous results could be made, but then increase up to a large yet still computationally feasible limit.
The maximum value of 640 should be significantly large enough such that comments about the $\alpha \longmapsto \infty$ limit can be made.

\begin{table}[!t]
\centering
\begin{tabular}{|c|c|}
\hline
\begin{tabular}[c]{@{}c@{}}Layer Size\\ $\alpha$\end{tabular} & \begin{tabular}[c]{@{}c@{}}Average\\ Accuracy\end{tabular} \\ \hline
10 & $ 0.8886 \pm 0.0023 $ \\ \hline %0.888623 0.0023171838233079396
40 & $ 0.9666 \pm 0.0001 $ \\ \hline %0.966613 8.13832353743694e-05
160 & $ 0.9779 \pm 0.0001 $ \\ \hline %0.9778669999999999 7.189792069316059e-05
640 & $ 0.9796 \pm 0.0022 $ \\ \hline %0.979644 0.002186610161871569
\end{tabular}
\caption{Final accuracies of the trained NN models for increasing intermediate layer sizes $\alpha$, after the 50 epochs of training for the
MNIST classification investigation and Gaussian initialisation. Accuracies are averaged over the 1000 runs with standard error reported also.}
\label{tab:asymp_accuracies}
\end{table}

The average final accuracy performance scores across the 1000 runs for each value of $\alpha$ are shown in Table \ref{tab:asymp_accuracies}.
They are all sufficiently high to indicate good learning, and in the large
$\alpha$ limit express increasingly better performances.

\begin{figure}[!t]
    \centering
    \begin{subfigure}{0.24\textwidth}
        \centering
        \includegraphics[width=0.98\textwidth]{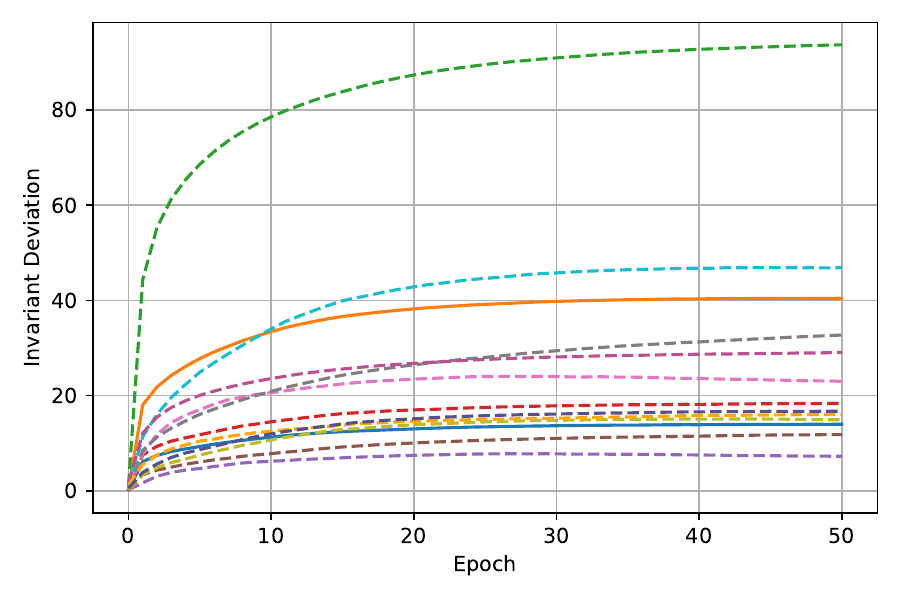}
        \caption{$\alpha=10$}
    \end{subfigure} 
    \begin{subfigure}{0.24\textwidth}
        \centering
        \includegraphics[width=0.98\textwidth]{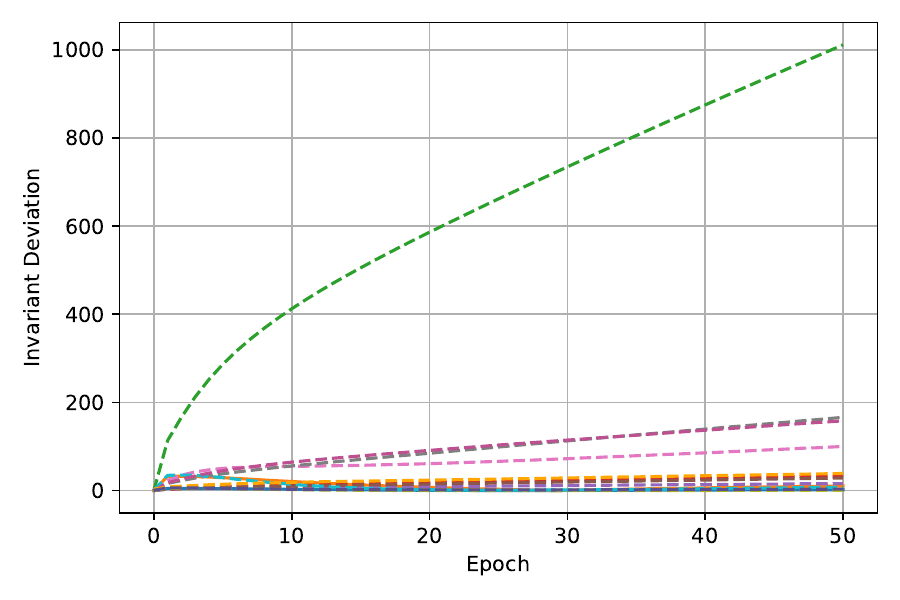}
        \caption{$\alpha=40$}
    \end{subfigure}
    \begin{subfigure}{0.24\textwidth}
        \centering
        \includegraphics[width=0.98\textwidth]{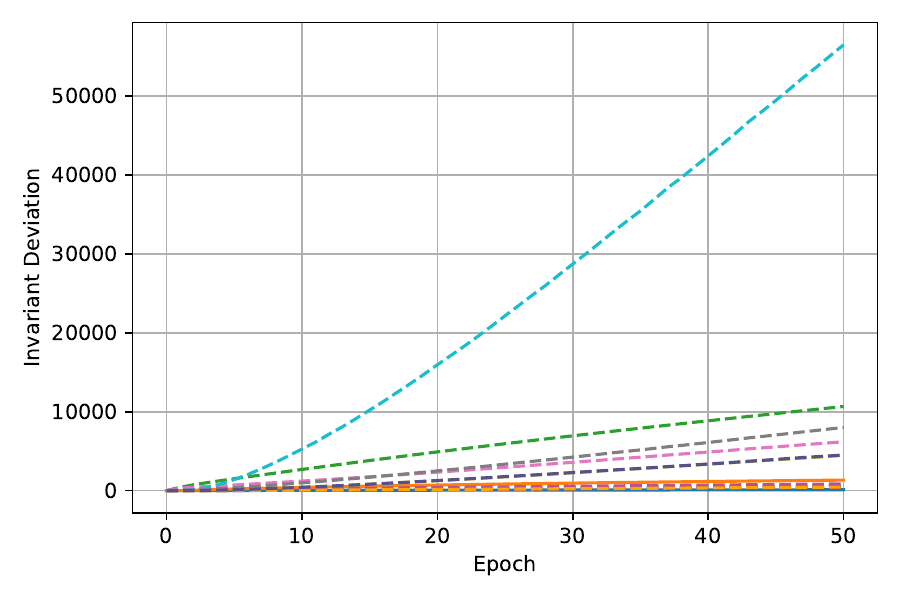}
        \caption{$\alpha=160$}
    \end{subfigure}
    \begin{subfigure}{0.24\textwidth}
        \centering
        \includegraphics[width=0.98\textwidth]{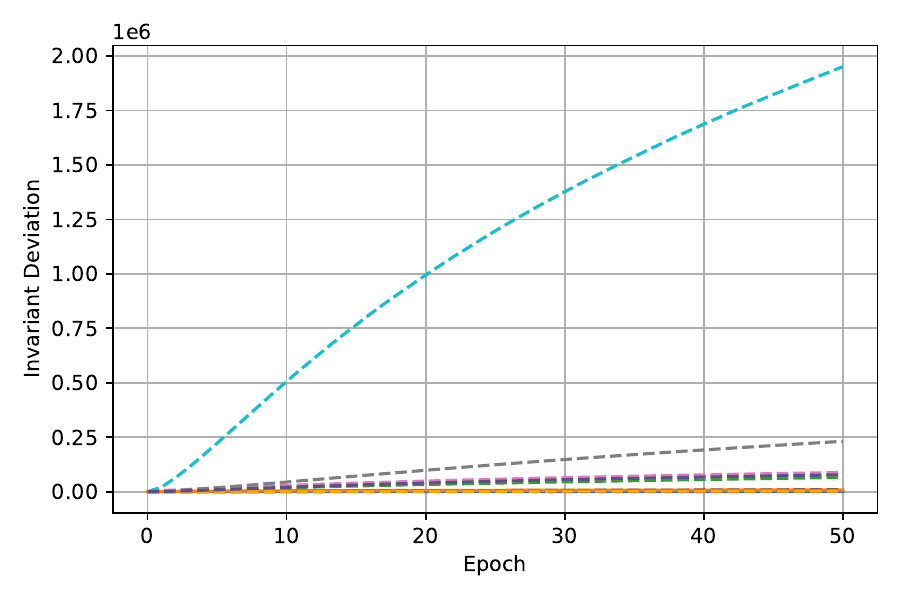}
        \caption{$\alpha=640$}
    \end{subfigure}\\ 
    \begin{subfigure}{0.98\textwidth}
        \centering
        \includegraphics[width=0.98\textwidth]{Figures/Deviations/LQ_Legend.pdf}
        \caption{Legend}
    \end{subfigure}
    \caption{Variation of the linear invariant deviations (solid lines) and quadratic invariant deviations (dashed lines), labelled respectively by their invariants $I_1-I_{13}$, across the 50 epochs of training. Displayed for NNs of increasing width with $\alpha$ neurons per central layer. The legend is the same throughout and collectively shown at the bottom (e) for readability. We emphasise the varying scales in y-axis.}
    \label{fig:LQasymp_during}
\end{figure}

Equivalent deviation plots for the linear and quadratic invariants are shown in Figure \ref{fig:LQasymp_during}.
These plots demonstrate that in the large width limit, as $\alpha \longmapsto \infty$, training causes the deviation from the simple Gaussian model to vary increasingly.
Aligned with the expected behaviour for pre-training as dictated by the traditional large-width limit, the deviations are very small before training as the weight distributions are well modelled by the simple Gaussian, however over training the weights change greatly from this form.
Whilst the lower $\alpha$ values appear to qualitatively show convergence in their deviations, the larger values do not, indicating more training is needed for larger models for the weight distributions to stabilise.

The most deviant invariant is not consistent between $\alpha$ values, being $I_3 = \sum_{i,j}W_{ij}^2$ for $\alpha \in \{10, 40\}$ (matching the results seen in Figure \ref{fig:LQ_during}), for $\alpha \in \{160, 640\}$ the most deviant invariant is $I_{10}= \sum_{i,j,k,l}W_{ij}W_{kl}$.
These invariants are both quadratic, and have a similar form, in fact $I_{10} = I_3 + \sum_{i, j, k \neq i , l \neq j}W_{ij}W_{kl}$, and this indicates that in the limit of increasing NN layer width this latter term becomes more significant in the trained weight distribution.

\begin{figure}[!t]
    \centering
    \begin{subfigure}{0.24\textwidth}
        \centering
        \includegraphics[width=0.98\textwidth]{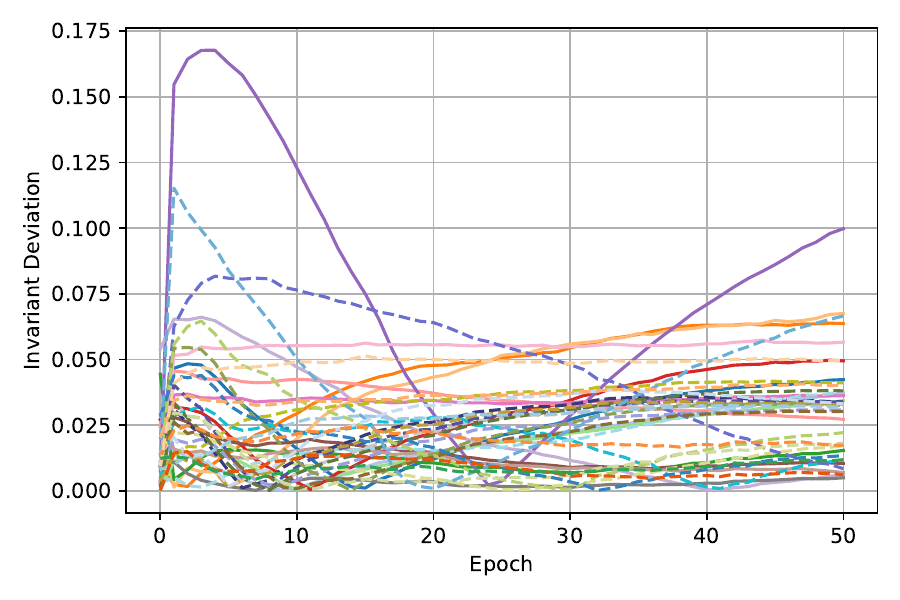}
        \caption{$\alpha=10$}
    \end{subfigure} 
    \begin{subfigure}{0.24\textwidth}
        \centering
        \includegraphics[width=0.98\textwidth]{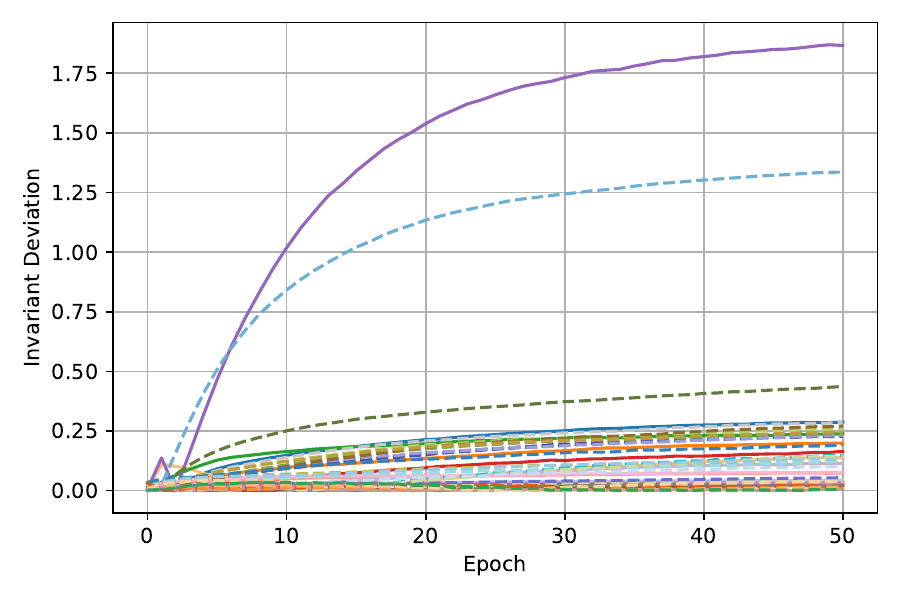}
        \caption{$\alpha=40$}
    \end{subfigure} 
    \begin{subfigure}{0.24\textwidth}
        \centering
        \includegraphics[width=0.98\textwidth]{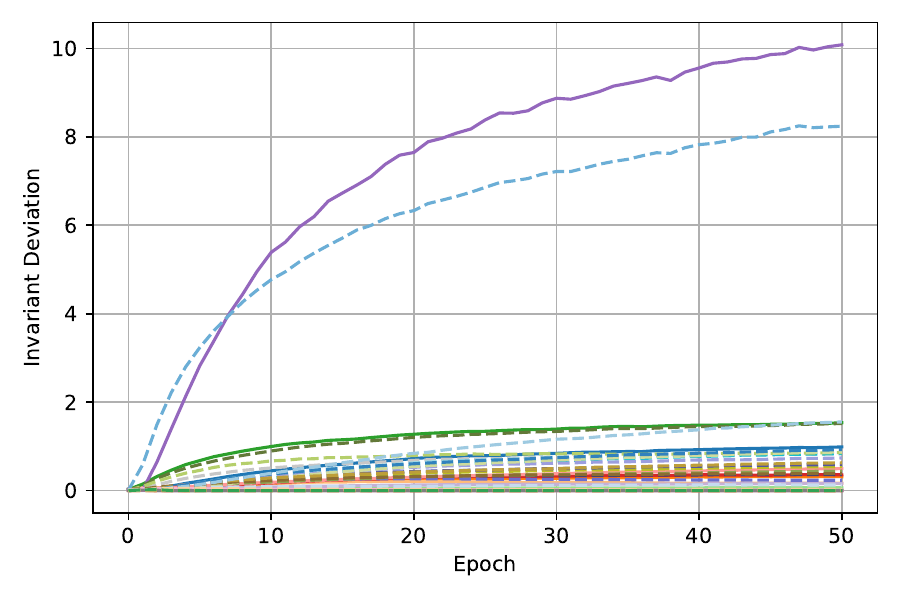}
        \caption{$\alpha=160$}
    \end{subfigure} 
    \begin{subfigure}{0.24\textwidth}
        \centering
        \includegraphics[width=0.98\textwidth]{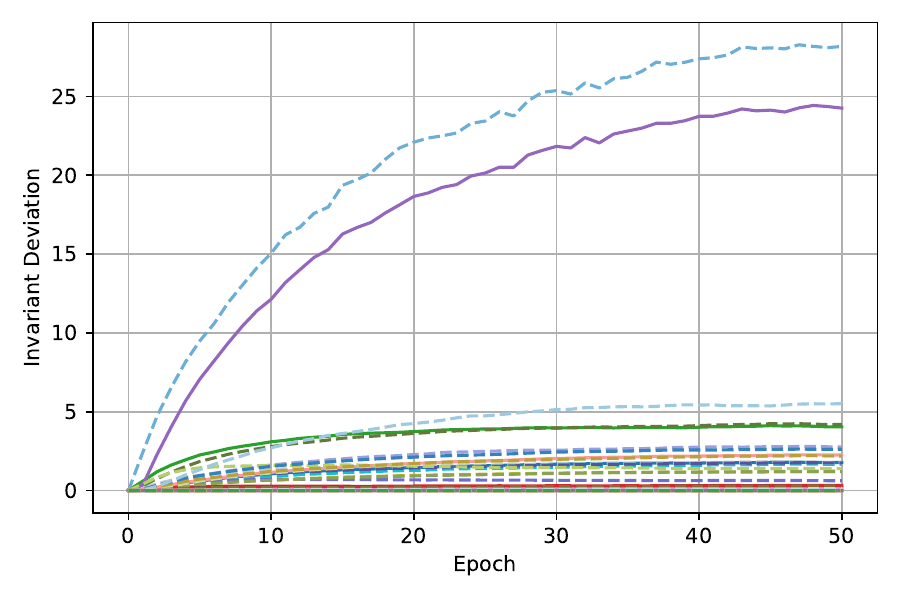}
        \caption{$\alpha=640$}
    \end{subfigure} \\
    \begin{subfigure}{0.98\textwidth}
        \centering
        \includegraphics[width=0.98\textwidth]{Figures/LHDeviations/CQ_Legend.pdf}
        \caption{Legend}
    \end{subfigure}
    \caption{Variation of the cubic deviations (solid lines) and quartic deviations (dashed lines), labelled respectively by their invariants $I_{14}-I_{52}$, across the 50 epochs of training. Displayed for NNs of increasing width with $\alpha$ neurons per central layer. The legend is the same throughout and collectively shown at the bottom (e) for readability. Note the y-axes scales are fixed within initialisations, but differ between them.}
    \label{fig:CQasymp_during}
\end{figure}

Equivalent deviation plots for the cubic and quartic invariants are shown in Figure \ref{fig:CQasymp_during}.
These plots assess the suitability of the more general fitted PIGMMs, and show that as layer width increases this form of PIGMM becomes less suitable for the final trained weight distributions, as the y-axes scales of the deviations increase with $\alpha$.

Some invariants do still have consistently low deviations across the $\alpha$ values, but most notably the most deviant invariants become much more significantly deviant.
The behaviour for $\alpha=10$ is similar to Figure \ref{fig:CQ_during}, however for larger $\alpha$ values the $I_9 = \sum_{i,j}W_{ij}^3$ and $I_{32} = \sum_{i,j} W_{ij}^4$ invariants deviate significantly from the fitted PIGMM predicted values.
These also deviate the most during training for $\alpha = 10$ in both architectures, but show substantially more deviation than the other invariants as $\alpha$ increases.
Where the cubic and quartic invariants are not well predicted by the PIGMM, this directly shows how the Gaussianity assumption (only up to quadratic order is needed) in the fitted models is broken, such that to improve the model beyond Gaussianity these exact invariant terms should be added into the model action and fitted to the data.

\begin{figure}[!t]
    \centering
    \begin{subfigure}{0.48\textwidth}
        \centering
        \includegraphics[width=0.98\textwidth]{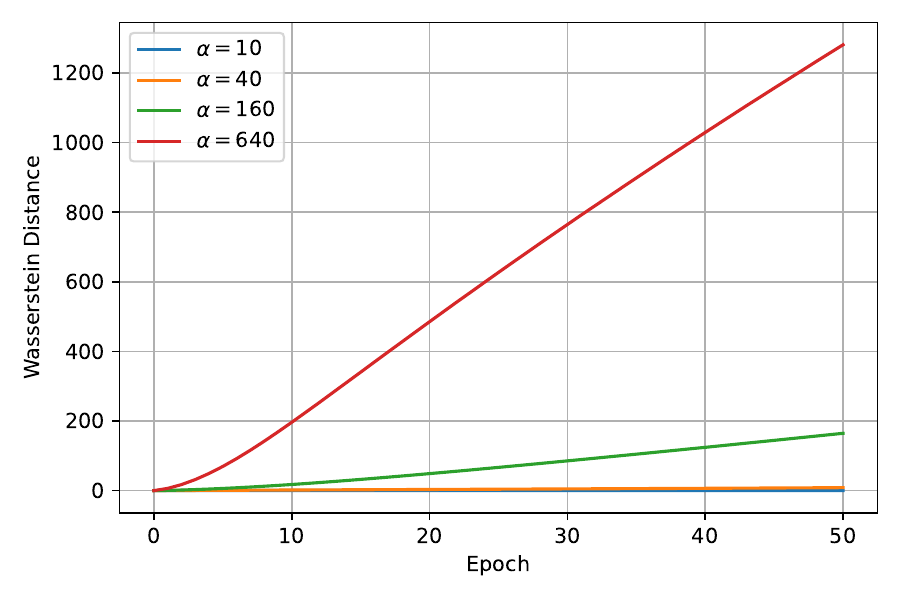}
        \caption{Linear scale}
    \end{subfigure} 
    \begin{subfigure}{0.48\textwidth}
        \centering
        \includegraphics[width=0.98\textwidth]{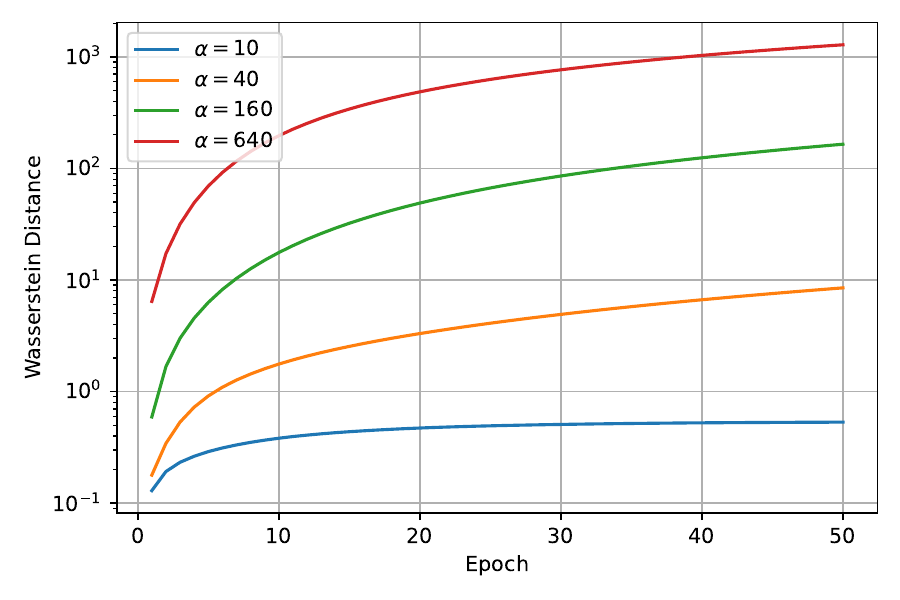}
        \caption{Log scale}
    \end{subfigure}
    \caption{Computations of the Wasserstein distance between the simple Gaussian model for the Gaussian initialisation and the permutation-invariant Gaussian matrix models of increasing layer sizes ($\alpha \in [10,40,160,640]$) at each epoch (based on the observed model parameters). Shown with the distances on a (a) linear scale and then (b) log scale (note the distances of 0 cannot be shown on a log-scale).}
    \label{fig:wasserstein_asymptotic}
\end{figure}

The final plots are for the Wasserstein distances of these asymptotic architectures from their simple Gaussian models in Figure \ref{fig:wasserstein_asymptotic}, with the same data on both a linear and log scale.
These distances are much more significant for larger layer sizes and on the linear scale plot show no sign of convergence through the training as the weight distributions move further away from their simple Gaussian models. %should have done more dimensional-normalisation here?
On the log-scale, the separation of final Wasserstein distances appears approximately constant with the increasing $\alpha$ values, which motivates a power law behaviour for the post-training Wasserstein distance from the initialisation simple Gaussian model with the NN layer size.

Overall, this section has shown that larger architectures with more weights to model have more complicated weight distributions.
As there're more weights the fixed 13 degrees of freedom in the PIGMM formalism become less representative, however the higher-order invariant deviations indicate exactly where the Gaussianity assumption is most violated and how more general matrix models should be built as better, yet still efficient, models for modelling weight matrix distributions.

%%%%%%%%%%%%%%%%%%%%%%%%%%%%%%%%%%%%%%%%%%%%%%%%%
\section{Summary \& Outlook}\label{sec:conc}

This work studied distributions of neural network weight matrices using matrix models. Motivated by permutation symmetries of dense feed forward neural networks, the framework of permutation invariant matrix models was used as an efficient and scalable model to describe the statistics of weight matrices. 

This was done by fitting general 13-parameter permutation-invariant Gaussian matrix models (PIGMM) \cite{Ramgoolam:2018xty} to weight-matrix data, and comparing theoretical expectation values of a selection of 39 cubic and quartic invariants to the experimental averages computed from the weight-matrix data.  
The quantitative measure of non-Gaussianity for each invariant $I_i$ used was the deviation  \eqref{eq:CQinv_deviation}, defined as the 
difference between the theoretical and experimental expectation values divided by the standard deviation obtained from the experimental data. 
The special point in the 13-parameter space is the 2-parameter simple Gaussian, where all the matrix entries are drawn from a one-variable Gaussian distribution, so that they are independent and identically distributed. 
The 13-parameter model generalises the simple Gaussian to include non-trivial dependencies between the variables encoded in linear and quadratic permutation invariants associated to directed graphs with up to two edges (Figure \ref{fig:LQ_invariant_graphs}). 

Using the standard Gaussian initialisation of the weight matrices, the non-Gaussianity measures start as expected at zero. 
They initially increase through the epochs of  training but for all the invariants, they decrease in the latter stages of training to finish at values less than $1$, showing that non-trivial Gaussians in the 13-parameter space provide a good fit to the distribution of weight matrices in the later stages of training (Figure \ref{fig:CQ_during}). 
The Wasserstein distance on the space of 13-parameter Gaussians, as outlined in §\ref{sec:wasserstein} and derived in §\ref{app:wasserstein_derivation} was calculated to confirm that the distance between the fitted Gaussians and the simple Gaussian increase monotonically through training (Figure \ref{fig:wasserstein}).  
The departure away from the simple Gaussian after initialisation is also evidenced in §\ref{sec:lq} where the deviations of the  expectation values of the linear and quadratic permutation invariants away from predictions of the simple Gaussian are quantified using the measures defined in \eqref{eq:LQinv_deviation} and  \eqref{eq:param_deviation}. 
As the Figures \ref{fig:CQ_during} and \ref{fig:wasserstein} show, similar results are obtained when the weight matrices are initialised using the standard one-variable uniform distribution. 
The simple Gaussian provides generally a good fit for the expectation values of the permutation invariants $I_i$ of the initial weight matrices because these invariants are sums over large number of matrix entries and the central limit theorem ensures that these can be obtained from a Gaussian approximation. 

Introducing regularisation into the training regime made the deviation from the simple Gaussian model of the initialisation distributions consistent between the layers, but changed which invariants deviated the most and hence where the Gaussianity assumption of the PIGMM class breaks down.
The monotonic smooth changes of the weight distributions no longer appear under regularisation, as the PIGMMs become worse (yet still sufficiently good) fits, yet have clearer non-Gaussianities emerging over training as some invariants deviate more.

Under a different architecture, first tests for the suitability of PIGMMs under the large-width limit were also performed.
Using the same techniques, as an architecture's number of parameters increases both the simple Gaussian model and PIGMMs become less suitable.
This effect is especially prominent for the simple Gaussian models, yet is more marginal for the PIGMMs as few higher-order invariants express the largest deviations indicating limited extension would be required to accommodate higher-order correlation effects in the weight distributions.

These asymptotic results  motivate the development of these PIGMMs to  more general classes of models, for example adding as perturbations to the 13-parameter Gaussians the specific higher degree invariants identified as having higher deviations.  This work has many other natural directions for generalisation, the first the authors hope to perform would include extending the formalism to accommodate non-square matrices, as is more common in traditional ML practical uses. 
A good starting point for the non-square matrices of size $ d_1 \times d_2 $ would be to consider the product symmetric group $ S_{ d_1} \times S_{d_2} $ to organise the invariants, which should be related to bipartite graphs. 
Developing the graph theoretic  enumeration formulae and associated Gaussian models is a concrete future direction. Another is extending the theoretical and computational framework of the PIGMM from one-matrix to one matrix coupled to a vector, and to multi-matrix systems with permutation actions motivated by neural networks.
The application of these extensions to weight-matrix data along the lines of the present paper, provide many fascinating future research directions. 

Whilst the code functionality in this work's  \href{https://github.com/edhirst/PIGWMM}{\texttt{GitHub}} repository accommodates application of this formalism to the CIFAR classification problem \cite{cifar}, compute restrictions encouraged this initial work to focus on the MNIST problem.
Broadening the study to more varied ML problems would be another interesting direction for development, as well as how this formalism applies to other architecture classes, including weight matrices in convolutional NNs and transformers. %vary loss, vary optimiser?

Overall, this work demonstrates the need for general matrix model classes that can be used for representing weight matrix distributions. 
With a better model, initialisation schemes can be optimised for a problem, but more importantly the class of high performing and optimally trained models can be manageably studied, using the reduced degrees of freedom defined in these models (especially as the architecture scales up). 
This work shows that PIGMMs are strong candidate models for general trained weight matrix distributions.

%%%%%%%%%%%%%%%%%%%%%%%%%%%%%%%%%%%%%%%%%%%%%%%%%
\section*{Acknowledgements}
The authors wish to thank Yang-Hui He,  Jurgis Pasukonis, Michael Stephanou and Michael Toomey for useful conversations related to this research.  
EH acknowledges support from Pierre Andurand over the course of this research.
SR is supported by the STFC consolidated grant ST/P000754/1 “String Theory, Gauge Theory and Duality”. SR is grateful for a Visiting Professorship at Dublin Institute for Advanced Studies,  held during 2024, while this project was in progress.  SR also gratefully acknowledges a visit to the Perimeter Institute in November 2024: this research was supported in part by Perimeter Institute for Theoretical Physics. Research at Perimeter Institute is supported by the Government of Canada through the Department of Innovation, Science, and Economic Development, and by the Province of Ontario through the Ministry of Colleges and Universities. 
This research utilised Queen Mary's Apocrita HPC facility \cite{apocrita}, supported by QMUL Research-IT.

%%%%%%%%%%%%%%%%%%%%%%%%%%%%%%%%%%%%%%%%%%%%%%%%%
%\section*{Data Accessibility}
\paragraph{\textbf{Data Accessibility}}\mbox{}\\
Code scripts for data generation and analysis are made publicly available at this work's respective GitHub repository: \href{https://github.com/edhirst/PIGWMM}{\texttt{https://github.com/edhirst/PIGWMM}}.
The weight matrix ensembles studied throughout this work amounted to $\sim$1TB of data, exceeding GitHub's cloud storage limits.
These are hence not remotely available but can be provided upon request, and the GitHub repository above contains all the code used in generating them.

%%%%%%%%%%%%%%%%%%%%%%%%%%%%%%%%%%%%%%%%%%%%%%%%%
\appendix
%%%%%%%%%%%%%%%%%%%%%%%%%%%%%%%%%%%%%%%%%%%%%%%%%
\section{Definition of Permutation Invariants via Graphs}\label{app:inv_graphs}
The invariants, $I_i$, of the diagonal permutation symmetry considered in this work, and introduced in §\ref{sec:gmms}, are defined below in terms of the square weight matrices $W_{ij}$. 
The order of the invariant refers to the number of occurrences of $W_{ij}$ in the definition, and the number of nodes refers to the number of free indices summed over in the definition.

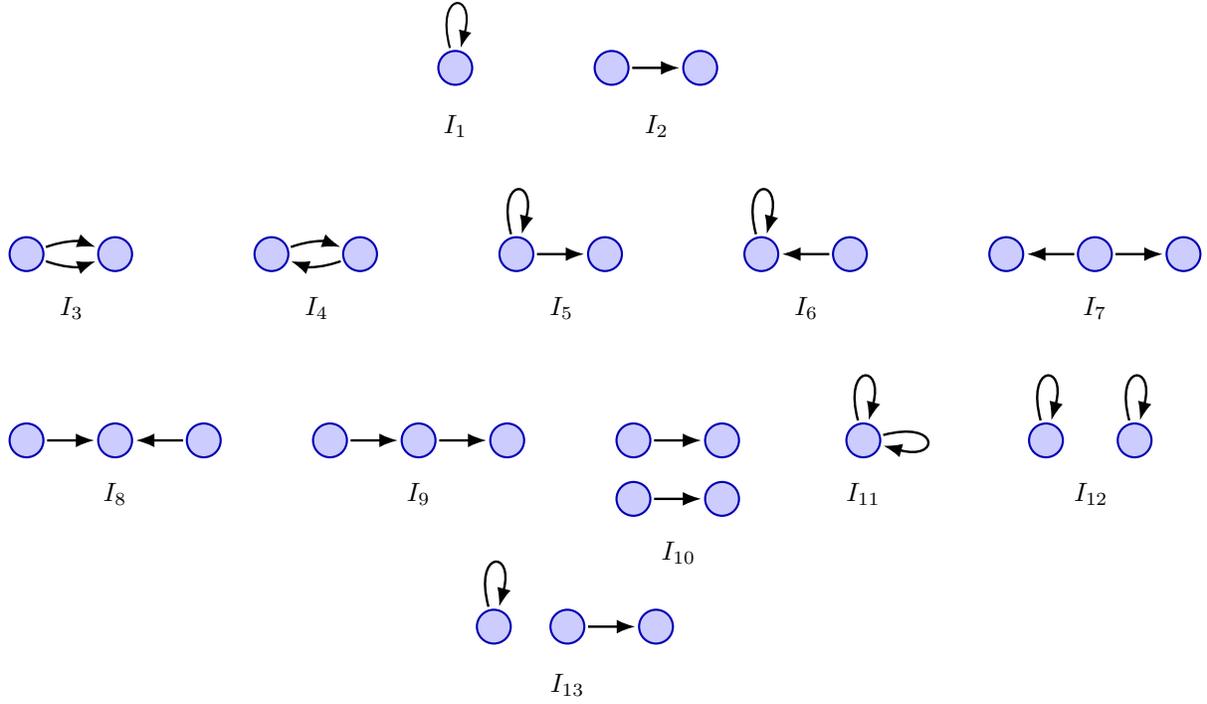
\begin{figure}[H]
\centering
\begin{tikzpicture}[edge, node distance=0.8cm and 0.8cm]
% I1 = sum_i W_{ii}
\node[node, xshift=5.7cm] (a1) {};
\path (a1) edge[loop above, thick, looseness=17] (a1);
\node[below=0.25cm of a1] {$I_1$};

% I2 = sum_{i,j} W_{ij}
\node[node,right=1.6cm of a1] (a2) {};
\node[node,right=0.7cm of a2] (a3) {};
\path (a2) edge (a3);
\node[below=0.25cm of a2, xshift=0.6cm] {$I_2$};

% I3 = W_{ij}^2
\node[node,below=2cm of a1, xshift=-5.7cm] (b1) {};
\node[node,right=0.7cm of b1] (b2) {};
\path (b1) edge[bend left=20] (b2)
      (b1) edge[bend right=20] (b2);
\node[below=0.2cm of b1, xshift=0.6cm] {$I_3$};

% I4 = W_{ij}W_{ji}
\node[node,right=1.6cm of b2] (b3) {};
\node[node,right=0.7cm of b3] (b4) {};
\path (b3) edge[bend left=20] (b4)
      (b4) edge[bend left=20] (b3);
\node[below=0.2cm of b3, xshift=0.6cm] {$I_4$};

% I5 = W_{ii}W_{ij}
\node[node,right=1.6cm of b4] (b5) {};
\node[node,right=0.7cm of b5] (b6) {};
\path (b5) edge[loop above, looseness=17] (b5)
      (b5) edge (b6);
\node[below=0.2cm of b5, xshift=0.6cm] {$I_5$};

% I6 = W_{ii}W_{ji}
\node[node,right=1.6cm of b6] (b7) {};
\node[node,right=0.7cm of b7] (b8) {};
\path (b7) edge[loop above, looseness=17] (b7)
      (b8) edge (b7);
\node[below=0.2cm of b7, xshift=0.6cm] {$I_6$};

% I7 = W_{ij}W_{ik}
\node[node,right=1.6cm of b8] (b9) {};
\node[node,right=0.7cm of b9] (b10) {};
\node[node,right=0.7cm of b10] (b11) {};
\path (b10) edge (b9)
      (b10) edge (b11);
\node[below=0.2cm of b10, xshift=0cm] {$I_7$};

% I8 = sum_{i,j,k} W_{ij}W_{jk}
\node[node,below=2cm of b1] (c1) {};
\node[node,right=0.7cm of c1] (c2) {};
\node[node,right=0.7cm of c2] (c3) {};
\path (c1) edge (c2)
      (c3) edge (c2);
\node[below=0.2cm of c2] {$I_8$};

% I9 = sum_{i,j,k,l} W_{ij}W_{kl}
\node[node,right=1.2cm of c3] (c4) {};
\node[node,right=0.7cm of c4] (c5) {};
\node[node,right=0.7cm of c5] (c6) {};
\path (c4) edge (c5)
      (c5) edge (c6);
\node[below=0.2cm of c5] {$I_9$};

% I10 = sum_i W_{ii}^2
\node[node,right=1.2cm of c6] (c7) {};
\node[node,right=0.7cm of c7] (c8) {};
\node[node,below=0.3cm of c7] (c9) {};
\node[node,right=0.7cm of c9] (c10) {};
\path (c7) edge (c8)
      (c9) edge (c10);
\node[below=0.2cm of c9, xshift=0.6cm] {$I_{10}$};

% I11 = sum_{i,j} W_{ii}W_{jj}
\node[node,right=1.4cm of c8] (c11) {};
\path (c11) edge[loop above, looseness=17] (c11)
      (c11) edge[loop right, looseness=17] (c11);
\node[below=0.2cm of c11] {$I_{11}$};

% I12 = sum_{i,j,k} W_{ii}W_{jk}
\node[node,right=1.95cm of c11] (c12) {};
\node[node,right=0.7cm of c12] (c13) {};
\path (c12) edge[loop above, looseness=17] (c12)
      (c13) edge[loop above, looseness=17] (c13);
\node[below=0.2cm of c12, xshift=0.6cm] {$I_{12}$};

% I13 = W_{ii}W_{jk}
\node[node,below=2cm of c5, xshift=1cm] (d1) {};
\node[node,right=0.5cm of d1] (d2) {};
\node[node,right=0.7cm of d2] (d3) {};
\path (d1) edge[loop above, looseness=17] (d1)
      (d2) edge (d3);
\node[below=0.25cm of d2] {$I_{13}$};
\end{tikzpicture}
\caption{Graphical representation of the linear ($I_1, I_2$) and quadratic ($I_3,\dots,I_{13}$) invariants. Each index corresponds to a node, and each $W_{ij}$ to a directed edge from $i$ to $j$.}
\label{fig:LQ_invariant_graphs}
\end{figure}

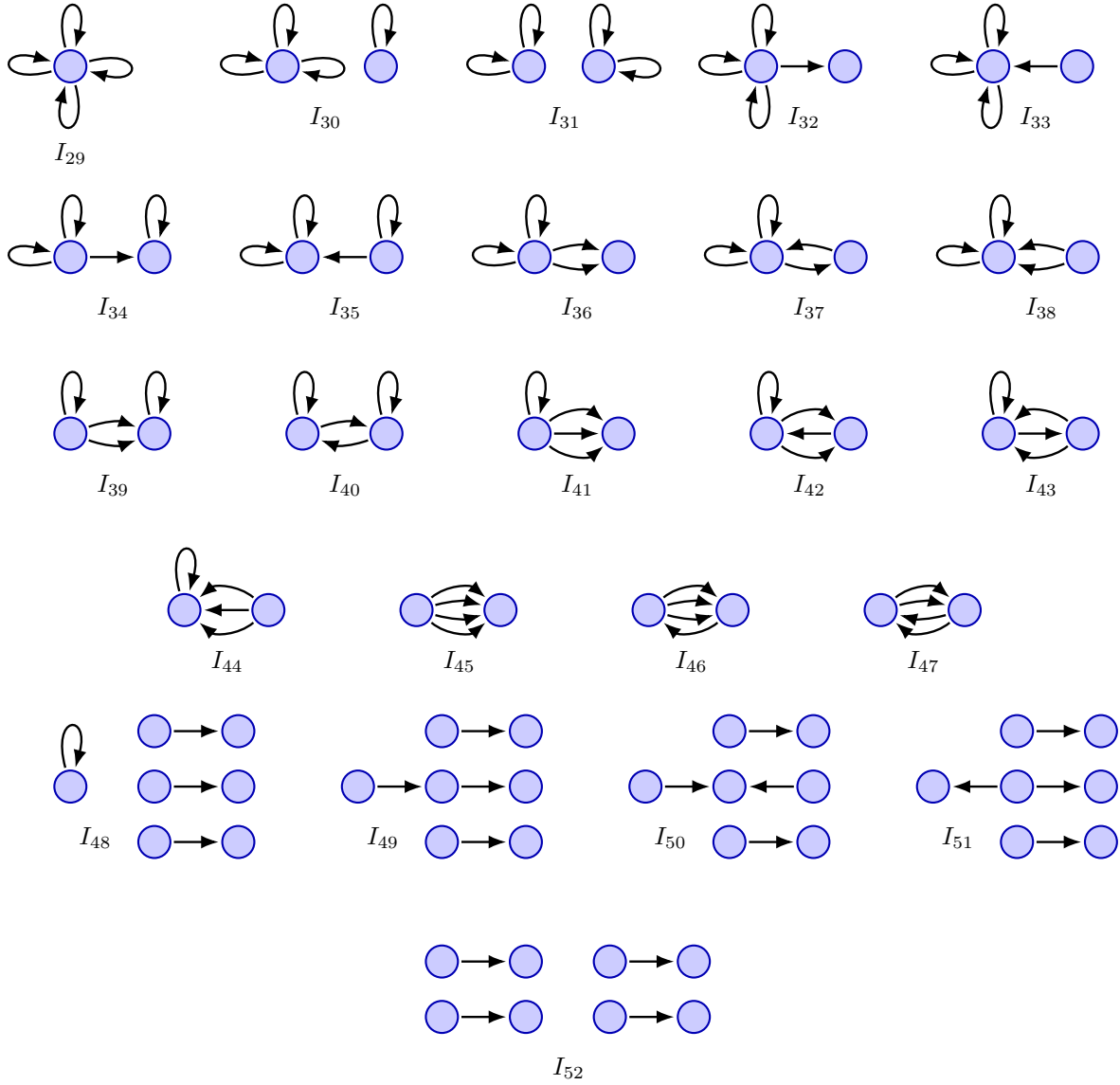
\begin{figure}[H]
\centering
\begin{tikzpicture}[edge, node distance=0.8cm and 0.8cm]
%%% Row 1: I29 -- I33 %%%
% I29 = W_{ii}^4
\node[node] (q1) {};
\path (q1) edge[loop above, looseness=17] (q1)
      (q1) edge[loop right, looseness=17] (q1)
      (q1) edge[loop left, looseness=17] (q1)
      (q1) edge[loop below, looseness=17] (q1);
\node[below=0.7cm of q1] {$I_{29}$};

% I30 = W_{ii}^3 W_{jj}
\node[node,right=2.5cm of q1] (q2) {};
\node[node,right=0.9cm of q2] (q3) {};
\path (q2) edge[loop above, looseness=17] (q2)
      (q2) edge[loop right, looseness=17] (q2)
      (q2) edge[loop left, looseness=17] (q2)
      (q3) edge[loop above, looseness=17] (q3);
\node[below=0.2cm of q2, xshift=0.6cm] {$I_{30}$};

% I31 = W_{ii}^2 W_{jj}^2
\node[node,right=1.6cm of q3] (q4) {};
\node[node,right=0.5cm of q4] (q5) {};
\path (q4) edge[loop above, looseness=17] (q4)
      (q4) edge[loop left, looseness=17] (q4)
      (q5) edge[loop above, looseness=17] (q5)
      (q5) edge[loop right, looseness=17] (q5);
\node[below=0.2cm of q4, xshift=0.5cm] {$I_{31}$};

% I32 = W_{ii}^3 W_{ij}
\node[node,right=1.8cm of q5] (q6) {};
\node[node,right=0.7cm of q6] (q7) {};
\path (q6) edge[loop above, looseness=17] (q6)
      (q6) edge[loop below, looseness=17] (q6)
      (q6) edge[loop left, looseness=17] (q6)
      (q6) edge (q7);
\node[below=0.2cm of q6, xshift=0.6cm] {$I_{32}$};

% I33 = W_{ji} W_{ii}^3
\node[node,right=1.6cm of q7] (q8) {};
\node[node,right=0.7cm of q8] (q9) {};
\path (q8) edge[loop above, looseness=17] (q8)
      (q8) edge[loop below, looseness=17] (q8)
      (q8) edge[loop left, looseness=17] (q8)
      (q9) edge (q8);
\node[below=0.2cm of q8, xshift=0.6cm] {$I_{33}$};

%%% Row 2: I34 -- I38 %%%
% I34 = W_{ii}^2 W_{ij} W_{jj}
\node[node,below=2.2cm of q1] (q10) {};
\node[node,right=0.7cm of q10] (q11) {};
\path (q10) edge[loop above, looseness=17] (q10)
      (q10) edge[loop left, looseness=17] (q10)
      (q10) edge (q11)
      (q11) edge[loop above, looseness=17] (q11);
\node[below=0.2cm of q10, xshift=0.6cm] {$I_{34}$};

% I35 = W_{jj} W_{ji} W_{ii}^2
\node[node,right=1.6cm of q11] (q12) {};
\node[node,right=0.7cm of q12] (q13) {};
\path (q12) edge[loop above, looseness=17] (q12)
      (q12) edge[loop left, looseness=17] (q12)
      (q13) edge[loop above, looseness=17] (q13)
      (q13) edge (q12);
\node[below=0.2cm of q12, xshift=0.6cm] {$I_{35}$};

% I36 = W_{ii}^2 W_{ij}^2
\node[node,right=1.6cm of q13] (q14) {};
\node[node,right=0.7cm of q14] (q15) {};
\path (q14) edge[loop above, looseness=17] (q14)
      (q14) edge[loop left, looseness=17] (q14)
      (q14) edge[bend right=20] (q15)
      (q14) edge[bend left=20] (q15);
\node[below=0.2cm of q14, xshift=0.6cm] {$I_{36}$};

% I37 = W_{ji} W_{ii}^2 W_{ij}
\node[node,right=1.6cm of q15] (q16) {};
\node[node,right=0.7cm of q16] (q17) {};
\path (q16) edge[loop above, looseness=17] (q16)
      (q16) edge[loop left, looseness=17] (q16)
      (q16) edge[bend right=20] (q17)
      (q17) edge[bend right=20] (q16);
\node[below=0.2cm of q16, xshift=0.6cm] {$I_{37}$};

% I38 = W_{ji}^2 W_{ii}^2
\node[node,right=1.6cm of q17] (q18) {};
\node[node,right=0.7cm of q18] (q19) {};
\path (q18) edge[loop above, looseness=17] (q18)
      (q18) edge[loop left, looseness=17] (q18)
      (q19) edge[bend left=20] (q18)
      (q19) edge[bend right=20] (q18);
\node[below=0.2cm of q18, xshift=0.6cm] {$I_{38}$};

%%% Row 3: I39 -- I43 %%%
% I39 = W_{ii} W_{ij}^2 W_{jj}
\node[node,below=2cm of q10] (q20) {};
\node[node,right=0.7cm of q20] (q21) {};
\path (q20) edge[loop above, looseness=17] (q20)
      (q20) edge[bend left=20] (q21)
      (q20) edge[bend right=20] (q21)
      (q21) edge[loop above, looseness=17] (q21);
\node[below=0.2cm of q20, xshift=0.6cm] {$I_{39}$};

% I40 = W_{ji} W_{ii} W_{ij} W_{jj}
\node[node,right=1.6cm of q21] (q22) {};
\node[node,right=0.7cm of q22] (q23) {};
\path (q22) edge[loop above, looseness=17] (q22)
      (q22) edge[bend left=20] (q23)
      (q23) edge[loop above, looseness=17] (q23)
      (q23) edge[bend left=20] (q22);
\node[below=0.2cm of q22, xshift=0.6cm] {$I_{40}$};

% I41 = W_{ii} W_{ij}^3
\node[node,right=1.6cm of q23] (q24) {};
\node[node,right=0.7cm of q24] (q25) {};
\path (q24) edge[loop above, looseness=17] (q24)
      (q24) edge[bend left=40] (q25)
      (q24) edge (q25)
      (q24) edge[bend right=40] (q25);
\node[below=0.2cm of q24, xshift=0.6cm] {$I_{41}$};

% I42 = W_{ji} W_{ii} W_{ij}^2
\node[node,right=1.6cm of q25] (q26) {};
\node[node,right=0.7cm of q26] (q27) {};
\path (q26) edge[loop above, looseness=17] (q26)
      (q27) edge (q26)
      (q26) edge[bend left=40] (q27)
      (q26) edge[bend right=40] (q27);
\node[below=0.2cm of q26, xshift=0.6cm] {$I_{42}$};

% I43 = W_{ji}^2 W_{ii} W_{ij}
\node[node,right=1.6cm of q27] (q28) {};
\node[node,right=0.7cm of q28] (q29) {};
\path (q28) edge[loop above, looseness=17] (q28)
      (q29) edge[bend left=40] (q28)
      (q29) edge[bend right=40] (q28)
      (q28) edge (q29);
\node[below=0.2cm of q28, xshift=0.6cm] {$I_{43}$};

%%% Row 4: I44 -- I47 %%%
% I44 = W_{ji}^3 W_{ii}
\node[node,below=2cm of q20, xshift=1.6cm] (q30) {};
\node[node,right=0.7cm of q30] (q31) {};
\path (q30) edge[loop above, looseness=17] (q30)
      (q31) edge[bend left=40] (q30)
      (q31) edge (q30)
      (q31) edge[bend right=40] (q30);
\node[below=0.2cm of q30, xshift=0.6cm] {$I_{44}$};

% I45 = W_{ij}^4
\node[node,right=1.6cm of q31] (q32) {};
\node[node,right=0.7cm of q32] (q33) {};
\path (q32) edge[bend left=40] (q33)
      (q32) edge[bend left=15] (q33)
      (q32) edge[bend right=15] (q33)
      (q32) edge[bend right=40] (q33);
\node[below=0.2cm of q32, xshift=0.6cm] {$I_{45}$};

% I46 = W_{ji} W_{ij}^3
\node[node,right=1.6cm of q33] (q34) {};
\node[node,right=0.7cm of q34] (q35) {};
\path (q34) edge[bend left=40] (q35)
      (q34) edge[bend left=15] (q35)
      (q34) edge[bend right=15] (q35)
      (q35) edge[bend left=40] (q34);
\node[below=0.2cm of q34, xshift=0.6cm] {$I_{46}$};

% I47 = W_{ji}^2 W_{ij}^2
\node[node,right=1.6cm of q35] (q36) {};
\node[node,right=0.7cm of q36] (q37) {};
\path (q36) edge[bend left=40] (q37)
      (q36) edge[bend left=15] (q37)
      (q37) edge[bend left=15] (q36)
      (q37) edge[bend left=40] (q36);
\node[below=0.2cm of q36, xshift=0.6cm] {$I_{47}$};

%%% Row 5: I48 -- I51 %%%
% I48 = W_{ii} W_{jk} W_{lm} W_{no}
\node[node,below=2cm of q30, xshift=-1.6cm] (q38) {};
\node[node,right=0.7cm of q38] (q39) {};
\node[node,right=0.7cm of q39] (q40) {};
\node[node,above=0.3cm of q39] (q41) {};
\node[node,right=0.7cm of q41] (q42) {};
\node[node,below=0.3cm of q39] (q43) {};
\node[node,right=0.7cm of q43] (q44) {};
\path (q38) edge[loop above, looseness=17] (q38)
      (q39) edge (q40)
      (q41) edge (q42)
      (q43) edge (q44);
\node[below=0.2cm of q38, xshift=0.35cm] {$I_{48}$};

% I49 = W_{ij} W_{jk} W_{lm} W_{no}
\node[node,right=1.2cm of q40] (q45) {};
\node[node,right=0.7cm of q45] (q46) {};
\node[node,right=0.7cm of q46] (q47) {};
\node[node,above=0.3cm of q46] (q48) {};
\node[node,right=0.7cm of q48] (q49) {};
\node[node,below=0.3cm of q46] (q50) {};
\node[node,right=0.7cm of q50] (q51) {};
\path (q45) edge (q46)
      (q46) edge (q47)
      (q48) edge (q49)
      (q50) edge (q51);
\node[below=0.2cm of q45, xshift=0.35cm] {$I_{49}$};

% I50 = W_{ij} W_{kj} W_{lm} W_{no}
\node[node,right=1.2cm of q47] (q52) {};
\node[node,right=0.7cm of q52] (q53) {};
\node[node,right=0.7cm of q53] (q54) {};
\node[node,above=0.3cm of q53] (q55) {};
\node[node,right=0.7cm of q55] (q56) {};
\node[node,below=0.3cm of q53] (q57) {};
\node[node,right=0.7cm of q57] (q58) {};
\path (q52) edge (q53)
      (q54) edge (q53)
      (q55) edge (q56)
      (q57) edge (q58);
\node[below=0.2cm of q52, xshift=0.35cm] {$I_{50}$};

% I51 = W_{ji} W_{jk} W_{lm} W_{no}
\node[node,right=1.2cm of q54] (q59) {};
\node[node,right=0.7cm of q59] (q60) {};
\node[node,right=0.7cm of q60] (q61) {};
\node[node,above=0.3cm of q60] (q62) {};
\node[node,right=0.7cm of q62] (q63) {};
\node[node,below=0.3cm of q60] (q64) {};
\node[node,right=0.7cm of q64] (q65) {};
\path (q60) edge (q59)
      (q60) edge (q61)
      (q62) edge (q63)
      (q64) edge (q65);
\node[below=0.2cm of q59, xshift=0.35cm] {$I_{51}$};

%%% Row 6: I52 %%%
% I52 = W_{ij} W_{kl} W_{mn} W_{op}
\node[node,below=1.2cm of q50] (q66) {};
\node[node,right=0.7cm of q66] (q67) {};
\node[node,right=0.7cm of q67] (q68) {};
\node[node,right=0.7cm of q68] (q69) {};
\node[node,below=0.3cm of q66] (q70) {};
\node[node,right=0.7cm of q70] (q71) {};
\node[node,right=0.7cm of q71] (q72) {};
\node[node,right=0.7cm of q72] (q73) {};
\path (q66) edge (q67)
      (q68) edge (q69)
      (q70) edge (q71)
      (q72) edge (q73);
\node[below=0.2cm of q71, xshift=0.6cm] {$I_{52}$};
\end{tikzpicture}
\caption{Graphical representation of the quartic ($I_{29}$–$I_{52}$) invariants. Each index corresponds to a node, and each $W_{ij}$ to a directed edge from $i$ to $j$.}
\label{fig:Q_invariant_graphs}
\end{figure}

The formulas for the studied invariants are given below, where focus was on all linear and quadratic order invariants, then all low (1 \& 2) and high ($2n$ \& $2n-1$) node invariants for order $n=3,4$ at cubic and quartic orders.
These invariants can be faithfully represented by directed graphs, such that each invariant may be represented as a graph.
The considered invariants are displayed for linear and quadratic orders in Figure \ref{fig:LQ_invariant_graphs}, then for cubic order in Figure \ref{fig:C_invariant_graphs}, and quartic order in Figure \ref{fig:Q_invariant_graphs}.
In these graphs, each summed index in the invariant defining formula is represented by a node and each directed arrow is an occurrence of the weight matrix. 

\begin{equation}
\begin{split}
    \text{Linear:}& \\
    \{I_1, I_2\} & = \biggl\{ \sum_i W_{ii}, \  \sum_{i,j} W_{ij} \biggr\}\;,\\
\end{split}
\end{equation}
\begin{equation}
\begin{split}
    \text{Quadratic:}& \\
    \{I_3, ..., I_{13}\} & = \biggl\{ \sum_{i,j} W_{ij}^2, \ \sum_{i,j} W_{ij}W_{ji}, \ \sum_{i,j} W_{ii}W_{ij}, \ \sum_{i,j} W_{ii}W_{ji},\\
    & \qquad \sum_{i,j,k} W_{ij}W_{ik}, \ \sum_{i,j,k} W_{ij}W_{kj}, \ \sum_{i,j,k} W_{ij}W_{jk}, \ \sum_{i,j,k,l} W_{ij}W_{kl}, \\
    & \qquad \sum_{i} W_{ii}^2, \ \sum_{i,j} W_{ii}W_{jj}, \ \sum_{i,j,k} W_{ii}W_{jk} \biggr\}\;.
\end{split}
\end{equation}

Chosen low and high node number invariants $\{1,2,2n-1,2n\}$ for the higher-orders cubic ($n=3$) and quartic ($n=4$).
\begin{equation}
\begin{split}
    \text{Cubic:}& \\
    \text{1-Node:}& \\
    \{I_{14}\} & = \biggl\{ \sum_i W_{ii}^3 \biggr\}\;,\\
    \text{2-Node:}& \\
    \{I_{15}, ..., I_{23}\} & = \biggl\{ \sum_{i,j} W_{ii}^2W_{jj}, \ \sum_{i,j} W_{ii}W_{ij}W_{jj}, \ \sum_{i,j} W_{ii}^2W_{ij}, \ \sum_{i,j} W_{ji}W_{ii}^2, \ \sum_{i,j} W_{ii}W_{ij}^2, \\
    & \qquad \sum_{i,j} W_{ji}W_{ii}W_{ij}, \ \sum_{i,j} W_{ji}^2W_{ii}, \ \sum_{i,j} W_{ij}^3, \ \sum_{i,j} W_{ij}^2W_{ji} \biggr\}\;,\\
    \text{5-Node:}& \\
    \{I_{24}, ..., I_{27}\} & = \biggl\{ \sum_{i,j,k,l,m} W_{ii}W_{jk}W_{lm}, \ \sum_{i,j,k,l,m} W_{ij}W_{jk}W_{lm}, \\
    & \qquad \sum_{i,j,k,l,m} W_{ij}W_{kj}W_{lm}, \ \sum_{i,j,k,l,m} W_{ji}W_{jk}W_{lm}, \  \biggr\}\;,\\
    \text{6-Node:}& \\
    \{I_{28}\} & = \biggl\{ \sum_{i,j,k,l,m,n} W_{ij}W_{kl}W_{mn} \biggr\}\;,
\end{split}
\end{equation}
\begin{equation}
\begin{split}
    \text{Quartic:}& \\
    \text{1-Node:}& \\
    \{I_{29}\} & = \biggl\{ \sum_i W_{ii}^4 \biggr\}\;,\\
    \text{2-Node:}& \\
    \{I_{30}, ..., I_{47}\} & = \biggl\{ \sum_{i,j} W_{ii}^3W_{jj}, \ \sum_{i,j} W_{ii}^2W_{jj}^2, \ \sum_{i,j} W_{ii}^3W_{ij}, \ \sum_{i,j} W_{ji}W_{ii}^3, \\
    & \qquad \sum_{i,j} W_{ii}^2W_{ij}W_{jj}, \ \sum_{i,j} W_{jj}W_{ji}W_{ii}^2, \ \sum_{i,j} W_{ii}^2W_{ij}^2, \ \sum_{i,j} W_{ji}W_{ii}^2W_{ij}, \\
    & \qquad \sum_{i,j} W_{ji}^2W_{ii}^2, \ \sum_{i,j} W_{ii}W_{ij}^2W_{jj}, \ \sum_{i,j} W_{ji}W_{ii}W_{ij}W_{jj}, \ \sum_{i,j} W_{ii}W_{ij}^3, \\
    & \qquad \sum_{i,j} W_{ji}W_{ii}W_{ij}^2, \ \sum_{i,j} W_{ji}^2W_{ii}W_{ij}, \ \sum_{i,j} W_{ji}^3W_{ii}, \ \sum_{i,j} W_{ij}^4, \\
    & \qquad \sum_{i,j} W_{ji}W_{ij}^3, \ \sum_{i,j} W_{ji}^2W_{ij}^2 \biggr\}\;,\\
    \text{7-Node:}& \\
    \{I_{48}, ..., I_{51}\} & = \biggl\{ \sum_{i,j,k,l,m,n,o} W_{ii}W_{jk}W_{lm}W_{no}, \ \sum_{i,j,k,l,m,n,o} W_{ij}W_{jk}W_{lm}W_{no}, \\
    & \qquad \sum_{i,j,k,l,m,n,o} W_{ij}W_{kj}W_{lm}W_{no}, \ \sum_{i,j,k,l,m,n,o} W_{ji}W_{jk}W_{lm}W_{no} \biggr\}\;,\\
    \text{8-Node:}& \\
    \{I_{52}\} & = \biggl\{ \sum_{i,j,k,l,m,n,o,p} W_{ij}W_{kl}W_{mn}W_{op} \biggr\}\;,
\end{split}
\end{equation}

\begin{figure}[H]
\centering
\begin{tikzpicture}[edge, node distance=0.8cm and 0.8cm]
%%% Row 1: I14 -- I18 %%%
% I14 = W_{ii}^3
\node[node] (d1) {};
\path (d1) edge[loop above, looseness=17] (d1)
      (d1) edge[loop right, looseness=17] (d1)
      (d1) edge[loop left, looseness=17] (d1);
\node[below=0.2cm of d1] {$I_{14}$};

% I15 = W_{ii}^2 W_{jj}
\node[node,right=2.75cm of d1] (d2) {};
\node[node,right=0.5cm of d2] (d3) {};
\path (d2) edge[loop above, looseness=17] (d2)
      (d3) edge[loop right, looseness=17] (d3)
      (d3) edge[loop above, looseness=17] (d3);
\node[below=0.2cm of d2, xshift=0.6cm] {$I_{15}$};

% I16 = W_{ii} W_{ij} W_{jj}
\node[node,right=1.8cm of d3] (d4) {};
\node[node,right=0.7cm of d4] (d5) {};
\path (d4) edge[loop above, looseness=17] (d4)
      (d4) edge (d5)
      (d5) edge[loop above, looseness=17] (d5);
\node[below=0.2cm of d4, xshift=0.6cm] {$I_{16}$};

% I17 = W_{ii}^2 W_{ij}
\node[node,right=1.6cm of d5] (d6) {};
\node[node,right=0.7cm of d6] (d7) {};
\path (d7) edge[loop above, looseness=17] (d7)
      (d7) edge[loop right, looseness=17] (d7)
      (d7) edge (d6);
\node[below=0.2cm of d6, xshift=0.6cm] {$I_{17}$};

% I18 = W_{ji} W_{ii}^2
\node[node,right=1.6cm of d7] (d8) {};
\node[node,right=0.7cm of d8] (d9) {};
\path (d9) edge[loop above, looseness=17] (d9)
      (d9) edge[loop right, looseness=17] (d9)
      (d8) edge (d9);
\node[below=0.2cm of d8, xshift=0.6cm] {$I_{18}$};

%%% Row 2: I19 -- I23 %%%
% I19 = W_{ii} W_{ij}^2
\node[node,below=2cm of d1] (d10) {};
\node[node,right=0.7cm of d10] (d11) {};
\path (d10) edge[loop above, looseness=17] (d10)
      (d10) edge[bend left=20] (d11)
      (d10) edge[bend right=20] (d11);
\node[below=0.2cm of d10, xshift=0.6cm] {$I_{19}$};

% I20 = W_{ji} W_{ii} W_{ij}
\node[node,right=1.6cm of d11] (d12) {};
\node[node,right=0.7cm of d12] (d13) {};
\path (d12) edge[loop above, looseness=17] (d12)
      (d12) edge[bend left=20] (d13)
      (d13) edge[bend left=20] (d12);
\node[below=0.2cm of d12, xshift=0.6cm] {$I_{20}$};

% I21 = W_{ji}^2 W_{ii}
\node[node,right=1.6cm of d13] (d14) {};
\node[node,right=0.7cm of d14] (d15) {};
\path (d14) edge[loop above, looseness=17] (d14)
      (d15) edge[bend left=20] (d14)
      (d15) edge[bend right=20] (d14);
\node[below=0.2cm of d14, xshift=0.6cm] {$I_{21}$};

% I22 = W_{ij}^3
\node[node,right=1.6cm of d15] (d16) {};
\node[node,right=0.7cm of d16] (d17) {};
\path (d16) edge[bend left=40] (d17)
      (d16) edge (d17)
      (d16) edge[bend right=40] (d17);
\node[below=0.2cm of d16, xshift=0.6cm] {$I_{22}$};

% I23 = W_{ij}^2 W_{ji}
\node[node,right=1.6cm of d17] (d18) {};
\node[node,right=0.7cm of d18] (d19) {};
\path (d18) edge[bend left=40] (d19)
      (d18) edge[bend right=40] (d19)
      (d19) edge (d18);
\node[below=0.2cm of d18, xshift=0.6cm] {$I_{23}$};

%%% Row 3: I24 -- I27 %%%
% I24 = W_{ii} W_{jk} W_{lm}
\node[node,below=2cm of d10] (d20) {};
\node[node,right=0.5cm of d20] (d21) {};
\node[node,right=0.7cm of d21] (d22) {};
\node[node,below=0.3cm of d21] (d23) {};
\node[node,right=0.7cm of d23] (d24) {};
\path (d20) edge[loop above, looseness=17] (d20)
      (d21) edge (d22)
      (d23) edge (d24);
\node[below=0.2cm of d20] {$I_{24}$};

% I25 = W_{ij} W_{jk} W_{lm}
\node[node,right=1.4cm of d22] (d25) {};
\node[node,right=0.7cm of d25] (d26) {};
\node[node,right=0.7cm of d26] (d27) {};
\node[node,below=0.3cm of d26] (d28) {};
\node[node,right=0.7cm of d28] (d29) {};
\path (d25) edge (d26)
      (d26) edge (d27)
      (d28) edge (d29);
\node[below=0.2cm of d25, xshift=0.35cm] {$I_{25}$};

% I26 = W_{ij} W_{kj} W_{lm}
\node[node,right=1.4cm of d27] (d30) {};
\node[node,right=0.7cm of d30] (d31) {};
\node[node,right=0.7cm of d31] (d32) {};
\node[node,below=0.3cm of d31] (d33) {};
\node[node,right=0.7cm of d33] (d34) {};
\path (d30) edge (d31)
      (d32) edge (d31)
      (d33) edge (d34);
\node[below=0.2cm of d30, xshift=0.35cm] {$I_{26}$};

% I27 = W_{ji} W_{jk} W_{lm}
\node[node,right=1.4cm of d32] (d35) {};
\node[node,right=0.7cm of d35] (d36) {};
\node[node,right=0.7cm of d36] (d37) {};
\node[node,below=0.3cm of d36] (d38) {};
\node[node,right=0.7cm of d38] (d39) {};
\path (d36) edge (d35)
      (d36) edge (d37)
      (d38) edge (d39);
\node[below=0.2cm of d35, xshift=0.35cm] {$I_{27}$};

%%% Row 4: I28 %%%
% I28 = W_{ij} W_{kl} W_{mn}
\node[node,below=1.2cm of d28, xshift=-0.9cm] (d40) {};
\node[node,right=0.7cm of d40] (d41) {};
\node[node,right=0.5cm of d41] (d42) {};
\node[node,right=0.7cm of d42] (d43) {};
\node[node,right=0.5cm of d43] (d44) {};
\node[node,right=0.7cm of d44] (d45) {};
\path (d40) edge (d41)
      (d42) edge (d43)
      (d44) edge (d45);
\node[below=0.2cm of d42, xshift=0.6cm] {$I_{28}$};
\end{tikzpicture}
\caption{Graphical representation of the cubic ($I_{14}$–$I_{28}$) invariants. Each index corresponds to a node, and each $W_{ij}$ to a directed edge from $i$ to $j$.}
\label{fig:C_invariant_graphs}
\end{figure}

%%%%%%%%%%%%%%%%%%%%%%%%%%%%%%%%%%%%%%%%%%%%%%%%%
\newpage
\section{Invariant Analytic Calculations}\label{app:inv_analytics}
To facilitate meaningful comparison of the observed invariant values across the initialisations/layers, expected analytic values of the invariants are calculated.
Since the network architecture is consistent, and the input layer is ignored in this study to maintain focusing on the square $10$ neuron layers, all layer weight matrices are the same shape ($10 \times 10$); therefore these results apply consistently across the layers.
However, as the two initialisations schemes lead to different weight distributions the results will differ here.

This section shows the calculation of each linear and quadratic expectation value, with their respective standard errors, for general initialisation scheme (allowing application to other schemes).
The results are then evaluated for the two initialisation schemes considered.

The two initialisation schemes considered, as introduced in §\ref{sec:data}, are the typical industry standards based on Gaussian and Uniform distributions.
The functions, repeated from \eqref{eq:gauss_init} \& \eqref{eq:uniform_init}, are:
\begin{align}
    f_{\text{Gaussian}}(w) & = \sqrt{\frac{d_{\text{in}}}{2\pi}}\text{exp}\bigg( -\frac{d_{\text{in}}w^2}{2} \bigg) \;,\\
    f_{\text{Uniform}}(w) & = \begin{cases} \frac{\sqrt{d_{\text{in}}}}{2} & \text{if } -\frac{1}{\sqrt{d_{\text{in}}}}\leq w \leq \frac{1}{\sqrt{d_{\text{in}}}}\\ 0,              & \text{otherwise} \end{cases}\;,
\end{align}
where the layer input size $d_{\text{in}}=10$ for all the layers considered.

The invariant expectations values reduce to functions of expectation values for powers of sampled weights, $w^{k\leq 4}$, using these distributions.
Defined generally as
\begin{equation}
    \langle w^k \rangle = \int_{-\infty}^{\infty} w^k f(w) \ dw
\end{equation}
which for the Gaussian distribution \eqref{eq:gauss_init} evaluate to
\begin{align}
    \langle w \rangle_{\text{Gaussian}} & = 0\;,\label{eq:gauss_w_exp_start}\\
    \langle w^2 \rangle_{\text{Gaussian}} & = \frac{1}{d_{\text{in}}}\;,\\
    \langle w^3 \rangle_{\text{Gaussian}} & = 0\;,\\
    \langle w^4 \rangle_{\text{Gaussian}} & = \frac{3}{d_{\text{in}}^2}\;,\label{eq:gauss_w_exp_final}
\end{align}
and for the Uniform distribution \eqref{eq:uniform_init} to
\begin{align}
    \langle w \rangle_{\text{Uniform}} & = 0\;,\label{eq:uniform_w_exp_start}\\
    \langle w^2 \rangle_{\text{Uniform}} & = \frac{1}{3d_{\text{in}}}\;,\\
    \langle w^3 \rangle_{\text{Uniform}} & = 0\;,\\
    \langle w^4 \rangle_{\text{Uniform}} & = \frac{1}{5d_{\text{in}}^2}\;.\label{eq:uniform_w_exp_final}
\end{align}
Equaling zero for odd powers since the initialisation distributions are both even.

The invariants are then functions of these weights, and hence their expectation values functions of the drawn weight sample's expectation values, for various powers based on both the invariant order and the statistic order.
In this section we show the calculation of the expectation values (1st order) and standard error (2nd order via the variance), for the linear (1st order) and quadratic (2nd order) invariants; however the methods can be straightforwardly generalised and applied for higher order invariants.

In the invariant computations, some standard statistical laws are used, for describing expectation values (denoted $\langle  \cdot  \rangle$), for samples of \textit{independent} random variables, $X_i$, and functions thereof.
They are stated here for clarity, using a general multiplicative constant $a$.
\begin{align}\label{eq:exp_rules}
    \langle aX_i \rangle & = a\langle X_i \rangle\;,\\
    \langle \Sigma_i X_i \rangle & = \Sigma_i \langle X_i \rangle\;,\label{eq:exp_rules_middle}\\
    \langle \Pi_{i}X_i \rangle & = \Pi_i\langle X_i \rangle\;,\label{eq:exp_rules_final}
\end{align}
and for variances, $\text{Var}(\cdot) = \langle \ \cdot^2 \rangle - \langle \cdot \rangle^2$, of \textit{independent} random variables they are
\begin{align}\label{eq:var_rules}
    \text{Var}(aX_i) & = a^2 \text{Var}(X_i)\;,\\
    \text{Var}(\Sigma_i X_i) & = \Sigma_i \text{Var}(X_i)\;,\label{eq:var_rules_middle}\\
    \text{Var}(\Pi_i X_i) & = \Pi_i (\text{Var}(X_i) + \langle X_i \rangle^2) - \Pi_i \langle X_i \rangle^2\;\label{eq:var_rules_final}\\
    & = \Pi_i \text{Var}(X_i)\;, \qquad \text{...where } \langle X_i \rangle=0\;,\label{eq:var_rules_simplify}
\end{align}
noting that in the last line, since $\langle w \rangle = 0$ for the even initialisation distributions considered, the equation can be significantly reduced.
From the variances, standard deviations are computed as $\hat{\sigma}(f(X_i)) = \sqrt{\text{Var}(f(X_i))}$.
Since these invariants are then averaged over the $N=1000$ runs, the variational measure of interest is the standard error of the average invariant expectation values, which is computed in terms of the invariants' standard deviations $\hat{\sigma}(I_i)$ via the variances by 
\begin{equation}
\begin{split}
    \text{Var}\bigg(\frac{1}{N}\Sigma_{r=1}^N I_i \bigg) & = \frac{1}{N^2}\text{Var}(\Sigma_{r=1}^N I_i)\;,\\
    & = \frac{1}{N^2}\Sigma_{r=1}^N \text{Var}(I_i)\;,\\
    & = \frac{1}{N}\text{Var}(I_i)\;,
\end{split}
\end{equation}
to give the standard error $\hat{\sigma}_{SE}$ as
\begin{equation}\label{eq:standarderror}
\begin{split}
    \hat{\sigma}_{SE}(I_i) & = \sqrt{\text{Var}\bigg(\frac{1}{N}\Sigma_{r=1}^N I_i \bigg)}\;,\\
    & = \frac{1}{\sqrt{N}} \hat{\sigma}(I_i)\;,
\end{split}
\end{equation}
noting that the sum is over the $N$ runs, within which each $I_i$ are independent $\forall \ r$.

\subsection{Expectation Values}
There are 2 linear and 11 quadratic invariants, as introduced in §\ref{sec:gmms}.
Their expectation values for the NN weight matrices considered (for the predefined architecture with $d_{\text{in}}=10$) are computed here; this largely relies on identifying where products in the functions are of dependent variables and separating those out.\\

\noindent\textit{Linear:}
\begin{equation}
\begin{split}
    \langle I_1 \rangle & = \langle \Sigma_{i=1}^{d_{\text{in}}} W_{ii} \rangle\;,\\
    & = \Sigma_{i=1}^{d_{\text{in}}} \langle W_{ii} \rangle\;,\\
    & = d_{\text{in}} \langle w \rangle\;,   
\end{split}
\end{equation}
\begin{equation}
\begin{split}
    \langle I_2 \rangle & = \langle \Sigma_{i,j=1}^{d_{\text{in}}} W_{ij} \rangle\;,\\
    & = \Sigma_{i,j=1}^{d_{\text{in}}} \langle W_{ij} \rangle\;,\\
    & = d_{\text{in}}^2 \langle w \rangle\;,  
\end{split}
\end{equation}

\noindent\textit{Quadratic:}
\begin{equation}
\begin{split}
    \langle I_3 \rangle & = \langle \Sigma_{i,j=1}^{d_{\text{in}}} W^2_{ij} \rangle\;,\\
    & = \Sigma_{i,j=1}^{d_{\text{in}}} \langle W^2_{ij} \rangle\;,\\
    & = d_{\text{in}}^2 \langle w^2 \rangle\;,
\end{split}
\end{equation}
\begin{equation}
\begin{split}
    \langle I_4 \rangle & = \langle \Sigma_{i,j=1}^{d_{\text{in}}} W_{ij}W_{ji} \rangle\;,\\
    & = \Sigma_{i,j=1|i=j}^{d_{\text{in}}} \langle W_{ii}^2 \rangle + \Sigma_{i,j=1|i \neq j}^{d_{\text{in}}} \langle W_{ij}W_{ji} \rangle\;,\\
    & = \Sigma_{i,j=1|i=j}^{d_{\text{in}}} \langle W_{ii}^2 \rangle + \Sigma_{i,j=1|i \neq j}^{d_{\text{in}}} \langle W_{ij} \rangle\langle W_{ji} \rangle\;,\\
    & = d_{\text{in}} \langle w^2 \rangle + d_{\text{in}}(d_{\text{in}}-1) \langle w \rangle^2\;,
\end{split}
\end{equation}
\begin{equation}
\begin{split}
    \langle I_5 \rangle & = \langle \Sigma_{i,j=1}^{d_{\text{in}}} W_{ii}W_{ij} \rangle\;,\\
    & = \Sigma_{i,j=1|i=j}^{d_{\text{in}}} \langle W_{ii}^2 \rangle + \Sigma_{i,j=1|i \neq j}^{d_{\text{in}}} \langle W_{ii}W_{ij} \rangle\;,\\
    & = d_{\text{in}} \langle w^2 \rangle + d_{\text{in}}(d_{\text{in}}-1) \langle w \rangle^2\;,
\end{split}
\end{equation}
\begin{equation}
\begin{split}
    \langle I_6 \rangle & = \langle \Sigma_{i,j=1}^{d_{\text{in}}} W_{ii}W_{ji} \rangle\;,\\
    & = \Sigma_{i,j=1|i=j}^{d_{\text{in}}} \langle W_{ii}^2 \rangle + \Sigma_{i,j=1|i \neq j}^{d_{\text{in}}} \langle W_{ii}W_{ji} \rangle\;,\\
    & = d_{\text{in}} \langle w^2 \rangle + d_{\text{in}}(d_{\text{in}}-1) \langle w \rangle^2\;,
\end{split}
\end{equation}
\begin{equation}
\begin{split}
    \langle I_7 \rangle & = \langle \Sigma_{i,j,k=1}^{d_{\text{in}}} W_{ij}W_{ik} \rangle\;,\\
    & = \Sigma_{i,j,k=1|j=k}^{d_{\text{in}}} \langle W_{ij}^2 \rangle + \Sigma_{i,j,k=1|j \neq k}^{d_{\text{in}}} \langle W_{ij}W_{ik} \rangle\;,\\
    & = d_{\text{in}}^2 \langle w^2 \rangle + d_{\text{in}}^2(d_{\text{in}}-1) \langle w \rangle^2\;,
\end{split}
\end{equation}
\begin{equation}
\begin{split}
    \langle I_8 \rangle & = \langle \Sigma_{i,j,k=1}^{d_{\text{in}}} W_{ij}W_{kj} \rangle\;,\\
    & = \Sigma_{i,j,k=1|i=k}^{d_{\text{in}}} \langle W_{ij}^2 \rangle + \Sigma_{i,j,k=1|i \neq k}^{d_{\text{in}}} \langle W_{ij}W_{kj} \rangle\;,\\
    & = d_{\text{in}}^2 \langle w^2 \rangle + d_{\text{in}}^2(d_{\text{in}}-1) \langle w \rangle^2\;,
\end{split}
\end{equation}
\begin{equation}
\begin{split}
    \langle I_9 \rangle & = \langle \Sigma_{i,j,k=1}^{d_{\text{in}}} W_{ij}W_{jk} \rangle\;,\\
    & = \Sigma_{i,j,k=1|i=j=k}^{d_{\text{in}}} \langle W_{ii}^2 \rangle + \Sigma_{i,j,k=1|\neg (i=j=k)}^{d_{\text{in}}} \langle W_{ij}W_{jk} \rangle\;,\\
    & = d_{\text{in}} \langle w^2 \rangle + (d_{\text{in}}^3-d_{\text{in}}) \langle w \rangle^2\;,
\end{split}
\end{equation}
%...wrote as \neg i=j=k, but could also write as: i \neq j \text{ or } i \neq k \text{ or } j \neq k
\begin{equation}
\begin{split}
    \langle I_{10} \rangle & = \langle \Sigma_{i,j,k,l=1}^{d_{\text{in}}} W_{ij}W_{kl} \rangle\;,\\
    & = \Sigma_{i,j,k,l=1|i=k \& j=l}^{d_{\text{in}}} \langle W_{ij}^2 \rangle + \Sigma_{i,j,k,l=1|\neg (i=k \& j=l)}^{d_{\text{in}}} \langle W_{ij}W_{kl} \rangle\;,\\
    & = d_{\text{in}}^2 \langle w^2 \rangle + (d_{\text{in}}^4-d_{\text{in}}^2) \langle w \rangle^2\;,
\end{split}
\end{equation}
\begin{equation}
\begin{split}
    \langle I_{11} \rangle & = \langle \Sigma_{i=1}^{d_{\text{in}}} W_{ii}^2 \rangle\;,\\
    & = \Sigma_{i=1}^{d_{\text{in}}} \langle W_{ii}^2 \rangle\;,\\
    & = d_{\text{in}} \langle w^2 \rangle\;,   
\end{split}
\end{equation}
\begin{equation}
\begin{split}
    \langle I_{12} \rangle & = \langle \Sigma_{i,j=1}^{d_{\text{in}}} W_{ii}W_{jj} \rangle\;,\\
    & = \Sigma_{i,j=1|i=j}^{d_{\text{in}}} \langle W_{ii}^2 \rangle + \Sigma_{i,j=1|i \neq j}^{d_{\text{in}}} \langle W_{ii}W_{jj} \rangle\;,\\
    & = d_{\text{in}} \langle w^2 \rangle + d_{\text{in}}(d_{\text{in}}-1) \langle w \rangle^2\;,
\end{split}
\end{equation}
\begin{equation}
\begin{split}
    \langle I_{13} \rangle & = \langle \Sigma_{i,j,k=1}^{d_{\text{in}}} W_{ii}W_{jk} \rangle\;,\\
    & = \Sigma_{i,j,k=1|i=j=k}^{d_{\text{in}}} \langle W_{ii}^2 \rangle + \Sigma_{i,j,k=1|\neg (i=j=k)}^{d_{\text{in}}} \langle W_{ii}W_{jk} \rangle\;,\\
    & = d_{\text{in}} \langle w^2 \rangle + (d_{\text{in}}^3-d_{\text{in}}) \langle w \rangle^2\;.
\end{split}
\end{equation}
%...as a check one notes that the coefficients of the sum decompositions still add up to $d_{\text{in}}^{\# \text{indices}}$.

We note that the invariant expectation values are equal for: $\langle I_4 \rangle = \langle I_5 \rangle = \langle I_6 \rangle = \langle I_{12} \rangle$, $\langle I_7 \rangle = \langle I_8 \rangle$, and $\langle I_9 \rangle = \langle I_{13} \rangle$.

For the investigations carried out in this work, $d_{\text{in}}=10$, and $\langle w^k \rangle$ are given as in \eqref{eq:gauss_w_exp_start}-\eqref{eq:gauss_w_exp_final} and \eqref{eq:uniform_w_exp_start}-\eqref{eq:uniform_w_exp_final} for the Gaussian and Uniform initialisation schemes respectively.
With these values substituted, the final expectation values are given in Table \ref{tab:inv_analytic_subbed}, noting there are now more accidental equalities.

\begin{table}[!t]
\centering
\addtolength{\leftskip}{-1cm}
\addtolength{\rightskip}{-1cm}
\begin{tabular}{|cc|cc|ccccccccccc|}
\hline
\multicolumn{2}{|c|}{\multirow{2}{*}{Invariant}}                                                 & \multicolumn{2}{c|}{Linear}        & \multicolumn{11}{c|}{Quadratic} \\ \cline{3-15} 
\multicolumn{2}{|c|}{}                                                                           & \multicolumn{1}{c|}{$I_1$} & $I_2$ & \multicolumn{1}{c|}{$I_3$} & \multicolumn{1}{c|}{$I_4$} & \multicolumn{1}{c|}{$I_5$} & \multicolumn{1}{c|}{$I_6$} & \multicolumn{1}{c|}{$I_7$} & \multicolumn{1}{c|}{$I_8$} & \multicolumn{1}{c|}{$I_9$} & \multicolumn{1}{c|}{$I_{10}$} & \multicolumn{1}{c|}{$I_{11}$} & \multicolumn{1}{c|}{$I_{12}$} & $I_{13}$ \\ \hline
\multicolumn{1}{|c|}{\multirow{2}{*}{\begin{tabular}[c]{@{}c@{}}Init\\ Scheme\end{tabular}}} & G & \multicolumn{1}{c|}{0}     & 0     & \multicolumn{1}{c|}{10}      & \multicolumn{1}{c|}{1}      & \multicolumn{1}{c|}{1}      & \multicolumn{1}{c|}{1}      & \multicolumn{1}{c|}{10}      & \multicolumn{1}{c|}{10}      & \multicolumn{1}{c|}{1}      & \multicolumn{1}{c|}{10}         & \multicolumn{1}{c|}{1}         & \multicolumn{1}{c|}{1}         &     1     \\ \cline{2-15} 
\multicolumn{1}{|c|}{}                                                                       & U & \multicolumn{1}{c|}{0}     & 0     & \multicolumn{1}{c|}{$\frac{10}{3}$}      & \multicolumn{1}{c|}{$\frac{1}{3}$}      & \multicolumn{1}{c|}{$\frac{1}{3}$}      & \multicolumn{1}{c|}{$\frac{1}{3}$}      & \multicolumn{1}{c|}{$\frac{10}{3}$}      & \multicolumn{1}{c|}{$\frac{10}{3}$}      & \multicolumn{1}{c|}{$\frac{1}{3}$}      & \multicolumn{1}{c|}{$\frac{10}{3}$}         & \multicolumn{1}{c|}{$\frac{1}{3}$}         & \multicolumn{1}{c|}{$\frac{1}{3}$}         &   $\frac{1}{3}$       \\ \hline
\end{tabular}
\caption{Expected Values for the linear and quadratic invariants for all layers, shown for the 2 initialisation schemes considered: Gaussian (G) and Uniform (U) respectively. Specifically these values have the explicit $d_{\text{in}}$ and $\langle w^k \rangle$ values substituted.}
\label{tab:inv_analytic_subbed}
\end{table}

\subsection{Standard Errors}
To now compute the standard errors of these 2 linear and 11 quadratic invariants, an equivalent calculation procedure is followed separating out terms which are products of dependent variables, then using the variance propagation rules for independent variables as dictated in \eqref{eq:var_rules}-\eqref{eq:var_rules_final}, and that 
\begin{equation}
\begin{split}
    \text{Var}(X_i^2) & = \langle (X_i^2)^2 \rangle - ( \langle X_i^2 \rangle )^2\;,\\
    & = \langle X_i^4 \rangle - \langle X_i^2 \rangle^2\;,
\end{split}
\end{equation}
for the terms which are products of the same (i.e. dependent) variables.
To simplify the presentation, we specify our calculations to the case $\langle w \rangle = 0$, applicable to both initialisations and all the work carried out here; this allows us to use $\text{Var}(w) = \langle w^2 \rangle - \langle w \rangle^2 = \langle w^2 \rangle$; as well as the simplification \eqref{eq:var_rules_simplify} such that $\text{Var}(\Pi_i w) = \Pi_i \text{Var}(w) = \Pi_i \langle w^2 \rangle$ for independent $w$.

Here we show in each case the computation of the variance, standard deviation, and then the standard error (from systematically applying \eqref{eq:standarderror}).\\

\noindent\textit{Linear:}
\begin{equation}
\begin{split}
    \text{Var}(I_1) & = \text{Var}(\Sigma_{i=1}^{d_{\text{in}}} W_{ii})\;,\\
    & = \Sigma_{i=1}^{d_{\text{in}}}\text{Var}(W_{ii})\;,\\
    & = d_{\text{in}} \text{Var}(w)\;,\\
    & = d_{\text{in}} \langle w^2 \rangle\;,\\
    \implies \quad\, \hat\sigma(I_1) & = \sqrt{d_{\text{in}} \langle w^2 \rangle}\;,\\
    \implies \hat\sigma_{SE}(I_1) & = \sqrt{\frac{d_{\text{in}} \langle w^2 \rangle}{N}}\;,
\end{split}
\end{equation}
\begin{equation}
\begin{split}
    \text{Var}(I_2) & = \text{Var}(\Sigma_{i,j=1}^{d_{\text{in}}} W_{ij})\;,\\
    & = \Sigma_{i,j=1}^{d_{\text{in}}}\text{Var}(W_{ij})\;,\\
    & = d_{\text{in}}^2 \langle w^2 \rangle\;,\\
    \implies \quad\, \hat\sigma(I_2) & = d_{\text{in}}\sqrt{\langle w^2 \rangle}\;,\\
    \implies \hat\sigma_{SE}(I_2) & = d_{\text{in}}\sqrt{\frac{\langle w^2 \rangle}{N}}\;,
\end{split}
\end{equation}

\noindent\textit{Quadratic:}
\begin{equation}
\begin{split}
    \text{Var}(I_3) & = \text{Var}(\Sigma_{i,j=1}^{d_{\text{in}}} W_{ij}^2)\;,\\
    & = \Sigma_{i,j=1}^{d_{\text{in}}}\text{Var}(W_{ij}^2)\;,\\
    & = d_{\text{in}}^2 \text{Var}(w^2)\;,\\
    & = d_{\text{in}}^2 (\langle w^4 \rangle - \langle w^2 \rangle^2)\;,\\
    \implies \quad\, \hat\sigma(I_3) & = d_{\text{in}}\sqrt{\langle w^4 \rangle - \langle w^2 \rangle^2}\;,\\
    \implies \hat\sigma_{SE}(I_3) & = d_{\text{in}}\sqrt{\frac{\langle w^4 \rangle - \langle w^2 \rangle^2}{N}}\;,
\end{split}
\end{equation}
\begin{equation}
\begin{split}
    \text{Var}(I_4) & = \text{Var}(\Sigma_{i,j=1}^{d_{\text{in}}} W_{ij}W_{ji})\;,\\
    & = \Sigma_{i,j=1|i=j}^{d_{\text{in}}}\text{Var}(W_{ii}^2) + \Sigma_{i,j=1|i \neq j}^{d_{\text{in}}} \text{Var}(W_{ij}W_{ji}) \;,\\
    & = d_{\text{in}} \text{Var}(w^2) + d_{\text{in}}(d_{\text{in}}-1) \text{Var}(w)^2\;,\\
    & = d_{\text{in}} (\langle w^4 \rangle - \langle w^2 \rangle^2) + d_{\text{in}}(d_{\text{in}}-1) \langle w^2 \rangle^2\;,\\
    & = d_{\text{in}} \langle w^4 \rangle + d_{\text{in}}(d_{\text{in}}-2) \langle w^2 \rangle^2\;,\\
    \implies \quad\, \hat\sigma(I_4) & = \sqrt{d_{\text{in}} \langle w^4 \rangle + d_{\text{in}}(d_{\text{in}}-2) \langle w^2 \rangle^2}\;,\\
    \implies \hat\sigma_{SE}(I_4) & = \sqrt{\frac{d_{\text{in}} \langle w^4 \rangle + d_{\text{in}}(d_{\text{in}}-2) \langle w^2 \rangle^2}{N}}\;,
\end{split}
\end{equation}
\begin{equation}
\begin{split}
    \text{Var}(I_5) & = \text{Var}(\Sigma_{i,j=1}^{d_{\text{in}}} W_{ii}W_{ij})\;,\\
    & = \Sigma_{i,j=1|i=j}^{d_{\text{in}}}\text{Var}(W_{ii}^2) + \Sigma_{i,j=1|i \neq j}^{d_{\text{in}}} \text{Var}(W_{ii}W_{ij}) \;,\\
    & = d_{\text{in}} \text{Var}(w^2) + d_{\text{in}}(d_{\text{in}}-1) \text{Var}(w)^2\;,\\
    & = d_{\text{in}} (\langle w^4 \rangle - \langle w^2 \rangle^2) + d_{\text{in}}(d_{\text{in}}-1) \langle w^2 \rangle^2\;,\\
    & = d_{\text{in}} \langle w^4 \rangle + d_{\text{in}}(d_{\text{in}}-2) \langle w^2 \rangle^2\;,\\
    \implies \quad\, \hat\sigma(I_5) & = \sqrt{d_{\text{in}} \langle w^4 \rangle + d_{\text{in}}(d_{\text{in}}-2) \langle w^2 \rangle^2}\;,\\
    \implies \hat\sigma_{SE}(I_5) & = \sqrt{\frac{d_{\text{in}} \langle w^4 \rangle + d_{\text{in}}(d_{\text{in}}-2) \langle w^2 \rangle^2}{N}}\;,
\end{split}
\end{equation}
\begin{equation}
\begin{split}
    \text{Var}(I_6) & = \text{Var}(\Sigma_{i,j=1}^{d_{\text{in}}} W_{ii}W_{ji})\;,\\
    & = \Sigma_{i,j=1|i=j}^{d_{\text{in}}}\text{Var}(W_{ii}^2) + \Sigma_{i,j=1|i \neq j}^{d_{\text{in}}} \text{Var}(W_{ii}W_{ji}) \;,\\
    & = d_{\text{in}} \text{Var}(w^2) + d_{\text{in}}(d_{\text{in}}-1) \text{Var}(w)^2\;,\\
    & = d_{\text{in}} (\langle w^4 \rangle - \langle w^2 \rangle^2) + d_{\text{in}}(d_{\text{in}}-1) \langle w^2 \rangle^2\;,\\
    & = d_{\text{in}} \langle w^4 \rangle + d_{\text{in}}(d_{\text{in}}-2) \langle w^2 \rangle^2\;,\\
    \implies \quad\, \hat\sigma(I_6) & = \sqrt{d_{\text{in}} \langle w^4 \rangle + d_{\text{in}}(d_{\text{in}}-2) \langle w^2 \rangle^2}\;,\\
    \implies \hat\sigma_{SE}(I_6) & = \sqrt{\frac{d_{\text{in}} \langle w^4 \rangle + d_{\text{in}}(d_{\text{in}}-2) \langle w^2 \rangle^2}{N}}\;,
\end{split}
\end{equation}
\begin{equation}
\begin{split}
    \text{Var}(I_7) & = \text{Var}(\Sigma_{i,j,k=1}^{d_{\text{in}}} W_{ij}W_{ik})\;,\\
    & = \Sigma_{i,j,k=1|j=k}^{d_{\text{in}}}\text{Var}(W_{ij}^2) + \Sigma_{i,j,k=1|j \neq k}^{d_{\text{in}}} \text{Var}(W_{ij}W_{ik}) \;,\\
    & = d_{\text{in}}^2 \text{Var}(w^2) + d_{\text{in}}^2(d_{\text{in}}-1) \text{Var}(w)^2\;,\\
    & = d_{\text{in}}^2 (\langle w^4 \rangle - \langle w^2 \rangle^2) + d_{\text{in}}^2(d_{\text{in}}-1) \langle w^2 \rangle^2\;,\\
    & = d_{\text{in}}^2 \langle w^4 \rangle + d_{\text{in}}^2(d_{\text{in}}-2) \langle w^2 \rangle^2\;,\\
    \implies \quad\, \hat\sigma(I_7) & = \sqrt{d_{\text{in}}^2 \langle w^4 \rangle + d_{\text{in}}^2(d_{\text{in}}-2) \langle w^2 \rangle^2}\;,\\
    \implies \hat\sigma_{SE}(I_7) & = \sqrt{\frac{d_{\text{in}}^2 \langle w^4 \rangle + d_{\text{in}}^2(d_{\text{in}}-2) \langle w^2 \rangle^2}{N}}\;,
\end{split}
\end{equation}
\begin{equation}
\begin{split}
    \text{Var}(I_8) & = \text{Var}(\Sigma_{i,j,k=1}^{d_{\text{in}}} W_{ij}W_{kj})\;,\\
    & = \Sigma_{i,j,k=1|i=k}^{d_{\text{in}}}\text{Var}(W_{ij}^2) + \Sigma_{i,j,k=1|i \neq k}^{d_{\text{in}}} \text{Var}(W_{ij}W_{kj}) \;,\\
    & = d_{\text{in}}^2 \text{Var}(w^2) + d_{\text{in}}^2(d_{\text{in}}-1) \text{Var}(w)^2\;,\\
    & = d_{\text{in}}^2 (\langle w^4 \rangle - \langle w^2 \rangle^2) + d_{\text{in}}^2(d_{\text{in}}-1) \langle w^2 \rangle^2\;,\\
    & = d_{\text{in}}^2 \langle w^4 \rangle + d_{\text{in}}^2(d_{\text{in}}-2) \langle w^2 \rangle^2\;,\\
    \implies \quad\, \hat\sigma(I_8) & = \sqrt{d_{\text{in}}^2 \langle w^4 \rangle + d_{\text{in}}^2(d_{\text{in}}-2) \langle w^2 \rangle^2}\;,\\
    \implies \hat\sigma_{SE}(I_8) & = \sqrt{\frac{d_{\text{in}}^2 \langle w^4 \rangle + d_{\text{in}}^2(d_{\text{in}}-2) \langle w^2 \rangle^2}{N}}\;,
\end{split}
\end{equation}
\begin{equation}
\begin{split}
    \text{Var}(I_9) & = \text{Var}(\Sigma_{i,j,k=1}^{d_{\text{in}}} W_{ij}W_{jk})\;,\\
    & = \Sigma_{i,j,k=1|i=j=k}^{d_{\text{in}}}\text{Var}(W_{ii}^2) + \Sigma_{i,j,k=1|\neg (i=j=k)}^{d_{\text{in}}} \text{Var}(W_{ij}W_{jk}) \;,\\
    & = d_{\text{in}} \text{Var}(w^2) + (d_{\text{in}}^3-d_{\text{in}}) \text{Var}(w)^2\;,\\
    & = d_{\text{in}} (\langle w^4 \rangle - \langle w^2 \rangle^2) + (d_{\text{in}}^3-d_{\text{in}}) \langle w^2 \rangle^2\;,\\
    & = d_{\text{in}} \langle w^4 \rangle + d_{\text{in}}(d_{\text{in}}^2-2) \langle w^2 \rangle^2\;,\\
    \implies \quad\, \hat\sigma(I_9) & = \sqrt{d_{\text{in}} \langle w^4 \rangle + d_{\text{in}}(d_{\text{in}}^2-2) \langle w^2 \rangle^2}\;,\\
    \implies \hat\sigma_{SE}(I_9) & = \sqrt{\frac{d_{\text{in}} \langle w^4 \rangle + d_{\text{in}}(d_{\text{in}}^2-2) \langle w^2 \rangle^2}{N}}\;,
\end{split}
\end{equation}
\begin{equation}
\begin{split}
    \text{Var}(I_{10}) & = \text{Var}(\Sigma_{i,j,k,l=1}^{d_{\text{in}}} W_{ij}W_{kl})\;,\\
    & = \Sigma_{i,j,k,l=1|i=k \& j=l}^{d_{\text{in}}}\text{Var}(W_{ij}^2) + \Sigma_{i,j,k,l=1|\neg (i=k \& j=l)}^{d_{\text{in}}} \text{Var}(W_{ij}W_{kl}) \;,\\
    & = d_{\text{in}}^2 \text{Var}(w^2) + (d_{\text{in}}^4-d_{\text{in}}^2) \text{Var}(w)^2\;,\\
    & = d_{\text{in}}^2 (\langle w^4 \rangle - \langle w^2 \rangle^2) + (d_{\text{in}}^4-d_{\text{in}}^2) \langle w^2 \rangle^2\;,\\
    & = d_{\text{in}}^2 \langle w^4 \rangle + d_{\text{in}}^2(d_{\text{in}}^2-2) \langle w^2 \rangle^2\;,\\
    \implies \quad\, \hat\sigma(I_{10}) & = \sqrt{d_{\text{in}}^2 \langle w^4 \rangle + d_{\text{in}}^2(d_{\text{in}}^2-2) \langle w^2 \rangle^2}\;,\\
    \implies \hat\sigma_{SE}(I_{10}) & = \sqrt{\frac{d_{\text{in}}^2 \langle w^4 \rangle + d_{\text{in}}^2(d_{\text{in}}^2-2) \langle w^2 \rangle^2}{N}}\;,
\end{split}
\end{equation}
\begin{equation}
\begin{split}
    \text{Var}(I_{11}) & = \text{Var}(\Sigma_{i=1}^{d_{\text{in}}} W_{ii}^2)\;,\\
    & = \Sigma_{i=1}^{d_{\text{in}}}\text{Var}(W_{ii}^2)\;,\\
    & = d_{\text{in}} \text{Var}(w^2)\;,\\
    & = d_{\text{in}} (\langle w^4 \rangle - \langle w^2 \rangle^2)\;,\\
    \implies \quad\, \hat\sigma(I_{11}) & = \sqrt{d_{\text{in}} (\langle w^4 \rangle - \langle w^2 \rangle^2)}\;,\\
    \implies \hat\sigma_{SE}(I_{11}) & = \sqrt{\frac{d_{\text{in}} (\langle w^4 \rangle - \langle w^2 \rangle^2)}{N}}\;,
\end{split}
\end{equation}
\begin{equation}
\begin{split}
    \text{Var}(I_{12}) & = \text{Var}(\Sigma_{i,j=1}^{d_{\text{in}}} W_{ii}W_{jj})\;,\\
    & = \Sigma_{i,j=1|i=j}^{d_{\text{in}}}\text{Var}(W_{ii}^2) + \Sigma_{i,j=1|i \neq j}^{d_{\text{in}}} \text{Var}(W_{ii}W_{jj}) \;,\\
    & = d_{\text{in}} \text{Var}(w^2) + d_{\text{in}}(d_{\text{in}}-1) \text{Var}(w)^2\;,\\
    & = d_{\text{in}} (\langle w^4 \rangle - \langle w^2 \rangle^2) + d_{\text{in}}(d_{\text{in}}-1) \langle w^2 \rangle^2\;,\\
    & = d_{\text{in}} \langle w^4 \rangle + d_{\text{in}}(d_{\text{in}}-2) \langle w^2 \rangle^2\;,\\
    \implies \quad\, \hat\sigma(I_{12}) & = \sqrt{d_{\text{in}} \langle w^4 \rangle + d_{\text{in}}(d_{\text{in}}-2) \langle w^2 \rangle^2}\;,\\
    \implies \hat\sigma_{SE}(I_{12}) & = \sqrt{\frac{d_{\text{in}} \langle w^4 \rangle + d_{\text{in}}(d_{\text{in}}-2) \langle w^2 \rangle^2}{N}}\;,
\end{split}
\end{equation}
\begin{equation}
\begin{split}
    \text{Var}(I_{13}) & = \text{Var}(\Sigma_{i,j,k=1}^{d_{\text{in}}} W_{ii}W_{jk})\;,\\
    & = \Sigma_{i,j,k=1|i=j=k}^{d_{\text{in}}}\text{Var}(W_{ii}^2) + \Sigma_{i,j,k=1|\neg (i=j=k)}^{d_{\text{in}}} \text{Var}(W_{ii}W_{jk}) \;,\\
    & = d_{\text{in}} \text{Var}(w^2) + (d_{\text{in}}^3-d_{\text{in}}) \text{Var}(w)^2\;,\\
    & = d_{\text{in}} (\langle w^4 \rangle - \langle w^2 \rangle^2) + (d_{\text{in}}^3-d_{\text{in}}) \langle w^2 \rangle^2\;,\\
    & = d_{\text{in}} \langle w^4 \rangle + d_{\text{in}}(d_{\text{in}}^2-2) \langle w^2 \rangle^2\;,\\
    \implies \quad\, \hat\sigma(I_{13}) & = \sqrt{d_{\text{in}} \langle w^4 \rangle + d_{\text{in}}(d_{\text{in}}^2-2) \langle w^2 \rangle^2}\;,\\
    \implies \hat\sigma_{SE}(I_{13}) & = \sqrt{\frac{d_{\text{in}} \langle w^4 \rangle + d_{\text{in}}(d_{\text{in}}^2-2) \langle w^2 \rangle^2}{N}}\;.
\end{split}
\end{equation}

We note that the invariant standard errors are equal for: $\hat\sigma(I_4) = \hat\sigma(I_5) = \hat\sigma(I_6) = \hat\sigma(I_{12})$, $\hat\sigma(I_7) = \hat\sigma(I_8)$, and $\hat\sigma(I_9) = \hat\sigma(I_{13})$.

For the investigations carried out in this work, $d_{\text{in}}=10$, $N=1000$, and $\langle w^k \rangle$ are given as in \eqref{eq:gauss_w_exp_start}-\eqref{eq:gauss_w_exp_final} and \eqref{eq:uniform_w_exp_start}-\eqref{eq:uniform_w_exp_final} for the Gaussian and Uniform initialisation schemes respectively.
With these values substituted, the final standard error values are given in Table \ref{tab:inv_analyticstdev_subbed}.

\begin{table}[!t]
\centering
\addtolength{\leftskip}{-1cm}
\addtolength{\rightskip}{-1cm}
\begin{tabular}{|cc|cc|cccccc|}
\hline
\multicolumn{2}{|c|}{\multirow{2}{*}{Invariant}}                                                 & \multicolumn{2}{c|}{Linear}        & \multicolumn{6}{c|}{Quadratic}                                                                                                                                     \\ \cline{3-10} 
\multicolumn{2}{|c|}{}                                                                           & \multicolumn{1}{c|}{$I_1$} & $I_2$ & \multicolumn{1}{c|}{$I_3$} & \multicolumn{1}{c|}{$I_4$}    & \multicolumn{1}{c|}{$I_5$}    & \multicolumn{1}{c|}{$I_6$}    & \multicolumn{1}{c|}{$I_7$}    & $I_8$ \\ \hline
\multicolumn{1}{|c|}{\multirow{2}{*}{\begin{tabular}[c]{@{}c@{}}Init\\ Scheme\end{tabular}}} & G & \multicolumn{1}{c|}{$\sqrt{\frac{1}{1000}}$}     & $\sqrt{\frac{1}{100}}$     & \multicolumn{1}{c|}{$\sqrt{\frac{2}{1000}}$}      & \multicolumn{1}{c|}{$\sqrt{\frac{11}{10000}}$}         & \multicolumn{1}{c|}{$\sqrt{\frac{11}{10000}}$}         & \multicolumn{1}{c|}{$\sqrt{\frac{11}{10000}}$}         & \multicolumn{1}{c|}{$\sqrt{\frac{11}{1000}}$}      &    $\sqrt{\frac{11}{1000}}$   \\ \cline{2-10} 
\multicolumn{1}{|c|}{}                                                                       & U & \multicolumn{1}{c|}{$\sqrt{\frac{1}{3000}}$}     & $\sqrt{\frac{1}{300}}$     & \multicolumn{1}{c|}{$\sqrt{\frac{4}{45000}}$}      & \multicolumn{1}{c|}{$\sqrt{\frac{49}{450000}}$}         & \multicolumn{1}{c|}{$\sqrt{\frac{49}{450000}}$}         & \multicolumn{1}{c|}{$\sqrt{\frac{49}{450000}}$}         & \multicolumn{1}{c|}{$\sqrt{\frac{49}{45000}}$}         &    $\sqrt{\frac{49}{45000}}$   \\ \hline
\multicolumn{2}{|c|}{Invariant}                                                                  & \multicolumn{2}{c|}{-}             & \multicolumn{1}{c|}{$I_9$} & \multicolumn{1}{c|}{$I_{10}$} & \multicolumn{1}{c|}{$I_{11}$} & \multicolumn{1}{c|}{$I_{12}$} & \multicolumn{1}{c|}{$I_{13}$} & -     \\ \hline
\multicolumn{1}{|c|}{\multirow{2}{*}{\begin{tabular}[c]{@{}c@{}}Init\\ Scheme\end{tabular}}} & G & \multicolumn{2}{c|}{-}              & \multicolumn{1}{c|}{$\sqrt{\frac{101}{10000}}$}      & \multicolumn{1}{c|}{$\sqrt{\frac{11}{1000}}$}         & \multicolumn{1}{c|}{$\sqrt{\frac{2}{10000}}$}         & \multicolumn{1}{c|}{$\sqrt{\frac{11}{10000}}$}         & \multicolumn{1}{c|}{$\sqrt{\frac{101}{10000}}$}   & -     \\ \cline{2-10} 
\multicolumn{1}{|c|}{}                                                                       & U & \multicolumn{2}{c|}{-}             & \multicolumn{1}{c|}{$\sqrt{\frac{499}{450000}}$}      & \multicolumn{1}{c|}{$\sqrt{\frac{499}{45000}}$}         & \multicolumn{1}{c|}{$\sqrt{\frac{4}{450000}}$}         & \multicolumn{1}{c|}{$\sqrt{\frac{49}{450000}}$}         & \multicolumn{1}{c|}{$\sqrt{\frac{499}{450000}}$}         & -     \\ \hline
\end{tabular}
\caption{Standard errors for the linear and quadratic invariants for both problems and all layers, shown for the 2 initialisation schemes considered: Gaussian (G) and Uniform (U) respectively. Specifically these values have the explicit $d_{\text{in}}$, $N$, and $\langle w^k \rangle$ values substituted.}
\label{tab:inv_analyticstdev_subbed}
\end{table}

%Match these tables to main text (combine these and add to main text table).

%%%%%%%%%%%%%%%%%%%%%%%%%%%%%%%%%%%%%%%%%%%%%%%%%
\section{Model Parameter Analytic Calculations}\label{app:param_analytics}
The model parameters are computed as implicit functions of the average invariants.
Their derivation was originally performed in 
\cite{Ramgoolam:2018xty}, and used quantum field theory inspired Gaussian integral techniques, they are restated in \cite{Ramgoolam:2019ldg} in the form of a system of 13 dependent equations\footnote{We note a notational difference here where the matrices we denote with $W_{ij}$ are $M_{ij}$, the matrix dimension we denote $d_{\text{in}}$ is $D$, and the model parameters we denote $f_i$ have a series of names $\{\tilde{\mu}_1, \tilde{\mu}_2, (\Lambda^{-1}_{V_0})_{11}, (\Lambda^{-1}_{V_0})_{12}, (\Lambda^{-1}_{V_0})_{22}, (\Lambda^{-1}_{V_H})_{11}, (\Lambda^{-1}_{V_H})_{12}, (\Lambda^{-1}_{V_H})_{13}, (\Lambda^{-1}_{V_H})_{22}, (\Lambda^{-1}_{V_H})_{23}, (\Lambda^{-1}_{V_H})_{33}, \Lambda^{-1}_{V_2}, \Lambda^{-1}_{V_3}\}$.}.

For this work, the system of equations was explicitly inverted to provide defining equations for these model parameters for the first time, they are thus defined:
\begin{align}\label{eq:model_param_formulas}
    f_1 & = \frac{\hat{I}_2}{d_{\text{in}}}\;,\\
    f_2 & = \frac{1}{d_{\text{in}}\sqrt{d_{\text{in}}-1}} \bigg( d_{\text{in}}\hat{I}_1 - \hat{I}_2 \bigg)\;,\\
    f_3 & = -\frac{1}{d_{\text{in}}^2} \bigg(\hat{I}_2^2 - \hat{I}_{10}\bigg)\;,\\
    f_4 & = -\frac{1}{d_{\text{in}}^2 \sqrt{d_{\text{in}}-1}} \bigg(d_{\text{in}}\hat{I}_1\hat{I}_2 - \hat{I}_2^2 + \hat{I}_{10} - d_{\text{in}}\hat{I}_{13} \bigg)\;,\\
    f_5 & = -\frac{1}{d_{\text{in}}^2 (d_{\text{in}}-1)} \bigg( d_{\text{in}}^2 \hat{I}_1^2 - 2 d_{\text{in}} \hat{I}_1\hat{I}_2 + \hat{I}_2^2 - \hat{I}_{10} - d_{\text{in}}^2 \hat{I}_{12} + 2 d_{\text{in}} \hat{I}_{13} \bigg)\;,\\
    f_6 & = \frac{1}{d_{\text{in}}^2 (d_{\text{in}}-1)} \bigg( d_{\text{in}}\hat{I}_8 - \hat{I}_{10} \bigg)\;,\\
    f_7 & = \frac{1}{d_{\text{in}}^2 (d_{\text{in}}-1)} \bigg( d_{\text{in}}\hat{I}_9 - \hat{I}_{10} \bigg)\;,\\
    f_8 & = \frac{1}{d_{\text{in}}^2(d_{\text{in}}-1)\sqrt{d_{\text{in}}-2}} \bigg( d_{\text{in}}^2\hat{I}_6 - d_{\text{in}}\hat{I}_8 - d_{\text{in}}\hat{I}_9 + 2\hat{I}_{10} - d_{\text{in}}\hat{I}_{13} \bigg)\;,\\
    f_9 & = \frac{1}{d_{\text{in}}^2 (d_{\text{in}}-1)} \bigg( d_{\text{in}}\hat{I}_7 - \hat{I}_{10} \bigg)\;,\\
    f_{10} & =  \frac{1}{d_{\text{in}}^2(d_{\text{in}}-1)\sqrt{d_{\text{in}}-2}} \bigg( d_{\text{in}}^2\hat{I}_5 - d_{\text{in}}\hat{I}_7  - d_{\text{in}}\hat{I}_9 + 2\hat{I}_{10} - d_{\text{in}}\hat{I}_{13} \bigg)\;,\\
    f_{11} & = -\frac{1}{d_{\text{in}}^2(d_{\text{in}}-1)(d_{\text{in}}-2)} \bigg( 
    2d_{\text{in}}^2\hat{I}_5 + 2d_{\text{in}}^2\hat{I}_6 - d_{\text{in}}\hat{I}_7 - d_{\text{in}}\hat{I}_8 \\
    & \qquad - 2d_{\text{in}}\hat{I}_9 +4\hat{I}_{10} - d_{\text{in}}^3\hat{I}_{11} + d_{\text{in}}^2\hat{I}_{12} - 4d_{\text{in}}\hat{I}_{13} \bigg)\;,\nonumber\\
    f_{12} & = \frac{1}{d_{\text{in}}(d_{\text{in}}-1)(d_{\text{in}}-2)(d_{\text{in}}-3)} \bigg( (d_{\text{in}}-1)(d_{\text{in}}-2)\hat{I}_3 \\
    & \qquad + (d_{\text{in}}-1)(d_{\text{in}}-2)\hat{I}_4 + 4(d_{\text{in}}-1)\hat{I}_5 + 4(d_{\text{in}}-1)\hat{I}_6 \nonumber\\
    & \qquad - (d_{\text{in}}-1)\hat{I}_7 - (d_{\text{in}}-1)\hat{I}_8 - 2(d_{\text{in}}-1)\hat{I}_9 + 2\hat{I}_{10} \nonumber\\
    & \qquad - 2d_{\text{in}}(d_{\text{in}}-1)\hat{I}_{11} + 2\hat{I}_{12} - 4\hat{I}_{13} \bigg)\;,\nonumber\\
    f_{13} & = \frac{1}{d_{\text{in}}(d_{\text{in}}-1)(d_{\text{in}}-2)} \bigg( d_{\text{in}}\hat{I}_3 - d_{\text{in}}\hat{I}_4 - \hat{I}_7 - \hat{I}_8 + 2\hat{I}_9 \bigg)\;,\label{eq:model_param_formulas_final}
\end{align}

as functions of the matrix size $d_{\text{in}}$ and average invariant value $I_i$, where the averaging is over the $N=1000$ runs.
The expectation values and standard deviations can hence be calculated by propagating the results for the invariants in §\ref{app:inv_analytics}.

Importantly, since the average invariants are used in these equations these will be related to the expectations of the invariants via
\begin{equation}
\begin{split}
    \langle \hat{I_{i}} \rangle & = \langle \frac{1}{N} \Sigma_{r=1}^N I_i^r \rangle\;,\\
    & =  \frac{1}{N} \Sigma_{r=1}^N \langle I_i^r \rangle\;,\\
    & = \langle I_i \rangle\;,
\end{split}
\end{equation}
since each of the $r$ runs\footnote{We emphasise here that the $r$ appearing in these equations is an index, and not a power.} are independent, using relations \eqref{eq:exp_rules}-\eqref{eq:exp_rules_final}. 
Whilst the variances will be related via
\begin{equation}
\begin{split}
    \text{Var}(\hat{I_{i}}) & = \text{Var}\bigg( \frac{1}{N} \Sigma_{r=1}^N I_i^r \bigg)\;,\\
    & = \frac{1}{N^2} \Sigma_{r=1}^N \text{Var}(  I_i^r )\;,\\
    & = \frac{1}{N} \text{Var}( I_i )\;,
\end{split}
\end{equation}
again since each of the $r$ runs are independent, and now using relations \eqref{eq:var_rules}-\eqref{eq:var_rules_final}.

The model parameter formulas in \eqref{eq:model_param_formulas}-\eqref{eq:model_param_formulas_final} are only non-linear in the first 2 invariants (since these are linear this matches these order 2 quadratic parameters).
Therefore to compute expectation values and variances of these model parameters, the invariant expectations and variances of §\ref{app:inv_analytics} are needed, as well as the expectations and variances for these non-linear terms.
These are computed here such they can be reused in each model parameter's computation.

\begin{equation}\label{eq:dep_inv_exps}
\begin{split}
    \langle \hat{I}_1^2 \rangle & = \bigg\langle \bigg( \frac{1}{N}\Sigma_{r_1,i=1,1}^{N,d_{\text{in}}} W_{ii}^{r_1} \bigg) \bigg( \frac{1}{N}\Sigma_{r_2,j=1,1}^{N,d_{\text{in}}} W_{jj}^{r_2} \bigg) \bigg\rangle \;,\\
    %& = \frac{1}{N^2} \bigg( \Sigma_{r_1,r_2,i,j=1,1,1,1|r_1=r_2 \& i=j}^{N,N,d_{\text{in}},d_{\text{in}}} (W_{ii}^{r_1})^2  + \Sigma_{r_1,r_2,i,j=1,1,1,1|\neg(r_1=r_2 \& i=j)}^{N,d_{\text{in}}} W_{ii}^{r_1}W_{jj}^{r_2} \bigg)\;,\\
    & = \frac{1}{N^2} (Nd_{\text{in}} \langle w^2 \rangle + (N^2d_{\text{in}}^2 - Nd_{\text{in}}) \langle w \rangle^2)\;,\\
    & = \frac{d_{\text{in}}}{N} \langle w^2 \rangle + d_{\text{in}}\bigg(d_{\text{in}} - \frac{1}{N}\bigg) \langle w \rangle^2\;,
\end{split}
\end{equation}
\begin{equation}
\begin{split}
    \langle \hat{I}_1\hat{I}_2 \rangle & = \bigg\langle \bigg( \frac{1}{N}\Sigma_{r_1,i=1,1}^{N,d_{\text{in}}} W_{ii}^{r_1} \bigg) \bigg( \frac{1}{N}\Sigma_{r_2,j,k=1,1,1}^{N,d_{\text{in}},d_{\text{in}}} W_{jk}^{r_2} \bigg) \bigg\rangle \;,\\
    & = \frac{1}{N^2} (Nd_{\text{in}} \langle w^2 \rangle + (N^2d_{\text{in}}^3 - Nd_{\text{in}}) \langle w \rangle^2)\;,\\
    & = \frac{d_{\text{in}}}{N} \langle w^2 \rangle + d_{\text{in}}\bigg(d_{\text{in}}^2 - \frac{1}{N}\bigg) \langle w \rangle^2\;,
\end{split}
\end{equation}
\begin{equation}\label{eq:dep_inv_exps_final}
\begin{split}
    \langle \hat{I}_2^2 \rangle & = \bigg\langle \bigg( \frac{1}{N}\Sigma_{r_1,i,j=1,1,1}^{N,d_{\text{in}},d_{\text{in}}} W_{ij}^{r_1} \bigg) \bigg( \frac{1}{N}\Sigma_{r_2,k,l=1,1,1}^{N,d_{\text{in}},d_{\text{in}}} W_{kl}^{r_2} \bigg) \bigg\rangle \;,\\
    & = \frac{1}{N^2} (Nd_{\text{in}}^2 \langle w^2 \rangle + (N^2d_{\text{in}}^4 - Nd_{\text{in}}^2) \langle w \rangle^2)\;,\\
    & = \frac{d_{\text{in}}^2}{N} \langle w^2 \rangle + d_{\text{in}}^2 \bigg(d_{\text{in}}^2 - \frac{1}{N}\bigg) \langle w \rangle^2\;,
\end{split}
\end{equation}

using techniques as used in §\ref{app:inv_analytics} for identifying what index equalities cause the weights to repeat and thus be dependent.
Whilst for the variances they are
\begin{equation}\label{eq:dep_inv_vars}
\begin{split}
    \text{Var}(\hat{I}_1^2) & = \text{Var}\bigg( \bigg(\frac{1}{N}\Sigma_{r_1,i=1,1}^{N,d_{\text{in}}} W_{ii}^{r_1} \bigg) \bigg( \frac{1}{N}\Sigma_{r_2,j=1,1}^{N,d_{\text{in}}} W_{jj}^{r_2}\bigg) \bigg) \;,\\
    & = \frac{1}{N^4} \Sigma_{r_1,r_2,i,j=1,1,1,1}^{N,N,d_{\text{in}},d_{\text{in}}} \text{Var}(W_{ii}^{r_1}W_{jj}^{r_2}) \;,\\
    & = \frac{1}{N^4} \bigg(Nd_{\text{in}} \text{Var}(w^2) + (N^2d_{\text{in}}^2 - Nd_{\text{in}})\text{Var}(w)^2\bigg)\;,\\
    & = \frac{1}{N^4} \bigg(Nd_{\text{in}} (\langle w^4 \rangle - \langle w^2 \rangle^2) + (N^2d_{\text{in}}^2 - Nd_{\text{in}}) \langle w^2 \rangle^2 \bigg)\;,\\
    & = \frac{d_{\text{in}}}{N^3} \langle w^4 \rangle + \frac{d_{\text{in}}}{N^2}\bigg( d_{\text{in}} - \frac{2}{N}\bigg) \langle w^2 \rangle^2 \;,
\end{split}
\end{equation}
\begin{equation}
\begin{split}
    \text{Var}(\hat{I}_1\hat{I}_2) & = \text{Var}\bigg( \bigg( \frac{1}{N}\Sigma_{r_1,i=1,1}^{N,d_{\text{in}}} W_{ii}^{r_1} \bigg) \bigg( \frac{1}{N}\Sigma_{r_2,j,k=1,1,1}^{N,d_{\text{in}},d_{\text{in}}} W_{jk}^{r_2} \bigg) \bigg) \;,\\
    & = \frac{1}{N^4} \Sigma_{r_1,r_2,i,j,k=1,1,1,1,1}^{N,N,d_{\text{in}},d_{\text{in}},d_{\text{in}}} \text{Var}(W_{ii}^{r_1}W_{jk}^{r_2}) \;,\\
    & = \frac{1}{N^4} \bigg(Nd_{\text{in}} \text{Var}(w^2) + (N^2d_{\text{in}}^3 - Nd_{\text{in}})\text{Var}(w)^2\bigg)\;,\\
    & = \frac{1}{N^4} \bigg(Nd_{\text{in}} (\langle w^4 \rangle - \langle w^2 \rangle^2) + (N^2d_{\text{in}}^3 - Nd_{\text{in}}) \langle w^2 \rangle^2 \bigg)\;,\\
    & = \frac{d_{\text{in}}}{N^3} \langle w^4 \rangle + \frac{d_{\text{in}}}{N^2}\bigg( d_{\text{in}}^2 - \frac{2}{N}\bigg) \langle w^2 \rangle^2 \;,
\end{split}
\end{equation}
\begin{equation}\label{eq:dep_inv_vars_final}
\begin{split}
    \text{Var}(\hat{I}_2^2) & = \text{Var}\bigg( \bigg( \frac{1}{N}\Sigma_{r_1,i,j=1,1,1}^{N,d_{\text{in}},d_{\text{in}}} W_{ij}^{r_1} \bigg) \bigg( \frac{1}{N}\Sigma_{r_2,k,l=1,1,1}^{N,d_{\text{in}},d_{\text{in}}} W_{kl}^{r_2} \bigg) \bigg) \;,\\
    & = \frac{1}{N^4} \Sigma_{r_1,r_2,i,j,k,l=1,1,1,1,1,1}^{N,N,d_{\text{in}},d_{\text{in}},d_{\text{in}},d_{\text{in}}} \text{Var}(W_{ij}^{r_1}W_{kl}^{r_2}) \;,\\
    & = \frac{1}{N^4} \bigg(Nd_{\text{in}}^2 \text{Var}(w^2) + (N^2d_{\text{in}}^4 - Nd_{\text{in}}^2)\text{Var}(w)^2\bigg)\;,\\
    & = \frac{1}{N^4} \bigg(Nd_{\text{in}}^2 (\langle w^4 \rangle - \langle w^2 \rangle^2) + (N^2d_{\text{in}}^4 - Nd_{\text{in}}^2) \langle w^2 \rangle^2 \bigg)\;,\\
    & = \frac{d_{\text{in}}^2}{N^3} \langle w^4 \rangle + \frac{d_{\text{in}}^2}{N^2}\bigg( d_{\text{in}}^2 - \frac{2}{N}\bigg) \langle w^2 \rangle^2 \;.
\end{split}
\end{equation}

\subsection{Expectation Values}
Equivalent to the invariant case in §\ref{app:inv_analytics}, there are 2 linear and 11 quadratic order model parameters, as introduced in §\ref{sec:gmms}.
Their expectation values for the NN weight matrices considered (for the predefined architecture with $d_{\text{in}}=10$) are computed here, as functions of the invariant expectation values.

Since the model parameters are sums of known invariants, despite the invariants not necessarily being independent (as they are functions of the same matrices), some of the same expectation rules can still be used, in particular the scaling \eqref{eq:exp_rules} and summation \eqref{eq:exp_rules_middle} rules, since these also apply to dependent variables.
The product rule \eqref{eq:exp_rules_final} does not generalise as straightforwardly to dependent variables, so these examples were computed directly in \eqref{eq:dep_inv_exps}-\eqref{eq:dep_inv_exps_final}.\\

\noindent\textit{Linear:}
\begin{equation}
\begin{split}
    \langle f_1 \rangle & = \frac{\langle \hat{I}_2 \rangle}{d_{\text{in}}} \;,\\
    & = \frac{d_{\text{in}}^2 \langle w \rangle}{d_{\text{in}}}\;,\\
    & = d_{\text{in}} \langle w \rangle\;,
\end{split}
\end{equation}
\begin{equation}
\begin{split}
    \langle f_2 \rangle & = \frac{1}{d_{\text{in}}\sqrt{d_{\text{in}}-1}} \bigg( d_{\text{in}} \langle \hat{I}_1 \rangle - \langle \hat{I}_2 \rangle \bigg)\;,\\
    & = \frac{1}{d_{\text{in}}\sqrt{d_{\text{in}}-1}} \bigg( d_{\text{in}} (d_{\text{in}} \langle w \rangle) - d_{\text{in}}^2 \langle w \rangle \bigg)\;,\\
    & = 0\;,
\end{split}
\end{equation}

\noindent\textit{Quadratic:}
\begin{equation}
\begin{split}
    \langle f_3 \rangle & = -\frac{1}{d_{\text{in}}^2} \bigg(\langle \hat{I}_2^2 \rangle - \langle \hat{I}_{10} \rangle \bigg)\;,\\
    & = -\frac{1}{d_{\text{in}}^2} \bigg(\bigg( \frac{d_{\text{in}}^2}{N} \langle w^2 \rangle + d_{\text{in}}^2 \bigg(d_{\text{in}}^2 - \frac{1}{N}\bigg) \langle w \rangle^2 \bigg) \\
    & \qquad\qquad - \bigg(d_{\text{in}}^2 \langle w^2 \rangle + d_{\text{in}}^2(d_{\text{in}}^2-2) \langle w \rangle^2\bigg) \bigg)\;,\\
    & = \bigg(1 - \frac{1}{N}\bigg) \langle w^2 \rangle - \bigg(2 - \frac{1}{N} \bigg) \langle w \rangle^2 \;,
\end{split}
\end{equation}
\begin{equation}
\begin{split}
    \langle f_4 \rangle & = -\frac{1}{d_{\text{in}}^2 \sqrt{d_{\text{in}}-1}} \bigg(d_{\text{in}} \langle \hat{I}_1\hat{I}_2 \rangle - \langle \hat{I}_2^2 \rangle + \langle \hat{I}_{10} \rangle - d_{\text{in}} \langle \hat{I}_{13} \rangle \bigg)\;,\\
    & = -\frac{1}{d_{\text{in}}^2 \sqrt{d_{\text{in}}-1}} \bigg( d_{\text{in}} \bigg( \frac{d_{\text{in}}}{N} \langle w^2 \rangle + d_{\text{in}}\bigg(d_{\text{in}}^2 - \frac{1}{N}\bigg) \langle w \rangle^2 \bigg) \\
    & \qquad\qquad - \bigg( \frac{d_{\text{in}}^2}{N} \langle w^2 \rangle + d_{\text{in}}^2 \bigg(d_{\text{in}}^2 - \frac{1}{N}\bigg) \langle w \rangle^2 \bigg) \\
    & \qquad\qquad + \bigg( d_{\text{in}}^2 \langle w^2 \rangle + (d_{\text{in}}^4-d_{\text{in}}^2) \langle w \rangle^2 \bigg) \\
    & \qquad\qquad - d_{\text{in}} \bigg( d_{\text{in}} \langle w^2 \rangle + (d_{\text{in}}^3-d_{\text{in}}) \langle w \rangle^2 \bigg) \bigg)\;,\\
    & = 0\;,
\end{split}
\end{equation}
\begin{equation}
\begin{split}
    \langle f_5 \rangle & = -\frac{1}{d_{\text{in}}^2 (d_{\text{in}}-1)} \bigg( d_{\text{in}}^2 \langle \hat{I}_1^2 \rangle - 2 d_{\text{in}} \langle \hat{I}_1\hat{I}_2 \rangle \\
    & \qquad\qquad + \langle \hat{I}_2^2 \rangle - \langle \hat{I}_{10} \rangle - d_{\text{in}}^2 \langle \hat{I}_{12} \rangle + 2 d_{\text{in}} \langle \hat{I}_{13} \rangle \bigg)\;,\\
    & = -\frac{1}{d_{\text{in}}^2 (d_{\text{in}}-1)} \bigg( d_{\text{in}}^2 \bigg( \frac{d_{\text{in}}}{N} \langle w^2 \rangle + d_{\text{in}}\bigg(d_{\text{in}} - \frac{1}{N}\bigg) \langle w \rangle^2 \bigg) \\
    & \qquad\qquad - 2 d_{\text{in}} \bigg( \frac{d_{\text{in}}}{N} \langle w^2 \rangle + d_{\text{in}}\bigg(d_{\text{in}}^2 - \frac{1}{N}\bigg) \langle w \rangle^2 \bigg) \\
    & \qquad\qquad + \bigg( \frac{d_{\text{in}}^2}{N} \langle w^2 \rangle + d_{\text{in}}^2 \bigg(d_{\text{in}}^2 - \frac{1}{N}\bigg) \langle w \rangle^2 \bigg) \\
    & \qquad\qquad - \bigg( d_{\text{in}}^2 \langle w^2 \rangle + (d_{\text{in}}^4-d_{\text{in}}^2) \langle w \rangle^2 \bigg) \\
    & \qquad\qquad - d_{\text{in}}^2 \bigg( d_{\text{in}} \langle w^2 \rangle + d_{\text{in}}(d_{\text{in}}-1) \langle w \rangle^2 \bigg) \\
    & \qquad\qquad + 2d_{\text{in}} \bigg( d_{\text{in}} \langle w^2 \rangle + (d_{\text{in}}^3-d_{\text{in}}) \langle w \rangle^2 \bigg)\bigg)\;,\\
    & = \bigg(1 - \frac{1}{N} \bigg) \langle w^2 \rangle - \bigg(1 - \frac{1}{N}\bigg) \langle w \rangle^2\;,
\end{split}
\end{equation}
\begin{equation}
\begin{split}
    \langle f_6 \rangle & = \frac{1}{d_{\text{in}}^2 (d_{\text{in}}-1)} \bigg( d_{\text{in}} \langle \hat{I}_8 \rangle - \langle \hat{I}_{10} \rangle \bigg)\\
    & = \frac{1}{d_{\text{in}}^2 (d_{\text{in}}-1)} \bigg( d_{\text{in}} \bigg( d_{\text{in}}^2 \langle w^2 \rangle + d_{\text{in}}^2(d_{\text{in}}-1) \langle w \rangle^2 \bigg)\\
    & \qquad\qquad - \bigg( d_{\text{in}}^2 \langle w^2 \rangle + (d_{\text{in}}^4-d_{\text{in}}^2) \langle w \rangle^2 \bigg)\bigg)\;,\\
    & = \langle w^2 \rangle - d_{\text{in}}\langle w \rangle^2\;,
\end{split}
\end{equation}
\begin{equation}
\begin{split}
    \langle f_7 \rangle & = \frac{1}{d_{\text{in}}^2 (d_{\text{in}}-1)} \bigg( d_{\text{in}} \langle \hat{I}_9 \rangle - \langle \hat{I}_{10} \rangle \bigg)\;,\\
    & = \frac{1}{d_{\text{in}}^2 (d_{\text{in}}-1)} \bigg(  d_{\text{in}} \bigg( d_{\text{in}} \langle w^2 \rangle + (d_{\text{in}}^3-d_{\text{in}}) \langle w \rangle^2 \bigg) \\
    & \qquad\qquad - \bigg( d_{\text{in}}^2 \langle w^2 \rangle + (d_{\text{in}}^4-d_{\text{in}}^2) \langle w \rangle^2 \bigg) \bigg)\;,\\
    & = 0\;,
\end{split}
\end{equation}
\begin{equation}
\begin{split}
    \langle f_8 \rangle & = \frac{1}{d_{\text{in}}^2(d_{\text{in}}-1)\sqrt{d_{\text{in}}-2}} \bigg( d_{\text{in}}^2 \langle \hat{I}_6 \rangle - d_{\text{in}} \langle \hat{I}_8 \rangle \\
    & \qquad\qquad - d_{\text{in}} \langle \hat{I}_9 \rangle + 2 \langle \hat{I}_{10} \rangle - d_{\text{in}} \langle \hat{I}_{13} \rangle \bigg)\;,\\
    & = \frac{1}{d_{\text{in}}^2(d_{\text{in}}-1)\sqrt{d_{\text{in}}-2}} \bigg( d_{\text{in}}^2 \bigg( d_{\text{in}} \langle w^2 \rangle + d_{\text{in}}(d_{\text{in}}-1) \langle w \rangle^2 \bigg)\\
    & \qquad\qquad - d_{\text{in}} \bigg( d_{\text{in}}^2 \langle w^2 \rangle + d_{\text{in}}^2(d_{\text{in}}-1) \langle w \rangle^2 \bigg)\\
    & \qquad\qquad - d_{\text{in}} \bigg( d_{\text{in}} \langle w^2 \rangle + (d_{\text{in}}^3-d_{\text{in}}) \langle w \rangle^2 \bigg)\\
    & \qquad\qquad + 2\bigg( d_{\text{in}}^2 \langle w^2 \rangle + (d_{\text{in}}^4-d_{\text{in}}^2) \langle w \rangle^2 \bigg)\\
    & \qquad\qquad - d_{\text{in}}\bigg( d_{\text{in}} \langle w^2 \rangle + (d_{\text{in}}^3-d_{\text{in}}) \langle w \rangle^2 \bigg)\bigg)\;,\\
    & = 0\;,
\end{split}
\end{equation}
\begin{equation}
\begin{split}
    \langle f_9 \rangle & = \frac{1}{d_{\text{in}}^2 (d_{\text{in}}-1)} \bigg( d_{\text{in}} \langle \hat{I}_7 \rangle - \langle \hat{I}_{10} \rangle \bigg)\;,\\
    & = \frac{1}{d_{\text{in}}^2 (d_{\text{in}}-1)} \bigg( d_{\text{in}} \bigg( d_{\text{in}}^2 \langle w^2 \rangle + d_{\text{in}}^2(d_{\text{in}}-1) \langle w \rangle^2 \bigg)\\
    & \qquad\qquad - \bigg( d_{\text{in}}^2 \langle w^2 \rangle + (d_{\text{in}}^4-d_{\text{in}}^2) \langle w \rangle^2 \bigg)\bigg)\;,\\
    & = \langle w^2 \rangle - d_{\text{in}} \langle w \rangle^2\;, 
\end{split}
\end{equation}
\begin{equation}
\begin{split}
    \langle f_{10} \rangle & = \frac{1}{d_{\text{in}}^2(d_{\text{in}}-1)\sqrt{d_{\text{in}}-2}} \bigg( d_{\text{in}}^2 \langle \hat{I}_5 \rangle - d_{\text{in}} \langle \hat{I}_7 \rangle \\
    & \qquad\qquad - d_{\text{in}} \langle \hat{I}_9 \rangle + 2 \langle \hat{I}_{10} \rangle - d_{\text{in}} \langle \hat{I}_{13} \rangle \bigg)\;,\\ 
    & = \frac{1}{d_{\text{in}}^2(d_{\text{in}}-1)\sqrt{d_{\text{in}}-2}} \bigg( d_{\text{in}}^2 \bigg( d_{\text{in}} \langle w^2 \rangle + d_{\text{in}}(d_{\text{in}}-1) \langle w \rangle^2 \bigg)\\
    & \qquad\qquad - d_{\text{in}} \bigg( d_{\text{in}}^2 \langle w^2 \rangle + d_{\text{in}}^2(d_{\text{in}}-1) \langle w \rangle^2 \bigg)\\
    & \qquad\qquad - d_{\text{in}} \bigg( d_{\text{in}} \langle w^2 \rangle + (d_{\text{in}}^3-d_{\text{in}}) \langle w \rangle^2 \bigg)\\
    & \qquad\qquad + 2\bigg( d_{\text{in}}^2 \langle w^2 \rangle + (d_{\text{in}}^4-d_{\text{in}}^2) \langle w \rangle^2 \bigg)\\
    & \qquad\qquad - d_{\text{in}} \bigg( d_{\text{in}} \langle w^2 \rangle + (d_{\text{in}}^3-d_{\text{in}}) \langle w \rangle^2 \bigg)\;,\\
    & = 0\;,
\end{split}
\end{equation}
\begin{equation}
\begin{split}
    \langle f_{11} \rangle & = -\frac{1}{d_{\text{in}}^2(d_{\text{in}}-1)(d_{\text{in}}-2)} \bigg( 
    2d_{\text{in}}^2 \langle\hat{I}_5\rangle + 2d_{\text{in}}^2 \langle\hat{I}_6\rangle - d_{\text{in}} \langle\hat{I}_7\rangle - d_{\text{in}} \langle\hat{I}_8\rangle \\
    & \qquad - 2d_{\text{in}} \langle\hat{I}_9\rangle + 4 \langle\hat{I}_{10}\rangle - d_{\text{in}}^3 \langle\hat{I}_{11}\rangle + d_{\text{in}}^2 \langle\hat{I}_{12}\rangle - 4d_{\text{in}} \langle\hat{I}_{13}\rangle \bigg)\;,\\
    & = -\frac{1}{d_{\text{in}}^2(d_{\text{in}}-1)(d_{\text{in}}-2)} \bigg( 2d_{\text{in}}^2 \bigg( d_{\text{in}} \langle w^2 \rangle + d_{\text{in}}(d_{\text{in}}-1) \langle w \rangle^2 \bigg)\\
    & \qquad\qquad + 2d_{\text{in}}^2 \bigg( d_{\text{in}} \langle w^2 \rangle + d_{\text{in}}(d_{\text{in}}-1) \langle w \rangle^2 \bigg)\\
    & \qquad\qquad - d_{\text{in}} \bigg(  d_{\text{in}}^2 \langle w^2 \rangle + d_{\text{in}}^2(d_{\text{in}}-1) \langle w \rangle^2 \bigg)\\
    & \qquad\qquad - d_{\text{in}} \bigg( d_{\text{in}}^2 \langle w^2 \rangle + d_{\text{in}}^2(d_{\text{in}}-1) \langle w \rangle^2 \bigg)\\
    & \qquad\qquad - 2d_{\text{in}} \bigg( d_{\text{in}} \langle w^2 \rangle + (d_{\text{in}}^3-d_{\text{in}}) \langle w \rangle^2 \bigg)\\
    & \qquad\qquad + 4 \bigg( d_{\text{in}}^2 \langle w^2 \rangle + (d_{\text{in}}^4-d_{\text{in}}^2) \langle w \rangle^2 \bigg)\\
    & \qquad\qquad - d_{\text{in}}^3 \bigg( d_{\text{in}} \langle w^2 \rangle \bigg)\\
    & \qquad\qquad + d_{\text{in}}^2 \bigg( d_{\text{in}} \langle w^2 \rangle + d_{\text{in}}(d_{\text{in}}-1) \langle w \rangle^2 \bigg)\\
    & \qquad\qquad - 4d_{\text{in}} \bigg( d_{\text{in}} \langle w^2 \rangle + (d_{\text{in}}^3-d_{\text{in}}) \langle w \rangle^2 \bigg)\;,\\
    & = \langle w^2 \rangle - \langle w \rangle^2\;,
\end{split}
\end{equation}
\begin{equation}
\begin{split}
    \langle f_{12} \rangle & = \frac{1}{d_{\text{in}}(d_{\text{in}}-1)(d_{\text{in}}-2)(d_{\text{in}}-3)} \bigg( (d_{\text{in}}-1)(d_{\text{in}}-2) \langle\hat{I}_3\rangle \\
    & \qquad + (d_{\text{in}}-1)(d_{\text{in}}-2) \langle\hat{I}_4\rangle + 4(d_{\text{in}}-1) \langle\hat{I}_5\rangle + 4(d_{\text{in}}-1) \langle\hat{I}_6\rangle \\
    & \qquad - (d_{\text{in}}-1) \langle\hat{I}_7\rangle - (d_{\text{in}}-1) \langle\hat{I}_8\rangle - 2(d_{\text{in}}-1) \langle\hat{I}_9\rangle + 2 \langle\hat{I}_{10}\rangle \\
    & \qquad - 2d_{\text{in}}(d_{\text{in}}-1) \langle\hat{I}_{11}\rangle + 2 \langle\hat{I}_{12}\rangle - 4 \langle\hat{I}_{13}\rangle \bigg)\;,\\
    & = \frac{1}{d_{\text{in}}(d_{\text{in}}-1)(d_{\text{in}}-2)(d_{\text{in}}-3)} \bigg( (d_{\text{in}}-1)(d_{\text{in}}-2) \bigg( d_{\text{in}}^2 \langle w^2 \rangle \bigg)\\
    & \qquad\qquad + (d_{\text{in}}-1)(d_{\text{in}}-2) \bigg( d_{\text{in}} \langle w^2 \rangle + d_{\text{in}}(d_{\text{in}}-1) \langle w \rangle^2 \bigg)\\
    & \qquad\qquad + 4(d_{\text{in}}-1) \bigg( d_{\text{in}} \langle w^2 \rangle + d_{\text{in}}(d_{\text{in}}-1) \langle w \rangle^2 \bigg)\\
    & \qquad\qquad + 4(d_{\text{in}}-1) \bigg( d_{\text{in}} \langle w^2 \rangle + d_{\text{in}}(d_{\text{in}}-1) \langle w \rangle^2 \bigg)\\
    & \qquad\qquad - (d_{\text{in}}-1) \bigg( d_{\text{in}}^2 \langle w^2 \rangle + d_{\text{in}}^2(d_{\text{in}}-1) \langle w \rangle^2 \bigg)\\
    & \qquad\qquad - (d_{\text{in}}-1) \bigg( d_{\text{in}}^2 \langle w^2 \rangle + d_{\text{in}}^2(d_{\text{in}}-1) \langle w \rangle^2 \bigg)\\
    & \qquad\qquad - 2(d_{\text{in}}-1) \bigg( d_{\text{in}} \langle w^2 \rangle + (d_{\text{in}}^3-d_{\text{in}}) \langle w \rangle^2 \bigg)\\
    & \qquad\qquad + 2 \bigg( d_{\text{in}}^2 \langle w^2 \rangle + (d_{\text{in}}^4-d_{\text{in}}^2) \langle w \rangle^2 \bigg)\\
    & \qquad\qquad - 2d_{\text{in}}(d_{\text{in}}-1) \bigg( d_{\text{in}} \langle w^2 \rangle \bigg)\\
    & \qquad\qquad + 2 \bigg( d_{\text{in}} \langle w^2 \rangle + d_{\text{in}}(d_{\text{in}}-1) \langle w \rangle^2 \bigg)\\
    & \qquad\qquad - 4 \bigg( d_{\text{in}} \langle w^2 \rangle + (d_{\text{in}}^3-d_{\text{in}}) \langle w \rangle^2 \bigg)\bigg)\;,\\
    & = \langle w^2 \rangle - \langle w \rangle^2 \;,
\end{split}
\end{equation}
\begin{equation}
\begin{split}
    \langle f_{13} \rangle & = \frac{1}{d_{\text{in}}(d_{\text{in}}-1)(d_{\text{in}}-2)} \bigg( d_{\text{in}} \langle\hat{I}_3\rangle - d_{\text{in}} \langle\hat{I}_4\rangle - \langle\hat{I}_7\rangle - \langle\hat{I}_8\rangle + 2\langle\hat{I}_9\rangle \bigg)\;,\\
    & = \frac{1}{d_{\text{in}}(d_{\text{in}}-1)(d_{\text{in}}-2)} \bigg( d_{\text{in}} \bigg( d_{\text{in}}^2 \langle w^2 \rangle \bigg)\\
    & \qquad\qquad - d_{\text{in}} \bigg( d_{\text{in}} \langle w^2 \rangle + d_{\text{in}}(d_{\text{in}}-1) \langle w \rangle^2 \bigg)\\
    & \qquad\qquad - \bigg( d_{\text{in}}^2 \langle w^2 \rangle + d_{\text{in}}^2(d_{\text{in}}-1) \langle w \rangle^2 \bigg)\\
    & \qquad\qquad - \bigg( d_{\text{in}}^2 \langle w^2 \rangle + d_{\text{in}}^2(d_{\text{in}}-1) \langle w \rangle^2 \bigg)\\
    & \qquad\qquad + 2 \bigg( d_{\text{in}} \langle w^2 \rangle + (d_{\text{in}}^3-d_{\text{in}}) \langle w \rangle^2 \bigg)\;,\\
    & = \langle w^2 \rangle - \langle w \rangle^2\;.
\end{split}
\end{equation}

We note that the model parameter expectation values are identically zero for $(f_2, f_4, f_7, f_8, f_{10})$, and are equal for $\langle f_6 \rangle = \langle f_9 \rangle$, $\langle f_{11} \rangle = \langle f_{12} \rangle = \langle f_{13} \rangle$.

\newpage 
For the investigations carried out in this work, $d_{\text{in}}=10$, $N=1000$, and $\langle w^k \rangle$ are given as in \eqref{eq:gauss_w_exp_start}-\eqref{eq:gauss_w_exp_final} and \eqref{eq:uniform_w_exp_start}-\eqref{eq:uniform_w_exp_final} for the Gaussian and Uniform initialisation schemes respectively.
With these values substituted, the final expectation values are given in Table \ref{tab:param_analytic_subbed}, noting there are now more accidental equalities.

\begin{table}[!t]
\centering
\addtolength{\leftskip}{-2cm}
\addtolength{\rightskip}{-2cm}
\begin{tabular}{|cc|cc|ccccccccccc|}
\hline
\multicolumn{2}{|c|}{\multirow{2}{*}{Model Parameter}}                                                 & \multicolumn{2}{c|}{Linear}        & \multicolumn{11}{c|}{Quadratic} \\ \cline{3-15} 
\multicolumn{2}{|c|}{}                                                                           & \multicolumn{1}{c|}{$f_1$} & $f_2$ & \multicolumn{1}{c|}{$f_3$} & \multicolumn{1}{c|}{$f_4$} & \multicolumn{1}{c|}{$f_5$} & \multicolumn{1}{c|}{$f_6$} & \multicolumn{1}{c|}{$f_7$} & \multicolumn{1}{c|}{$f_8$} & \multicolumn{1}{c|}{$f_9$} & \multicolumn{1}{c|}{$f_{10}$} & \multicolumn{1}{c|}{$f_{11}$} & \multicolumn{1}{c|}{$f_{12}$} & $f_{13}$ \\ \hline
\multicolumn{1}{|c|}{\multirow{2}{*}{\begin{tabular}[c]{@{}c@{}}Init\\ Scheme\end{tabular}}} & G & \multicolumn{1}{c|}{0}     & 0     & \multicolumn{1}{c|}{$\frac{999}{10000}$}      & \multicolumn{1}{c|}{0}      & \multicolumn{1}{c|}{$\frac{999}{10000}$}      & \multicolumn{1}{c|}{$\frac{1}{10}$}      & \multicolumn{1}{c|}{0}      & \multicolumn{1}{c|}{0}      & \multicolumn{1}{c|}{$\frac{1}{10}$}      & \multicolumn{1}{c|}{0}         & \multicolumn{1}{c|}{$\frac{1}{10}$}         & \multicolumn{1}{c|}{$\frac{1}{10}$}         &     $\frac{1}{10}$     \\ \cline{2-15} 
\multicolumn{1}{|c|}{}                                                                       & U & \multicolumn{1}{c|}{0}     & 0     & \multicolumn{1}{c|}{$\frac{333}{10000}$}      & \multicolumn{1}{c|}{0}      & \multicolumn{1}{c|}{$\frac{333}{10000}$}      & \multicolumn{1}{c|}{$\frac{1}{30}$}      & \multicolumn{1}{c|}{0}      & \multicolumn{1}{c|}{0}      & \multicolumn{1}{c|}{$\frac{1}{30}$}      & \multicolumn{1}{c|}{0}         & \multicolumn{1}{c|}{$\frac{1}{30}$}         & \multicolumn{1}{c|}{$\frac{1}{30}$}         &   $\frac{1}{30}$       \\ \hline
\end{tabular}
\caption{Expected Values for the linear and quadratic model parameters for all layers, shown for the 2 initialisation schemes considered: Gaussian (G) and Uniform (U) respectively. Specifically these values have the explicit $d_{\text{in}}$ and $\langle w^k \rangle$ values substituted.}
\label{tab:param_analytic_subbed}
\end{table}

\subsection{Standard Deviations}
To now compute the standard deviations of these 2 linear and 11 quadratic model parameters, an equivalent calculation procedure is followed as for the invariants, propagating the variances uses the rules outlined in \eqref{eq:var_rules}-\eqref{eq:var_rules_final}.
Since only a single model parameter is calculated from average invariants, there is no set of model parameters to take a mean over, and thus we use the standard deviation as the error measure (we do not calculate a standard error, which would anyway equal the standard deviation over this sample of 1 model parameter).

Unlike for expectation values, the relations in \eqref{eq:var_rules}-\eqref{eq:var_rules_final} do not directly apply to dependent variables.
The constant scaling \eqref{eq:var_rules} does apply, the summation \eqref{eq:var_rules_middle} generalises as:
\begin{equation}
    \text{Var}(\Sigma_i X_i) = \Sigma_i \text{Var}(X_i) + 2 \Sigma_{i<j} \text{Covar}(X_i, X_j)\;,
\end{equation}
in terms of the covariance, $\text{Covar}(\cdot, \cdot)$, whereas the product \eqref{eq:var_rules_final} generalisation does not have a simple closed form.
The variances of products of invariants in the model parameter expressions where hence computed directly in \eqref{eq:dep_inv_vars}-\eqref{eq:dep_inv_vars_final}.
However the covariances between all the invariants are particularly expensive to compute.
Therefore to make computation of a standard deviation feasible we make the \textit{assumption} that the invariants are negligibly dependent, such that the covariances are all 0, and hence continue to use the expression \eqref{eq:var_rules_middle} for summation of invariants. 
Additionally, some analytic computation of the covariances between invariants with simpler expressions further motivated this assumption experimentally. \\

\noindent\textit{Linear:}
\begin{equation}
\begin{split}
    \text{Var}(f_1) & = \text{Var}\bigg( \frac{\hat{I}_2}{d_{\text{in}}} \bigg)\;,\\
    & = \frac{1}{d_{\text{in}}^2}\text{Var}(\hat{I}_2)\;,\\
    & = \frac{1}{Nd_{\text{in}}^2}\text{Var}(I_2)\;,\\
    & = \frac{1}{Nd_{\text{in}}^2}d_{\text{in}}^2 \langle w^2 \rangle\;,\\
    & = \frac{1}{N} \langle w^2 \rangle\;,\\
    \implies \hat{\sigma}(f_1) & = \sqrt{\frac{\langle w^2 \rangle}{N}}\;,
\end{split}
\end{equation}
\begin{equation}
\begin{split}
    \text{Var}(f_2) & = \text{Var}\bigg( \frac{1}{d_{\text{in}}\sqrt{d_{\text{in}}-1}} \bigg( d_{\text{in}}\hat{I}_1 - \hat{I}_2 \bigg) \bigg)\;,\\
    & = \frac{1}{Nd_{\text{in}}^2(d_{\text{in}}-1)} \bigg(d_{\text{in}}^2\text{Var}(I_1) + \text{Var}(I_2)\bigg)\;,\\
    & = \frac{1}{Nd_{\text{in}}^2(d_{\text{in}}-1)} \bigg(d_{\text{in}}^2\bigg( d_{\text{in}} \langle w^2 \rangle \bigg) + d_{\text{in}}^2 \langle w^2 \rangle \bigg)\;,\\
    & = \frac{d_{\text{in}}+1}{N(d_{\text{in}}-1)}\langle w^2 \rangle\;,\\
    \implies \hat{\sigma}(f_2) & = \sqrt{\frac{(d_{\text{in}}+1)}{N(d_{\text{in}}-1)} \langle w^2 \rangle}\;,
\end{split}
\end{equation}

\noindent\textit{Quadratic:}
\begin{equation}
\begin{split}
    \text{Var}(f_3) & = \text{Var}\bigg( -\frac{1}{d_{\text{in}}^2} \bigg(\hat{I}_2^2 - \hat{I}_{10}\bigg) \bigg)\;,\\
    & = \frac{1}{d_{\text{in}}^4} \bigg( \text{Var}(\hat{I}_2^2) + \frac{1}{N} \text{Var}(I_{10}) \bigg)\;,\\
    & = \frac{1}{d_{\text{in}}^4} \bigg( \bigg( \frac{d_{\text{in}}^2}{N^3} \langle w^4 \rangle + \frac{d_{\text{in}}^2}{N^2}\bigg( d_{\text{in}}^2 - \frac{2}{N}\bigg) \langle w^2 \rangle^2 \bigg) \\
    & \qquad\qquad + \frac{1}{N}\bigg( d_{\text{in}}^2 \langle w^4 \rangle + d_{\text{in}}^2(d_{\text{in}}^2-2) \langle w^2 \rangle^2 \bigg)\bigg)\;,\\
    & = \frac{1}{d_{\text{in}}^2}\bigg(\frac{1}{N} + \frac{1}{N^3}\bigg) \langle w^4 \rangle \\
    & \qquad\qquad + \bigg( \frac{1}{N} + \frac{1}{N^2} - \frac{2}{Nd_{\text{in}}^2} - \frac{2}{N^3d_{\text{in}}^2}  \bigg) \langle w^2 \rangle^2\;,\\
    \implies \hat{\sigma}(f_3) & = \sqrt{\frac{1}{d_{\text{in}}^2}\bigg(\frac{1}{N} + \frac{1}{N^3}\bigg) \langle w^4 \rangle} \\
    & \qquad\qquad \overline{ + \bigg( \frac{1}{N} + \frac{1}{N^2} - \frac{2}{Nd_{\text{in}}^2} - \frac{2}{N^3d_{\text{in}}^2}  \bigg) \langle w^2 \rangle^2} \;,
\end{split}
\end{equation}
\begin{equation}
\begin{split}
    \text{Var}(f_4) & = \text{Var}\bigg( -\frac{1}{d_{\text{in}}^2 \sqrt{d_{\text{in}}-1}} \bigg(d_{\text{in}}\hat{I}_1\hat{I}_2 - \hat{I}_2^2 + \hat{I}_{10} - d_{\text{in}}\hat{I}_{13} \bigg) \bigg)\;,\\
    & = \frac{1}{d_{\text{in}}^4 (d_{\text{in}}-1)} \bigg( d_{\text{in}}^2 \text{Var}(\hat{I}_1\hat{I}_2) + \text{Var}(\hat{I}_2^2) \\
    & \qquad\qquad + \frac{1}{N} \text{Var}(I_{10}) + \frac{d_{\text{in}}^2}{N}\text{Var}(I_{13}) \bigg) \bigg)\;,\\
    & = \frac{1}{d_{\text{in}}^4 (d_{\text{in}}-1)} \bigg( d_{\text{in}}^2\bigg( \frac{d_{\text{in}}}{N^3} \langle w^4 \rangle + \frac{d_{\text{in}}}{N^2}\bigg( d_{\text{in}}^2 - \frac{2}{N}\bigg) \langle w^2 \rangle^2 \bigg) \\
    & \qquad\qquad + \bigg( \frac{d_{\text{in}}^2}{N^3} \langle w^4 \rangle + \frac{d_{\text{in}}^2}{N^2}\bigg( d_{\text{in}}^2 - \frac{2}{N}\bigg) \langle w^2 \rangle^2 \bigg) \\
    & \qquad\qquad + \frac{1}{N} \bigg( d_{\text{in}}^2 \langle w^4 \rangle + d_{\text{in}}^2(d_{\text{in}}^2-2) \langle w^2 \rangle^2 \bigg) \\
    & \qquad\qquad + \frac{d_{\text{in}}^2}{N} \bigg( d_{\text{in}} \langle w^4 \rangle + d_{\text{in}}(d_{\text{in}}^2-2) \langle w^2 \rangle^2 \bigg)\bigg)\;,\\
    & = \frac{d_{\text{in}}+1}{d_{\text{in}}^2 (d_{\text{in}}-1)}\bigg(\frac{1}{N} + \frac{1}{N^3}\bigg) \langle w^4 \rangle \\
    & \qquad\qquad + \frac{d_{\text{in}}+1}{d_{\text{in}}^2 (d_{\text{in}}-1)}\bigg( \frac{d_{\text{in}}^2-2}{N} + \frac{d_{\text{in}}^2}{N^2} - \frac{2}{N^3} \bigg) \langle w^2 \rangle^2 \;,\\ 
    \implies \hat{\sigma}(f_4) & = \sqrt{\frac{d_{\text{in}}+1}{d_{\text{in}}^2 (d_{\text{in}}-1)}\bigg(\frac{1}{N} + \frac{1}{N^3}\bigg) \langle w^4 \rangle} \\
    & \qquad\qquad \overline{ + \frac{d_{\text{in}}+1}{d_{\text{in}}^2 (d_{\text{in}}-1)}\bigg( \frac{d_{\text{in}}^2-2}{N} + \frac{d_{\text{in}}^2}{N^2} - \frac{2}{N^3} \bigg) \langle w^2 \rangle^2 }\;,
\end{split}
\end{equation}
\begin{equation}
\begin{split}
    \text{Var}(f_5) & = \text{Var}\bigg( -\frac{1}{d_{\text{in}}^2 (d_{\text{in}}-1)} \bigg( d_{\text{in}}^2 \hat{I}_1^2 - 2 d_{\text{in}} \hat{I}_1\hat{I}_2 \\
    & \qquad\qquad + \hat{I}_2^2 - \hat{I}_{10} - d_{\text{in}}^2 \hat{I}_{12} + 2 d_{\text{in}} \hat{I}_{13} \bigg) \bigg)\;,\\
    & = \frac{1}{d_{\text{in}}^4 (d_{\text{in}}-1)^2} \bigg( d_{\text{in}}^4 \text{Var}(\hat{I}_1^2) + 4d_{\text{in}}^2 \text{Var}(\hat{I}_1\hat{I}_2) \\
    & \qquad\qquad + \text{Var}(\hat{I}_2^2) + \frac{1}{N}\text{Var}(I_{10}) \\
    & \qquad\qquad + \frac{d_{\text{in}}^4}{N}\text{Var}(I_{12}) + \frac{4d_{\text{in}}^2}{N}\text{Var}(I_{13}) \bigg)\;,\\
    & = \frac{1}{d_{\text{in}}^4 (d_{\text{in}}-1)^2} \bigg( d_{\text{in}}^4 \bigg( \frac{d_{\text{in}}}{N^3} \langle w^4 \rangle + \frac{d_{\text{in}}}{N^2}\bigg( d_{\text{in}} - \frac{2}{N}\bigg) \langle w^2 \rangle^2 \bigg)\\
    & \qquad\qquad + 4d_{\text{in}}^2 \bigg( \frac{d_{\text{in}}}{N^3} \langle w^4 \rangle + \frac{d_{\text{in}}}{N^2}\bigg( d_{\text{in}}^2 - \frac{2}{N}\bigg) \langle w^2 \rangle^2  \bigg)\\
    & \qquad\qquad + \bigg( \frac{d_{\text{in}}^2}{N^3} \langle w^4 \rangle + \frac{d_{\text{in}}^2}{N^2}\bigg( d_{\text{in}}^2 - \frac{2}{N}\bigg) \langle w^2 \rangle^2 \bigg)\\
    & \qquad\qquad + \frac{1}{N} \bigg( d_{\text{in}}^2 \langle w^4 \rangle + d_{\text{in}}^2(d_{\text{in}}^2-2) \langle w^2 \rangle^2 \bigg)\\
    & \qquad\qquad + \frac{d_{\text{in}}^4}{N} \bigg( d_{\text{in}} \langle w^4 \rangle + d_{\text{in}}(d_{\text{in}}-2) \langle w^2 \rangle^2 \bigg)\\
    & \qquad\qquad + \frac{4d_{\text{in}}^2}{N} \bigg( d_{\text{in}} \langle w^4 \rangle + d_{\text{in}}(d_{\text{in}}^2-2) \langle w^2 \rangle^2 \bigg) \bigg)\;,\\
    & = \frac{d_{\text{in}}^3+4d_{\text{in}}+1}{d_{\text{in}}^2 (d_{\text{in}}-1)^2} \bigg( \frac{1}{N} + \frac{1}{N^3} \bigg) \langle w^4 \rangle\\
    & \qquad\qquad + \frac{1}{d_{\text{in}}^2 (d_{\text{in}}-1)^2} \bigg( \frac{d_{\text{in}}^4 + 4d_{\text{in}}^3 -d_{\text{in}}^2 -8d_{\text{in}} - 2}{N} \\
    & \qquad\qquad + \frac{d_{\text{in}}^4 + 4d_{\text{in}}^3 + d_{\text{in}}^2}{N^2} - \frac{2d_{\text{in}}^3 + 8d_{\text{in}} + 2}{N^3} \bigg)\langle w^2 \rangle^2 \;,\\
    \implies \hat{\sigma}(f_5) & = \sqrt{\frac{d_{\text{in}}^3+4d_{\text{in}}+1}{d_{\text{in}}^2 (d_{\text{in}}-1)^2} \bigg( \frac{1}{N} + \frac{1}{N^3} \bigg) \langle w^4 \rangle}\\
    & \qquad\qquad \overline{+ \frac{1}{d_{\text{in}}^2 (d_{\text{in}}-1)^2} \bigg( \frac{d_{\text{in}}^4 + 4d_{\text{in}}^3 -d_{\text{in}}^2 -8d_{\text{in}} - 2}{N}} \\
    & \qquad\qquad \overline{+ \frac{d_{\text{in}}^4 + 4d_{\text{in}}^3 + d_{\text{in}}^2}{N^2} - \frac{2d_{\text{in}}^3 + 8d_{\text{in}} + 2}{N^3} \bigg)\langle w^2 \rangle^2}\;,
\end{split}
\end{equation}
\begin{equation}
\begin{split}
    \text{Var}(f_6) & = \text{Var}\bigg( \frac{1}{d_{\text{in}}^2 (d_{\text{in}}-1)} \bigg( d_{\text{in}}\hat{I}_8 - \hat{I}_{10} \bigg) \bigg)\;,\\
    & = \frac{1}{d_{\text{in}}^4 (d_{\text{in}}-1)^2} \bigg( \frac{d_{\text{in}}^2}{N} \text{Var}(I_8) + \frac{1}{N}\text{Var}(I_{10}) \bigg) \;,\\
    & = \frac{1}{d_{\text{in}}^4 (d_{\text{in}}-1)^2} \bigg( \frac{d_{\text{in}}^2}{N} \bigg( d_{\text{in}}^2 \langle w^4 \rangle + d_{\text{in}}^2(d_{\text{in}}-2) \langle w^2 \rangle^2 \bigg)\\
    & \qquad\qquad + \frac{1}{N} \bigg( d_{\text{in}}^2 \langle w^4 \rangle + d_{\text{in}}^2(d_{\text{in}}^2-2) \langle w^2 \rangle^2 \bigg)\bigg)\;,\\
    & = \frac{d_{\text{in}}^2+1}{Nd_{\text{in}}^2 (d_{\text{in}}-1)^2} \langle w^4 \rangle + \frac{d_{\text{in}}^3-d_{\text{in}}^2-2}{Nd_{\text{in}}^2(d_{\text{in}}-1)^2} \langle w^2 \rangle^2 \;,\\
     \implies \hat{\sigma}(f_6) & = \sqrt{\frac{d_{\text{in}}^2+1}{Nd_{\text{in}}^2 (d_{\text{in}}-1)^2} \langle w^4 \rangle + \frac{d_{\text{in}}^3-d_{\text{in}}^2-2}{Nd_{\text{in}}^2(d_{\text{in}}-1)^2} \langle w^2 \rangle^2}\;,
\end{split}
\end{equation}
\begin{equation}
\begin{split}
    \text{Var}(f_7) & = \text{Var}\bigg( \frac{1}{d_{\text{in}}^2 (d_{\text{in}}-1)} \bigg( d_{\text{in}}\hat{I}_9 - \hat{I}_{10} \bigg) \bigg)\;,\\
    & = \frac{1}{d_{\text{in}}^4 (d_{\text{in}}-1)^2} \bigg( \frac{d_{\text{in}}^2}{N} \text{Var}(I_9) + \frac{1}{N}\text{Var}(I_{10}) \bigg) \;,\\
    & = \frac{1}{d_{\text{in}}^4 (d_{\text{in}}-1)^2} \bigg( \frac{d_{\text{in}}^2}{N} \bigg( d_{\text{in}} \langle w^4 \rangle + d_{\text{in}}(d_{\text{in}}^2-2) \langle w^2 \rangle^2 \bigg)\\
    & \qquad\qquad + \frac{1}{N} \bigg( d_{\text{in}}^2 \langle w^4 \rangle + d_{\text{in}}^2(d_{\text{in}}^2-2) \langle w^2 \rangle^2 \bigg)\bigg)\;,\\
    & = \frac{d_{\text{in}}+1}{Nd_{\text{in}}^2 (d_{\text{in}}-1)^2} \langle w^4 \rangle + \frac{(d_{\text{in}}+1)(d_{\text{in}}^2-2)}{Nd_{\text{in}}^2(d_{\text{in}}-1)^2} \langle w^2 \rangle^2 \;,\\
    \implies \hat{\sigma}(f_7) & = \sqrt{\frac{d_{\text{in}}+1}{Nd_{\text{in}}^2 (d_{\text{in}}-1)^2} \langle w^4 \rangle + \frac{(d_{\text{in}}+1)(d_{\text{in}}^2-2)}{Nd_{\text{in}}^2(d_{\text{in}}-1)^2} \langle w^2 \rangle^2}\;,
\end{split}
\end{equation}
\begin{equation}
\begin{split}
    \text{Var}(f_8) & = \text{Var}\bigg( \frac{1}{d_{\text{in}}^2(d_{\text{in}}-1)\sqrt{d_{\text{in}}-2}} \bigg( d_{\text{in}}^2\hat{I}_6 - d_{\text{in}}\hat{I}_8 \\
    & \qquad\qquad - d_{\text{in}}\hat{I}_9 + 2\hat{I}_{10} - d_{\text{in}}\hat{I}_{13} \bigg) \bigg)\;,\\
    & =  \frac{1}{d_{\text{in}}^4(d_{\text{in}}-1)^2(d_{\text{in}}-2)} \bigg( \frac{d_{\text{in}}^4}{N} \text{Var}(I_6) + \frac{d_{\text{in}}^2}{N} \text{Var}(I_8) \\
    & \qquad\qquad + \frac{d_{\text{in}}^2}{N} \text{Var}(I_9) + \frac{4}{N} \text{Var}(I_{10}) + \frac{d_{\text{in}}^2}{N} \text{Var}(I_{13}) \bigg)\;,\\
    & =  \frac{1}{d_{\text{in}}^4(d_{\text{in}}-1)^2(d_{\text{in}}-2)} \\
    & \qquad\qquad \bigg( \frac{d_{\text{in}}^4}{N} \bigg( d_{\text{in}} \langle w^4 \rangle + d_{\text{in}}(d_{\text{in}}-2) \langle w^2 \rangle^2 \bigg)\\
    & \qquad\qquad + \frac{d_{\text{in}}^2}{N} \bigg( d_{\text{in}}^2 \langle w^4 \rangle + d_{\text{in}}^2(d_{\text{in}}-2) \langle w^2 \rangle^2 \bigg)\\
    & \qquad\qquad + \frac{d_{\text{in}}^2}{N} \bigg( d_{\text{in}} \langle w^4 \rangle + d_{\text{in}}(d_{\text{in}}^2-2) \langle w^2 \rangle^2 \bigg)\\
    & \qquad\qquad + \frac{4}{N} \bigg( d_{\text{in}}^2 \langle w^4 \rangle + d_{\text{in}}^2(d_{\text{in}}^2-2) \langle w^2 \rangle^2 \bigg)\\
    & \qquad\qquad + \frac{d_{\text{in}}^2}{N} \bigg( d_{\text{in}} \langle w^4 \rangle + d_{\text{in}}(d_{\text{in}}^2-2) \langle w^2 \rangle^2 \bigg) \bigg)\;,\\
    & = \frac{d_{\text{in}}^3+d_{\text{in}}^2+2d_{\text{in}}+4}{Nd_{\text{in}}^2(d_{\text{in}}-1)^2(d_{\text{in}}-2)} \langle w^4 \rangle \\
    & \qquad\qquad + \frac{d_{\text{in}}^4+d_{\text{in}}^3+2d_{\text{in}}^2-4d_{\text{in}}-8}{Nd_{\text{in}}^2(d_{\text{in}}-1)^2(d_{\text{in}}-2)} \langle w^2 \rangle^2\;,\\
    \implies \hat{\sigma}(f_8) & = \sqrt{\frac{d_{\text{in}}^3+d_{\text{in}}^2+2d_{\text{in}}+4}{Nd_{\text{in}}^2(d_{\text{in}}-1)^2(d_{\text{in}}-2)} \langle w^4 \rangle} \\
    & \qquad\qquad \overline{+ \frac{d_{\text{in}}^4+d_{\text{in}}^3+2d_{\text{in}}^2-4d_{\text{in}}-8}{Nd_{\text{in}}^2(d_{\text{in}}-1)^2(d_{\text{in}}-2)} \langle w^2 \rangle^2}\;,
\end{split}
\end{equation}
\begin{equation}
\begin{split}
    \text{Var}(f_9) & = \text{Var}\bigg( \frac{1}{d_{\text{in}}^2 (d_{\text{in}}-1)} \bigg( d_{\text{in}}\hat{I}_7 - \hat{I}_{10} \bigg) \bigg)\;,\\
    & = \frac{1}{d_{\text{in}}^4 (d_{\text{in}}-1)^2} \bigg( \frac{d_{\text{in}}^2}{N}\text{Var}(I_7) + \frac{1}{N} \text{Var}(I_{10}) \bigg)\;,\\
    & = \frac{1}{d_{\text{in}}^4 (d_{\text{in}}-1)^2} \bigg( \frac{d_{\text{in}}^2}{N} \bigg( d_{\text{in}}^2 \langle w^4 \rangle + d_{\text{in}}^2(d_{\text{in}}-2) \langle w^2 \rangle^2 \bigg)\\
    & \qquad\qquad + \frac{1}{N} \bigg( d_{\text{in}}^2 \langle w^4 \rangle + d_{\text{in}}^2(d_{\text{in}}^2-2) \langle w^2 \rangle^2 \bigg) \bigg)\;,\\
    & = \frac{d_{\text{in}}^2+1}{Nd_{\text{in}}^2 (d_{\text{in}}-1)^2} \langle w^4 \rangle + \frac{d_{\text{in}}^3-d_{\text{in}}^2-2}{Nd_{\text{in}}^2 (d_{\text{in}}-1)^2} \langle w^2 \rangle^2\;,\\
    \implies \hat{\sigma}(f_9) & = \sqrt{\frac{d_{\text{in}}^2+1}{Nd_{\text{in}}^2 (d_{\text{in}}-1)^2} \langle w^4 \rangle + \frac{d_{\text{in}}^3-d_{\text{in}}^2-2}{Nd_{\text{in}}^2 (d_{\text{in}}-1)^2} \langle w^2 \rangle^2}\;,
\end{split}
\end{equation}
\begin{equation}
\begin{split}
    \text{Var}(f_{10}) & = \text{Var}\bigg( \frac{1}{d_{\text{in}}^2(d_{\text{in}}-1)\sqrt{d_{\text{in}}-2}} \bigg( d_{\text{in}}^2\hat{I}_5 - d_{\text{in}}\hat{I}_7 \\
    & \qquad\qquad - d_{\text{in}}\hat{I}_9 + 2\hat{I}_{10} - d_{\text{in}}\hat{I}_{13} \bigg) \bigg)\;,\\
    & = \frac{1}{d_{\text{in}}^4(d_{\text{in}}-1)^2(d_{\text{in}}-2)} \bigg( \frac{d_{\text{in}}^4}{N}\text{Var}(I_5) + \frac{d_{\text{in}}^2}{N}\text{Var}(I_7) \\
    & \qquad\qquad + \frac{d_{\text{in}}^2}{N}\text{Var}(I_9) + \frac{4}{N}\text{Var}(I_{10}) + \frac{d_{\text{in}}^2}{N}\text{Var}(I_{13}) \bigg) \bigg)\;,\\
    & = \frac{1}{d_{\text{in}}^4(d_{\text{in}}-1)^2(d_{\text{in}}-2)} \\
    & \qquad\qquad \bigg( \frac{d_{\text{in}}^4}{N} \bigg( d_{\text{in}} \langle w^4 \rangle + d_{\text{in}}(d_{\text{in}}-2) \langle w^2 \rangle^2 \bigg)\\
    & \qquad\qquad + \frac{d_{\text{in}}^2}{N} \bigg( d_{\text{in}}^2 \langle w^4 \rangle + d_{\text{in}}^2(d_{\text{in}}-2) \langle w^2 \rangle^2 \bigg)\\
    & \qquad\qquad + \frac{d_{\text{in}}^2}{N} \bigg( d_{\text{in}} \langle w^4 \rangle + d_{\text{in}}(d_{\text{in}}^2-2) \langle w^2 \rangle^2 \bigg)\\
    & \qquad\qquad + \frac{4}{N} \bigg( d_{\text{in}}^2 \langle w^4 \rangle + d_{\text{in}}^2(d_{\text{in}}^2-2) \langle w^2 \rangle^2 \bigg)\\
    & \qquad\qquad + \frac{d_{\text{in}}^2}{N} \bigg( d_{\text{in}} \langle w^4 \rangle + d_{\text{in}}(d_{\text{in}}^2-2) \langle w^2 \rangle^2 \bigg)\bigg)\;,\\
    & = \frac{d_{\text{in}}^3+d_{\text{in}}^2+2d_{\text{in}}+4}{Nd_{\text{in}}^2(d_{\text{in}}-1)^2(d_{\text{in}}-2)} \langle w^4 \rangle \\
    & \qquad\qquad + \frac{d_{\text{in}}^4+d_{\text{in}}^3+2d_{\text{in}}^2-4d_{\text{in}}-8}{Nd_{\text{in}}^2(d_{\text{in}}-1)^2(d_{\text{in}}-2)} \langle w^2 \rangle^2\;,\\
    \implies \hat{\sigma}(f_{10}) & = \sqrt{\frac{d_{\text{in}}^3+d_{\text{in}}^2+2d_{\text{in}}+4}{Nd_{\text{in}}^2(d_{\text{in}}-1)^2(d_{\text{in}}-2)} \langle w^4 \rangle} \\
    & \qquad\qquad \overline{+ \frac{d_{\text{in}}^4+d_{\text{in}}^3+2d_{\text{in}}^2-4d_{\text{in}}-8}{Nd_{\text{in}}^2(d_{\text{in}}-1)^2(d_{\text{in}}-2)} \langle w^2 \rangle^2}\;,
\end{split}
\end{equation}
\begin{equation}
\begin{split}
    \text{Var}(f_{11}) & = \text{Var}\bigg( -\frac{1}{d_{\text{in}}^2(d_{\text{in}}-1)(d_{\text{in}}-2)} \bigg( 
    2d_{\text{in}}^2\hat{I}_5 + 2d_{\text{in}}^2\hat{I}_6 \\
    & \qquad\qquad - d_{\text{in}}\hat{I}_7 - d_{\text{in}}\hat{I}_8 - 2d_{\text{in}}\hat{I}_9 +4\hat{I}_{10} - d_{\text{in}}^3\hat{I}_{11} \\
    & \qquad\qquad + d_{\text{in}}^2\hat{I}_{12} - 4d_{\text{in}}\hat{I}_{13} \bigg) \bigg)\;,\\
    & = \frac{1}{d_{\text{in}}^4(d_{\text{in}}-1)^2(d_{\text{in}}-2)^2} \bigg(  \frac{4d_{\text{in}}^4}{N}\text{Var}(I_5) + \frac{4d_{\text{in}}^4}{N}\text{Var}(I_6) \\
    & \qquad\qquad + \frac{d_{\text{in}}^2}{N}\text{Var}(I_7) + \frac{d_{\text{in}}^2}{N}\text{Var}(I_8) + \frac{4d_{\text{in}}^2}{N}\text{Var}(I_9) \\
    & \qquad\qquad + \frac{16}{N}\text{Var}(I_{10}) + \frac{d_{\text{in}}^6}{N}\text{Var}(I_{11}) + \frac{d_{\text{in}}^4}{N}\text{Var}(I_{12}) \\
    & \qquad\qquad + \frac{16d_{\text{in}}^2}{N}\text{Var}(I_{13})  \bigg)\;,\\
    & = \frac{1}{d_{\text{in}}^4(d_{\text{in}}-1)^2(d_{\text{in}}-2)^2} \\
    & \qquad\qquad \bigg( \frac{4d_{\text{in}}^4}{N} \bigg( d_{\text{in}} \langle w^4 \rangle + d_{\text{in}}(d_{\text{in}}-2) \langle w^2 \rangle^2 \bigg)\\
    & \qquad\qquad + \frac{4d_{\text{in}}^4}{N} \bigg( d_{\text{in}} \langle w^4 \rangle + d_{\text{in}}(d_{\text{in}}-2) \langle w^2 \rangle^2 \bigg)\\
    & \qquad\qquad + \frac{d_{\text{in}}^2}{N} \bigg( d_{\text{in}}^2 \langle w^4 \rangle + d_{\text{in}}^2(d_{\text{in}}-2) \langle w^2 \rangle^2 \bigg)\\
    & \qquad\qquad + \frac{d_{\text{in}}^2}{N} \bigg( d_{\text{in}}^2 \langle w^4 \rangle + d_{\text{in}}^2(d_{\text{in}}-2) \langle w^2 \rangle^2 \bigg)\\
    & \qquad\qquad + \frac{4d_{\text{in}}^2}{N} \bigg( d_{\text{in}} \langle w^4 \rangle + d_{\text{in}}(d_{\text{in}}^2-2) \langle w^2 \rangle^2 \bigg)\\
    & \qquad\qquad + \frac{16}{N} \bigg( d_{\text{in}}^2 \langle w^4 \rangle + d_{\text{in}}^2(d_{\text{in}}^2-2) \langle w^2 \rangle^2 \bigg)\\
    & \qquad\qquad + \frac{d_{\text{in}}^6}{N} \bigg( d_{\text{in}} (\langle w^4 \rangle - \langle w^2 \rangle^2) \bigg)\\
    & \qquad\qquad + \frac{d_{\text{in}}^4}{N} \bigg( d_{\text{in}} \langle w^4 \rangle + d_{\text{in}}(d_{\text{in}}-2) \langle w^2 \rangle^2 \bigg)\\
    & \qquad\qquad + \frac{16d_{\text{in}}^2}{N} \bigg( d_{\text{in}} \langle w^4 \rangle + d_{\text{in}}(d_{\text{in}}^2-2) \langle w^2 \rangle^2 \bigg)\bigg)\;,\\
    & = \frac{(d_{\text{in}}^2-d_{\text{in}}+4)(d_{\text{in}}^3+d_{\text{in}}^2+6d_{\text{in}}+4)}{Nd_{\text{in}}^2(d_{\text{in}}-1)^2(d_{\text{in}}-2)^2} \langle w^4 \rangle \\
    & \qquad\qquad - \frac{d_{\text{in}}^5-9d_{\text{in}}^4-4d_{\text{in}}^3-12d_{\text{in}}^2+40d_{\text{in}}+32}{Nd_{\text{in}}^2(d_{\text{in}}-1)^2(d_{\text{in}}-2)^2} \langle w^2 \rangle^2\;,\\
    \implies \hat{\sigma}(f_{11}) & = \sqrt{\frac{(d_{\text{in}}^2-d_{\text{in}}+4)(d_{\text{in}}^3+d_{\text{in}}^2+6d_{\text{in}}+4)}{Nd_{\text{in}}^2(d_{\text{in}}-1)^2(d_{\text{in}}-2)^2} \langle w^4 \rangle} \\
    & \qquad\qquad \overline{- \frac{d_{\text{in}}^5-9d_{\text{in}}^4-4d_{\text{in}}^3-12d_{\text{in}}^2+40d_{\text{in}}+32}{Nd_{\text{in}}^2(d_{\text{in}}-1)^2(d_{\text{in}}-2)^2} \langle w^2 \rangle^2}\;,
\end{split}
\end{equation}
\begin{align*}
%\begin{aligned}
    \text{Var}(f_{12}) & = \text{Var}\bigg( \frac{1}{d_{\text{in}}(d_{\text{in}}-1)(d_{\text{in}}-2)(d_{\text{in}}-3)} \bigg( (d_{\text{in}}-1)(d_{\text{in}}-2)\hat{I}_3 \\
    & \qquad\qquad + (d_{\text{in}}-1)(d_{\text{in}}-2)\hat{I}_4 + 4(d_{\text{in}}-1)\hat{I}_5 + 4(d_{\text{in}}-1)\hat{I}_6 \\
    & \qquad\qquad - (d_{\text{in}}-1)\hat{I}_7 - (d_{\text{in}}-1)\hat{I}_8 - 2(d_{\text{in}}-1)\hat{I}_9 + 2\hat{I}_{10} \\
    & \qquad\qquad - 2d_{\text{in}}(d_{\text{in}}-1)\hat{I}_{11} + 2\hat{I}_{12} - 4\hat{I}_{13} \bigg) \bigg)\;,\\
    & = \frac{1}{d_{\text{in}}^2(d_{\text{in}}-1)^2(d_{\text{in}}-2)^2(d_{\text{in}}-3)^2} \bigg( \frac{(d_{\text{in}}-1)^2(d_{\text{in}}-2)^2}{N}\text{Var}(I_3) \\
    & \qquad\qquad + \frac{(d_{\text{in}}-1)^2(d_{\text{in}}-2)^2}{N}\text{Var}(I_4) + \frac{16(d_{\text{in}}-1)^2}{N}\text{Var}(I_5) \\
    & \qquad\qquad + \frac{16(d_{\text{in}}-1)^2}{N}\text{Var}(I_6) + \frac{(d_{\text{in}}-1)^2}{N}\text{Var}(I_7) \\
    & \qquad\qquad + \frac{(d_{\text{in}}-1)^2}{N}\text{Var}(I_8) + \frac{4(d_{\text{in}}-1)^2}{N}\text{Var}(I_9) \\
    & \qquad\qquad + \frac{4}{N}\text{Var}(I_{10}) + \frac{4d_{\text{in}}^2(d_{\text{in}}-1)^2}{N}\text{Var}(I_{11}) + \frac{4}{N}\text{Var}(I_{12}) \\
    & \qquad\qquad + \frac{16}{N}\text{Var}(I_{13}) \bigg) \bigg)\;,\\
    & = \frac{1}{d_{\text{in}}^2(d_{\text{in}}-1)^2(d_{\text{in}}-2)^2(d_{\text{in}}-3)^2} \\
    & \qquad\qquad \bigg(\frac{(d_{\text{in}}-1)^2(d_{\text{in}}-2)^2}{N} \bigg( d_{\text{in}}^2 (\langle w^4 \rangle - \langle w^2 \rangle^2)  \bigg)\\
    & \qquad\qquad + \frac{(d_{\text{in}}-1)^2(d_{\text{in}}-2)^2}{N} \bigg( d_{\text{in}} \langle w^4 \rangle + d_{\text{in}}(d_{\text{in}}-2) \langle w^2 \rangle^2 \bigg)\\
    & \qquad\qquad + \frac{16(d_{\text{in}}-1)^2}{N} \bigg( d_{\text{in}} \langle w^4 \rangle + d_{\text{in}}(d_{\text{in}}-2) \langle w^2 \rangle^2 \bigg)\\
    & \qquad\qquad + \frac{16(d_{\text{in}}-1)^2}{N} \bigg( d_{\text{in}} \langle w^4 \rangle + d_{\text{in}}(d_{\text{in}}-2) \langle w^2 \rangle^2 \bigg)\\
    & \qquad\qquad + \frac{(d_{\text{in}}-1)^2}{N} \bigg( d_{\text{in}}^2 \langle w^4 \rangle + d_{\text{in}}^2(d_{\text{in}}-2) \langle w^2 \rangle^2 \bigg)\\
    & \qquad\qquad + \frac{(d_{\text{in}}-1)^2}{N} \bigg( d_{\text{in}}^2 \langle w^4 \rangle + d_{\text{in}}^2(d_{\text{in}}-2) \langle w^2 \rangle^2 \bigg)\\
    & \qquad\qquad + \frac{4(d_{\text{in}}-1)^2}{N} \bigg( d_{\text{in}} \langle w^4 \rangle + d_{\text{in}}(d_{\text{in}}^2-2) \langle w^2 \rangle^2 \bigg)\\
    & \qquad\qquad + \frac{4}{N} \bigg( d_{\text{in}}^2 \langle w^4 \rangle + d_{\text{in}}^2(d_{\text{in}}^2-2) \langle w^2 \rangle^2 \bigg)\\
    & \qquad\qquad + \frac{4d_{\text{in}}^2(d_{\text{in}}-1)^2}{N} \bigg( d_{\text{in}} (\langle w^4 \rangle - \langle w^2 \rangle^2) \bigg)\\
    & \qquad\qquad + \frac{4}{N} \bigg( d_{\text{in}} \langle w^4 \rangle + d_{\text{in}}(d_{\text{in}}-2) \langle w^2 \rangle^2 \bigg)\\
    & \qquad\qquad + \frac{16}{N} \bigg( d_{\text{in}} \langle w^4 \rangle + d_{\text{in}}(d_{\text{in}}^2-2) \langle w^2 \rangle^2 \bigg) \bigg)\;,\\
    & = \frac{d_{\text{in}}^5-d_{\text{in}}^4+d_{\text{in}}^3+37d_{\text{in}}^2-74d_{\text{in}}+60}{Nd_{\text{in}}(d_{\text{in}}-1)^2(d_{\text{in}}-2)^2(d_{\text{in}}-3)^2} \langle w^4 \rangle\\
    & \qquad\qquad + \frac{8(5d_{\text{in}}^3-17d_{\text{in}}^2+24d_{\text{in}}-15)}{Nd_{\text{in}}(d_{\text{in}}-1)^2(d_{\text{in}}-2)^2(d_{\text{in}}-3)^2} \langle w^2 \rangle^2 \;,\\
    \implies \hat{\sigma}(f_{12}) & = \sqrt{\frac{d_{\text{in}}^5-d_{\text{in}}^4+d_{\text{in}}^3+37d_{\text{in}}^2-74d_{\text{in}}+60}{Nd_{\text{in}}(d_{\text{in}}-1)^2(d_{\text{in}}-2)^2(d_{\text{in}}-3)^2} \langle w^4 \rangle}\\
    & \qquad\qquad \overline{+ \frac{8(5d_{\text{in}}^3-17d_{\text{in}}^2+24d_{\text{in}}-15)}{Nd_{\text{in}}(d_{\text{in}}-1)^2(d_{\text{in}}-2)^2(d_{\text{in}}-3)^2} \langle w^2 \rangle^2}\;,
%\end{aligned}
\end{align*}
\begin{equation}
\begin{split}
    \text{Var}(f_{13}) & = \text{Var}\bigg( \frac{1}{d_{\text{in}}(d_{\text{in}}-1)(d_{\text{in}}-2)} \bigg( d_{\text{in}}\hat{I}_3 - d_{\text{in}}\hat{I}_4 \\
    & \qquad\qquad - \hat{I}_7 - \hat{I}_8 + 2\hat{I}_9 \bigg) \bigg)\;,\\
    & = \frac{1}{d_{\text{in}}^2(d_{\text{in}}-1)^2(d_{\text{in}}-2)^2} \bigg(\frac{d_{\text{in}}^2}{N} \text{Var}(I_3) + \frac{d_{\text{in}}^2}{N}\text{Var}(I_4) \\
    & \qquad\qquad + \frac{1}{N}\text{Var}(I_7) + \frac{1}{N}\text{Var}(I_8) + \frac{4}{N}\text{Var}(I_9) \bigg) \bigg)\;,\\
    & = \frac{1}{d_{\text{in}}^2(d_{\text{in}}-1)^2(d_{\text{in}}-2)^2} \\
    & \qquad\qquad \bigg( \frac{d_{\text{in}}^2}{N} \bigg( d_{\text{in}}^2 (\langle w^4 \rangle - \langle w^2 \rangle^2) \bigg)\\
    & \qquad\qquad + \frac{d_{\text{in}}^2}{N} \bigg( d_{\text{in}} \langle w^4 \rangle + d_{\text{in}}(d_{\text{in}}-2) \langle w^2 \rangle^2 \bigg)\\
    & \qquad\qquad + \frac{1}{N} \bigg( d_{\text{in}}^2 \langle w^4 \rangle + d_{\text{in}}^2(d_{\text{in}}-2) \langle w^2 \rangle^2 \bigg)\\
    & \qquad\qquad + \frac{1}{N} \bigg( d_{\text{in}}^2 \langle w^4 \rangle + d_{\text{in}}^2(d_{\text{in}}-2) \langle w^2 \rangle^2 \bigg)\\
    & \qquad\qquad + \frac{4}{N} \bigg( d_{\text{in}} \langle w^4 \rangle + d_{\text{in}}(d_{\text{in}}^2-2) \langle w^2 \rangle^2 \bigg)\bigg)\;,\\
    & = \frac{d_{\text{in}}^3+d_{\text{in}}^2+2d_{\text{in}}+4}{Nd_{\text{in}}(d_{\text{in}}-1)^2(d_{\text{in}}-2)^2} \langle w^4 \rangle \\
    & \qquad\qquad + \frac{4(d_{\text{in}}+1)}{Nd_{\text{in}}(d_{\text{in}}-1)^2(d_{\text{in}}-2)} \langle w^2 \rangle^2\;,\\
    \implies \hat{\sigma}(f_{13}) & = \sqrt{\frac{d_{\text{in}}^3+d_{\text{in}}^2+2d_{\text{in}}+4}{Nd_{\text{in}}(d_{\text{in}}-1)^2(d_{\text{in}}-2)^2} \langle w^4 \rangle} \\
    & \qquad\qquad \overline{+ \frac{4(d_{\text{in}}+1)}{Nd_{\text{in}}(d_{\text{in}}-1)^2(d_{\text{in}}-2)} \langle w^2 \rangle^2}\;.
\end{split}
\end{equation}

For the investigations carried out in this work, $d_{\text{in}}=10$, $N=1000$, and $\langle w^k \rangle$ are given as in \eqref{eq:gauss_w_exp_start}-\eqref{eq:gauss_w_exp_final} and \eqref{eq:uniform_w_exp_start}-\eqref{eq:uniform_w_exp_final} for the Gaussian and Uniform initialisation schemes respectively.
With these values substituted, the final standard error values are still rather complicated, hence are only given directly in the main text rounded appropriately, in Table \ref{tab:param_analytic}.

%%%%%%%%%%%%%%%%%%%%%%%%%%%%%%%%%%%%%%%%%%%%%%%%%
\section{PIGMM Wasserstein Distances}\label{app:wasserstein_derivation}
PIGMMs are special cases of multi-variate Gaussian distributions for the $d^2$ matrix elements variables in $W_{ij}$. 
As reviewed in \cite{Ramgoolam:2018xty}, the construction of the PIGMM is conveniently done by changing variables to a set of $d^2$ linear combinations $ S^{\mathcal{A}} $ where $\mathcal{A}$ consists of representation-theoretic labels. 

Wasserstein distance between two multi-variate Gaussian distributions with means $ \mu_1 , \mu_2 $ and covariance matrices $ \Sigma_1 , \Sigma_2) $ is
\bea 
\mathfrak{d}^2 = ( \mu_1 - \mu_2) \cdot (\mu_1 - \mu_2 ) + \text{Tr} ( \Sigma_1 + \Sigma_2 - 2 ( \Sigma_1^{ 1/2} \Sigma_2 \Sigma_1^{1/2} )^{ 1/2 } )\;. 
\eea 

For reference the parameters $ \mu , \Sigma $ of a multi-variate Gaussian define the probability density function 
\begin{equation}\label{eq:multvar}
\begin{split}
   f ( x_1 , \cdots , x_N ) = & \ N \exp  \left (  - \frac{1}{2} \sum_{ i , j } ( x_i - \mu_i ) \Sigma^{-1}_{ ij} ( x_j - \mu_j )   \right ) \;, \\ 
  = & \ N  \exp  \left ( \sum_{ i , j } x_i \Sigma^{-1}_{ij} \mu_j   - \frac{1}{2}  \sum_{ i , j }  x_i\Sigma^{-1}_{ ij} x_j  - \frac{1}{2}  \sum_{ i , j } \mu_i \Sigma^{-1}_{ ij}  \mu_j \right ) \;,  
\end{split}
\end{equation}
where the normalisation factor $N$ is 
\bea 
N = \frac{1}{(2\pi)^{k/2} \sqrt{\text{Det}(\Sigma)}} \;,
\eea 
and $\Sigma$ is a positive semi-definite symmetric matrix. 

Remark that the covariance matrix $ \Sigma $ is schematically related to the inverses of the coupling matrices $ \Lambda $ in \cite{Ramgoolam:2018xty}. 
To get the square root matrices, we use the fact that symmetric matrices can be diagonalised. 
\bea 
\Sigma = U D U^{ T} \;,
\eea
where $D$ is diagonal and $U$ is orthogonal, i.e. $UU^T =1 $, such that the square root is 
\bea 
\Sigma^{ 1/2} = U D^{ 1/2} U^T \;.
\eea

In terms of representation theory variables $S^{ \mathcal{A} } $ the probability density function is a product over the irreducible representations. For the $V_0$ variables, namely $S^{ V_0 ; 1 } , S^{ V_0  ; 2 } $, the covariance matrix is a two by two matrix: 
\bea 
\Sigma_{ a b } =  ( \Lambda_{ V_0 }^{-1})_{ ab } \;.
\eea
For the linear parameters, let us temporarily call $ \mu^{PIGMM} $ the $\mu$ parameters used in \cite{Ramgoolam:2018xty}, and $ \mu^{MV} $ the linear parameters in \eqref{eq:multvar}, then
\bea 
\mu_a^{PIGMM} = \sum_b \Sigma^{-1}_{ab} \mu_b^{MV} = \sum_b (\Lambda_{V_0})_{ab} \mu_b^{MV} \;.
\eea
We also have from \cite{Ramgoolam:2018xty} that 
\bea 
\widetilde{ \mu}_a^{PIGMM} = \sum_{ b } ( \Lambda^{ -1}_{V_0} )_{ ab } \mu^{PIGMM}_{ b }  \;,
\eea
and hence 
\bea 
\widetilde \mu_{a}^{PIGMM} = \mu_{ a}^{MV} \;.
\eea

In the $V_H$ sector, we have $ 3 ( d-1) $ variables, namely $S^{ V_H ; a }_{ m } $ with 
$ a \in \{ 1,2,3 \}$ and $ m \in \{ 1, \cdots , ( d-1) \} $. The covariance matrix is 
\bea 
( \Sigma_{H} )_{ m_1  , a_1 ; m_2 , a_2 } = \delta_{ m_1 , m_2} (\Lambda^{-1}_{ V_H })_{ a_1 a_2 } \;,
\eea
and the linear parameter for these variables is zero. 

In the $ V_2$ sector we have $\frac{d(d-3)}{2}$ variables $ S^{ V_2}_{ m } $ with $ m \in \{ 1, 2, \cdots , \frac{d(d-3)}{2}\}  $ and the covariance matrix is
\bea 
( \Sigma_{V_2} )_{ m_1  ; m_2  } = \Lambda_{ V_2}^{-1} \delta_{ m_1 , m_2 } \;,
\eea
with the linear parameters zero. 

Similarly in the $V_3$ sector, we have $\frac{(d-1)(d-2)}{2}$ variables $ S^{V_3}_{ m }$. 
The covariance matrix is 
\bea 
( \Sigma_{V_3} )_{ m_1  ; m_2 } = \Lambda_{V_3}^{-1} \delta_{ m_1 , m_2 } \;,
\eea
also with the linear parameters zero. 

Using these facts, suppose we have two sets of PIGMM-parameters $\{ \tilde \mu_1 , \tilde \mu_2 , \Sigma = [ \Lambda_{ V_0} , \Lambda_{ V_H } , \Lambda_{ V_2 } , \Lambda_{ V_3} ] \}$ and $\{  \tilde \mu_1' , \tilde \mu_2' , \Sigma' = [ \Lambda_{ V_0}' , \Lambda_{ V_H }' , \Lambda_{ V_2 }' , \Lambda_{ V_3}' ]  \}$, the Wassermann distance is defined
\begin{equation}
    \mathfrak{d}^2 = ( \widetilde \mu_1 - \widetilde \mu_1')^2 + ( \widetilde \mu_2 - \widetilde \mu_2')^2 + \text{Tr} \Sigma + \text{Tr} \Sigma' - 2 \text{Tr}  ( \Sigma^{1/2}  \Sigma' \Sigma^{1/2} )^{1/2} \;,
\end{equation}
where the trace terms can be computed as
\begin{align}
    \text{Tr}  ( \Sigma )  &= \sum_{ a =1 } ^2 ( \Lambda_{ V_0 } )_{ aa} + (d-1) \sum_{ a=1}^{3}  ( \Lambda_{V_H}^{-1} )_{ aa} +  \frac{d(d -3)}{2}  \Lambda_{ V_2}^{-1} + \frac{(d-1)(d-2)}{2} \Lambda_{ V_3}^{ -1} \;,\\
    \text{Tr}  ( \Sigma' )  &= \sum_{ a =1 } ^2 ( \Lambda_{ V_0 }' )_{ aa} + (d-1) \sum_{ a =1}^{3} ( \Lambda_{V_H}^{'-1} )_{ aa} +  \frac{d(d-3)}{2} \Lambda_{ V_2}^{'-1} + \frac{(d-1)(d-2)}{2} \Lambda_{V_3}^{' -1} \;, 
\end{align}
\begin{equation}
\begin{split}
    -2 \text{Tr}  ( \Sigma^{1/2}  \Sigma' \Sigma^{1/2} )^{1/2} = & -2 \text{Tr}  \left (   (\Lambda_{ V_0}^{-1} )^{1/2}  (\Lambda_{ V_0}')^{-1} \Lambda_{ V_0}^{-1/2}   \right )^{1/2} \\
    & -2(d-1)  \text{Tr}  \left (   (\Lambda_{ V_H}^{-1} )^{1/2}  (\Lambda_{ V_H}')^{-1} \Lambda_{ V_H}^{-1/2}   \right )^{1/2} \cr 
    &  - d(d-3) ~~   \Lambda_{ V_2}^{-1/2} \Lambda_{ V_2'}^{ -1/2} \cr
    & - (d-1)(d-2)  \Lambda_{ V_3}^{ -1/2}  \Lambda_{ V_3}^{' -1/2} \;.
\end{split}
\end{equation}

%%%%%%%%%%%%%%%%%%%%%%%%%%%%%%%%%%%%%%%%%%%%%%%%%
\addcontentsline{toc}{section}{References}
\bibliographystyle{utphys}
\bibliography{references}{}

\end{document}